\documentclass[11pt]{article}

\usepackage[margin=1in]{geometry}
\usepackage[T1]{fontenc}
\usepackage{lmodern}
\usepackage{microtype}
\usepackage{amsmath,amssymb}
\usepackage{booktabs}
\usepackage{multirow}
\usepackage{graphicx}
\usepackage{placeins}
\usepackage{xcolor}
\usepackage{fancyvrb}
\usepackage{upquote}
\usepackage{tikz}
\usetikzlibrary{arrows.meta,decorations.pathreplacing}
\usepackage{pgfplots}
\pgfplotsset{compat=1.18}
\usepackage[numbers,sort&compress]{natbib}
\usepackage[colorlinks=true,linkcolor=blue!50!black,citecolor=blue!50!black,urlcolor=blue!50!black]{hyperref}
\usepackage{cleveref}

\newcommand{\tbd}{{\color{orange}\textbf{X}}}

\newcommand{\vect}[1]{\mathbf{#1}}

\providecommand{\res}[2]{\ifcsname res@#1@#2\endcsname\csname res@#1@#2\endcsname\else\tbd\fi}
\expandafter\def\csname res@R01-R07@ag_news\endcsname{$-$2.6}
\expandafter\def\csname res@R01-R07@ag_newsp\endcsname{4e-19}
\expandafter\def\csname res@R01-R07@banking77\endcsname{+0.5}
\expandafter\def\csname res@R01-R07@banking77p\endcsname{0.1}
\expandafter\def\csname res@R01-R07@belebele_en\endcsname{+0.3}
\expandafter\def\csname res@R01-R07@belebele_enp\endcsname{0.9}
\expandafter\def\csname res@R01-R07@boolq\endcsname{+1.0}
\expandafter\def\csname res@R01-R07@boolqp\endcsname{0.1}
\expandafter\def\csname res@R01-R07@clinc150\endcsname{$-$0.1}
\expandafter\def\csname res@R01-R07@clinc150p\endcsname{0.7}
\expandafter\def\csname res@R01-R07@cosmosqa\endcsname{+0.7}
\expandafter\def\csname res@R01-R07@cosmosqap\endcsname{0.4}
\expandafter\def\csname res@R01-R07@dbpedia14\endcsname{+0.3}
\expandafter\def\csname res@R01-R07@dbpedia14p\endcsname{0.0002}
\expandafter\def\csname res@R01-R07@emotion\endcsname{$-$0.2}
\expandafter\def\csname res@R01-R07@emotionp\endcsname{0.7}
\expandafter\def\csname res@R01-R07@helpsteer2\endcsname{$-$0.3}
\expandafter\def\csname res@R01-R07@helpsteer2p\endcsname{0.4}
\expandafter\def\csname res@R01-R07@id\endcsname{+0.11}
\expandafter\def\csname res@R01-R07@idhi\endcsname{+0.34}
\expandafter\def\csname res@R01-R07@idlo\endcsname{$-$0.14}
\expandafter\def\csname res@R01-R07@kev_transfer\endcsname{+0.0}
\expandafter\def\csname res@R01-R07@kev_transferp\endcsname{1}
\expandafter\def\csname res@R01-R07@massive_en\endcsname{+0.2}
\expandafter\def\csname res@R01-R07@massive_enp\endcsname{0.4}
\expandafter\def\csname res@R01-R07@mnli\endcsname{+0.0}
\expandafter\def\csname res@R01-R07@mnlip\endcsname{0.9}
\expandafter\def\csname res@R01-R07@oasst1\endcsname{$-$0.1}
\expandafter\def\csname res@R01-R07@oasst1p\endcsname{0.6}
\expandafter\def\csname res@R01-R07@ood\endcsname{$-$0.68}
\expandafter\def\csname res@R01-R07@oodhi\endcsname{+0.12}
\expandafter\def\csname res@R01-R07@oodlo\endcsname{$-$1.48}
\expandafter\def\csname res@R01-R07@procedural_typed_decisions\endcsname{+0.2}
\expandafter\def\csname res@R01-R07@procedural_typed_decisionsp\endcsname{0.3}
\expandafter\def\csname res@R01-R07@qasc\endcsname{$-$1.4}
\expandafter\def\csname res@R01-R07@qascp\endcsname{0.2}
\expandafter\def\csname res@R01-R07@snli\endcsname{$-$0.2}
\expandafter\def\csname res@R01-R07@snlip\endcsname{0.4}
\expandafter\def\csname res@R01-R07@squad2\endcsname{$-$1.0}
\expandafter\def\csname res@R01-R07@squad2p\endcsname{0.001}
\expandafter\def\csname res@R01-R07@sst5\endcsname{$-$0.5}
\expandafter\def\csname res@R01-R07@sst5p\endcsname{0.6}
\expandafter\def\csname res@R01-R07@tydiqa\endcsname{+0.3}
\expandafter\def\csname res@R01-R07@tydiqap\endcsname{0.8}
\expandafter\def\csname res@R01-R07@typed_decisions\endcsname{+1.2}
\expandafter\def\csname res@R01-R07@typed_decisionsp\endcsname{0.08}
\expandafter\def\csname res@R01-R68@ag_news\endcsname{$-$4.2}
\expandafter\def\csname res@R01-R68@ag_newsp\endcsname{6e-39}
\expandafter\def\csname res@R01-R68@banking77\endcsname{+0.4}
\expandafter\def\csname res@R01-R68@banking77p\endcsname{0.2}
\expandafter\def\csname res@R01-R68@belebele_en\endcsname{$-$0.3}
\expandafter\def\csname res@R01-R68@belebele_enp\endcsname{0.9}
\expandafter\def\csname res@R01-R68@boolq\endcsname{$-$0.3}
\expandafter\def\csname res@R01-R68@boolqp\endcsname{0.6}
\expandafter\def\csname res@R01-R68@clinc150\endcsname{$-$0.4}
\expandafter\def\csname res@R01-R68@clinc150p\endcsname{0.1}
\expandafter\def\csname res@R01-R68@cosmosqa\endcsname{+0.7}
\expandafter\def\csname res@R01-R68@cosmosqap\endcsname{0.4}
\expandafter\def\csname res@R01-R68@dbpedia14\endcsname{$-$1.8}
\expandafter\def\csname res@R01-R68@dbpedia14p\endcsname{2e-166}
\expandafter\def\csname res@R01-R68@emotion\endcsname{+0.4}
\expandafter\def\csname res@R01-R68@emotionp\endcsname{0.3}
\expandafter\def\csname res@R01-R68@helpsteer2\endcsname{$-$0.7}
\expandafter\def\csname res@R01-R68@helpsteer2p\endcsname{0.06}
\expandafter\def\csname res@R01-R68@id\endcsname{$-$0.10}
\expandafter\def\csname res@R01-R68@idhi\endcsname{+0.13}
\expandafter\def\csname res@R01-R68@idlo\endcsname{$-$0.33}
\expandafter\def\csname res@R01-R68@kev_transfer\endcsname{$-$0.8}
\expandafter\def\csname res@R01-R68@kev_transferp\endcsname{0.5}
\expandafter\def\csname res@R01-R68@massive_en\endcsname{+0.2}
\expandafter\def\csname res@R01-R68@massive_enp\endcsname{0.4}
\expandafter\def\csname res@R01-R68@mnli\endcsname{+0.1}
\expandafter\def\csname res@R01-R68@mnlip\endcsname{0.9}
\expandafter\def\csname res@R01-R68@oasst1\endcsname{$-$0.1}
\expandafter\def\csname res@R01-R68@oasst1p\endcsname{0.6}
\expandafter\def\csname res@R01-R68@ood\endcsname{$-$1.91}
\expandafter\def\csname res@R01-R68@oodhi\endcsname{$-$1.13}
\expandafter\def\csname res@R01-R68@oodlo\endcsname{$-$2.65}
\expandafter\def\csname res@R01-R68@procedural_typed_decisions\endcsname{$-$0.3}
\expandafter\def\csname res@R01-R68@procedural_typed_decisionsp\endcsname{0.07}
\expandafter\def\csname res@R01-R68@qasc\endcsname{$-$2.5}
\expandafter\def\csname res@R01-R68@qascp\endcsname{0.03}
\expandafter\def\csname res@R01-R68@snli\endcsname{+0.3}
\expandafter\def\csname res@R01-R68@snlip\endcsname{0.2}
\expandafter\def\csname res@R01-R68@squad2\endcsname{$-$0.7}
\expandafter\def\csname res@R01-R68@squad2p\endcsname{0.02}
\expandafter\def\csname res@R01-R68@sst5\endcsname{$-$0.5}
\expandafter\def\csname res@R01-R68@sst5p\endcsname{0.6}
\expandafter\def\csname res@R01-R68@tydiqa\endcsname{$-$0.4}
\expandafter\def\csname res@R01-R68@tydiqap\endcsname{0.6}
\expandafter\def\csname res@R01-R68@typed_decisions\endcsname{$-$0.2}
\expandafter\def\csname res@R01-R68@typed_decisionsp\endcsname{0.8}
\expandafter\def\csname res@R01@ag_news\endcsname{77.9}
\expandafter\def\csname res@R01@banking77\endcsname{93.2}
\expandafter\def\csname res@R01@belebele_en\endcsname{66.0}
\expandafter\def\csname res@R01@boolq\endcsname{84.5}
\expandafter\def\csname res@R01@clinc150\endcsname{92.3}
\expandafter\def\csname res@R01@cosmosqa\endcsname{74.9}
\expandafter\def\csname res@R01@dbpedia14\endcsname{95.1}
\expandafter\def\csname res@R01@emotion\endcsname{93.0}
\expandafter\def\csname res@R01@helpsteer2\endcsname{59.7}
\expandafter\def\csname res@R01@id\endcsname{81.9}
\expandafter\def\csname res@R01@idece\endcsname{0.034}
\expandafter\def\csname res@R01@kev_transfer\endcsname{55.4}
\expandafter\def\csname res@R01@massive_en\endcsname{91.3}
\expandafter\def\csname res@R01@mnli\endcsname{84.1}
\expandafter\def\csname res@R01@oasst1\endcsname{78.8}
\expandafter\def\csname res@R01@ood\endcsname{76.1}
\expandafter\def\csname res@R01@oodece\endcsname{0.077}
\expandafter\def\csname res@R01@peakgb\endcsname{6.7}
\expandafter\def\csname res@R01@procedural_typed_decisions\endcsname{81.7}
\expandafter\def\csname res@R01@qasc\endcsname{86.0}
\expandafter\def\csname res@R01@snli\endcsname{87.7}
\expandafter\def\csname res@R01@squad2\endcsname{83.7}
\expandafter\def\csname res@R01@sst5\endcsname{55.9}
\expandafter\def\csname res@R01@sstep\endcsname{4.41}
\expandafter\def\csname res@R01@thead\endcsname{1.09}
\expandafter\def\csname res@R01@tydiqa\endcsname{91.0}
\expandafter\def\csname res@R01@typed_decisions\endcsname{76.1}
\expandafter\def\csname res@R04-R07@ag_news\endcsname{$-$0.8}
\expandafter\def\csname res@R04-R07@ag_newsp\endcsname{0.002}
\expandafter\def\csname res@R04-R07@banking77\endcsname{+0.2}
\expandafter\def\csname res@R04-R07@banking77p\endcsname{0.7}
\expandafter\def\csname res@R04-R07@belebele_en\endcsname{+1.7}
\expandafter\def\csname res@R04-R07@belebele_enp\endcsname{0.2}
\expandafter\def\csname res@R04-R07@boolq\endcsname{+0.8}
\expandafter\def\csname res@R04-R07@boolqp\endcsname{0.2}
\expandafter\def\csname res@R04-R07@clinc150\endcsname{$-$0.5}
\expandafter\def\csname res@R04-R07@clinc150p\endcsname{0.1}
\expandafter\def\csname res@R04-R07@cosmosqa\endcsname{$-$0.7}
\expandafter\def\csname res@R04-R07@cosmosqap\endcsname{0.4}
\expandafter\def\csname res@R04-R07@dbpedia14\endcsname{+0.6}
\expandafter\def\csname res@R04-R07@dbpedia14p\endcsname{1e-18}
\expandafter\def\csname res@R04-R07@emotion\endcsname{$-$0.6}
\expandafter\def\csname res@R04-R07@emotionp\endcsname{0.2}
\expandafter\def\csname res@R04-R07@helpsteer2\endcsname{$-$0.3}
\expandafter\def\csname res@R04-R07@helpsteer2p\endcsname{0.4}
\expandafter\def\csname res@R04-R07@id\endcsname{+0.08}
\expandafter\def\csname res@R04-R07@idhi\endcsname{+0.32}
\expandafter\def\csname res@R04-R07@idlo\endcsname{$-$0.16}
\expandafter\def\csname res@R04-R07@kev_transfer\endcsname{+1.0}
\expandafter\def\csname res@R04-R07@kev_transferp\endcsname{0.5}
\expandafter\def\csname res@R04-R07@massive_en\endcsname{$-$0.2}
\expandafter\def\csname res@R04-R07@massive_enp\endcsname{0.3}
\expandafter\def\csname res@R04-R07@mnli\endcsname{+0.2}
\expandafter\def\csname res@R04-R07@mnlip\endcsname{0.5}
\expandafter\def\csname res@R04-R07@oasst1\endcsname{$-$0.1}
\expandafter\def\csname res@R04-R07@oasst1p\endcsname{0.7}
\expandafter\def\csname res@R04-R07@ood\endcsname{+0.07}
\expandafter\def\csname res@R04-R07@oodhi\endcsname{+0.86}
\expandafter\def\csname res@R04-R07@oodlo\endcsname{$-$0.73}
\expandafter\def\csname res@R04-R07@procedural_typed_decisions\endcsname{+0.4}
\expandafter\def\csname res@R04-R07@procedural_typed_decisionsp\endcsname{0.007}
\expandafter\def\csname res@R04-R07@qasc\endcsname{$-$2.1}
\expandafter\def\csname res@R04-R07@qascp\endcsname{0.05}
\expandafter\def\csname res@R04-R07@snli\endcsname{$-$0.2}
\expandafter\def\csname res@R04-R07@snlip\endcsname{0.5}
\expandafter\def\csname res@R04-R07@squad2\endcsname{+0.2}
\expandafter\def\csname res@R04-R07@squad2p\endcsname{0.5}
\expandafter\def\csname res@R04-R07@sst5\endcsname{$-$0.3}
\expandafter\def\csname res@R04-R07@sst5p\endcsname{0.7}
\expandafter\def\csname res@R04-R07@tydiqa\endcsname{$-$0.1}
\expandafter\def\csname res@R04-R07@tydiqap\endcsname{1}
\expandafter\def\csname res@R04-R07@typed_decisions\endcsname{+2.4}
\expandafter\def\csname res@R04-R07@typed_decisionsp\endcsname{0.0008}
\expandafter\def\csname res@R04@ag_news\endcsname{79.6}
\expandafter\def\csname res@R04@banking77\endcsname{93.0}
\expandafter\def\csname res@R04@belebele_en\endcsname{67.3}
\expandafter\def\csname res@R04@boolq\endcsname{84.3}
\expandafter\def\csname res@R04@clinc150\endcsname{91.9}
\expandafter\def\csname res@R04@cosmosqa\endcsname{73.5}
\expandafter\def\csname res@R04@dbpedia14\endcsname{95.5}
\expandafter\def\csname res@R04@emotion\endcsname{92.7}
\expandafter\def\csname res@R04@helpsteer2\endcsname{59.7}
\expandafter\def\csname res@R04@id\endcsname{81.8}
\expandafter\def\csname res@R04@idece\endcsname{0.034}
\expandafter\def\csname res@R04@kev_transfer\endcsname{56.3}
\expandafter\def\csname res@R04@massive_en\endcsname{90.9}
\expandafter\def\csname res@R04@mnli\endcsname{84.3}
\expandafter\def\csname res@R04@oasst1\endcsname{78.8}
\expandafter\def\csname res@R04@ood\endcsname{76.8}
\expandafter\def\csname res@R04@oodece\endcsname{0.068}
\expandafter\def\csname res@R04@peakgb\endcsname{6.3}
\expandafter\def\csname res@R04@procedural_typed_decisions\endcsname{81.9}
\expandafter\def\csname res@R04@qasc\endcsname{85.3}
\expandafter\def\csname res@R04@snli\endcsname{87.8}
\expandafter\def\csname res@R04@squad2\endcsname{84.9}
\expandafter\def\csname res@R04@sst5\endcsname{56.0}
\expandafter\def\csname res@R04@sstep\endcsname{3.31}
\expandafter\def\csname res@R04@thead\endcsname{1.08}
\expandafter\def\csname res@R04@tydiqa\endcsname{90.5}
\expandafter\def\csname res@R04@typed_decisions\endcsname{77.2}
\expandafter\def\csname res@R07@ag_news\endcsname{80.5}
\expandafter\def\csname res@R07@banking77\endcsname{92.8}
\expandafter\def\csname res@R07@belebele_en\endcsname{65.7}
\expandafter\def\csname res@R07@boolq\endcsname{83.5}
\expandafter\def\csname res@R07@clinc150\endcsname{92.4}
\expandafter\def\csname res@R07@cosmosqa\endcsname{74.2}
\expandafter\def\csname res@R07@dbpedia14\endcsname{94.9}
\expandafter\def\csname res@R07@emotion\endcsname{93.2}
\expandafter\def\csname res@R07@helpsteer2\endcsname{60.0}
\expandafter\def\csname res@R07@id\endcsname{81.7}
\expandafter\def\csname res@R07@idece\endcsname{0.033}
\expandafter\def\csname res@R07@kev_transfer\endcsname{55.4}
\expandafter\def\csname res@R07@massive_en\endcsname{91.1}
\expandafter\def\csname res@R07@mnli\endcsname{84.1}
\expandafter\def\csname res@R07@oasst1\endcsname{78.8}
\expandafter\def\csname res@R07@ood\endcsname{76.7}
\expandafter\def\csname res@R07@oodece\endcsname{0.071}
\expandafter\def\csname res@R07@peakgb\endcsname{6.3}
\expandafter\def\csname res@R07@procedural_typed_decisions\endcsname{81.5}
\expandafter\def\csname res@R07@qasc\endcsname{87.4}
\expandafter\def\csname res@R07@snli\endcsname{88.0}
\expandafter\def\csname res@R07@squad2\endcsname{84.7}
\expandafter\def\csname res@R07@sst5\endcsname{56.3}
\expandafter\def\csname res@R07@sstep\endcsname{3.24}
\expandafter\def\csname res@R07@thead\endcsname{1.07}
\expandafter\def\csname res@R07@tydiqa\endcsname{90.7}
\expandafter\def\csname res@R07@typed_decisions\endcsname{74.9}
\expandafter\def\csname res@R10-R07@ag_news\endcsname{$-$0.7}
\expandafter\def\csname res@R10-R07@ag_newsp\endcsname{0.01}
\expandafter\def\csname res@R10-R07@banking77\endcsname{+0.4}
\expandafter\def\csname res@R10-R07@banking77p\endcsname{0.3}
\expandafter\def\csname res@R10-R07@belebele_en\endcsname{+10.4}
\expandafter\def\csname res@R10-R07@belebele_enp\endcsname{6e-12}
\expandafter\def\csname res@R10-R07@boolq\endcsname{+0.6}
\expandafter\def\csname res@R10-R07@boolqp\endcsname{0.4}
\expandafter\def\csname res@R10-R07@clinc150\endcsname{$-$1.1}
\expandafter\def\csname res@R10-R07@clinc150p\endcsname{0.0001}
\expandafter\def\csname res@R10-R07@cosmosqa\endcsname{$-$0.1}
\expandafter\def\csname res@R10-R07@cosmosqap\endcsname{0.9}
\expandafter\def\csname res@R10-R07@dbpedia14\endcsname{+0.9}
\expandafter\def\csname res@R10-R07@dbpedia14p\endcsname{2e-30}
\expandafter\def\csname res@R10-R07@emotion\endcsname{+0.2}
\expandafter\def\csname res@R10-R07@emotionp\endcsname{0.6}
\expandafter\def\csname res@R10-R07@helpsteer2\endcsname{+0.0}
\expandafter\def\csname res@R10-R07@helpsteer2p\endcsname{0.9}
\expandafter\def\csname res@R10-R07@id\endcsname{$-$0.11}
\expandafter\def\csname res@R10-R07@idhi\endcsname{+0.13}
\expandafter\def\csname res@R10-R07@idlo\endcsname{$-$0.35}
\expandafter\def\csname res@R10-R07@kev_transfer\endcsname{+2.4}
\expandafter\def\csname res@R10-R07@kev_transferp\endcsname{0.06}
\expandafter\def\csname res@R10-R07@massive_en\endcsname{$-$0.1}
\expandafter\def\csname res@R10-R07@massive_enp\endcsname{0.7}
\expandafter\def\csname res@R10-R07@mnli\endcsname{$-$0.7}
\expandafter\def\csname res@R10-R07@mnlip\endcsname{0.02}
\expandafter\def\csname res@R10-R07@oasst1\endcsname{$-$0.0}
\expandafter\def\csname res@R10-R07@oasst1p\endcsname{1}
\expandafter\def\csname res@R10-R07@ood\endcsname{+2.64}
\expandafter\def\csname res@R10-R07@oodhi\endcsname{+3.50}
\expandafter\def\csname res@R10-R07@oodlo\endcsname{+1.76}
\expandafter\def\csname res@R10-R07@procedural_typed_decisions\endcsname{$-$2.2}
\expandafter\def\csname res@R10-R07@procedural_typed_decisionsp\endcsname{6e-55}
\expandafter\def\csname res@R10-R07@qasc\endcsname{+0.1}
\expandafter\def\csname res@R10-R07@qascp\endcsname{1}
\expandafter\def\csname res@R10-R07@snli\endcsname{$-$0.7}
\expandafter\def\csname res@R10-R07@snlip\endcsname{0.003}
\expandafter\def\csname res@R10-R07@squad2\endcsname{$-$0.8}
\expandafter\def\csname res@R10-R07@squad2p\endcsname{0.01}
\expandafter\def\csname res@R10-R07@sst5\endcsname{+0.2}
\expandafter\def\csname res@R10-R07@sst5p\endcsname{0.8}
\expandafter\def\csname res@R10-R07@tydiqa\endcsname{+1.1}
\expandafter\def\csname res@R10-R07@tydiqap\endcsname{0.2}
\expandafter\def\csname res@R10-R07@typed_decisions\endcsname{+1.7}
\expandafter\def\csname res@R10-R07@typed_decisionsp\endcsname{0.008}
\expandafter\def\csname res@R10-R68@ag_news\endcsname{$-$2.3}
\expandafter\def\csname res@R10-R68@ag_newsp\endcsname{1e-12}
\expandafter\def\csname res@R10-R68@banking77\endcsname{+0.4}
\expandafter\def\csname res@R10-R68@banking77p\endcsname{0.3}
\expandafter\def\csname res@R10-R68@belebele_en\endcsname{+9.8}
\expandafter\def\csname res@R10-R68@belebele_enp\endcsname{2e-10}
\expandafter\def\csname res@R10-R68@boolq\endcsname{$-$0.7}
\expandafter\def\csname res@R10-R68@boolqp\endcsname{0.3}
\expandafter\def\csname res@R10-R68@clinc150\endcsname{$-$1.4}
\expandafter\def\csname res@R10-R68@clinc150p\endcsname{4e-06}
\expandafter\def\csname res@R10-R68@cosmosqa\endcsname{$-$0.2}
\expandafter\def\csname res@R10-R68@cosmosqap\endcsname{0.9}
\expandafter\def\csname res@R10-R68@dbpedia14\endcsname{$-$1.1}
\expandafter\def\csname res@R10-R68@dbpedia14p\endcsname{3e-69}
\expandafter\def\csname res@R10-R68@emotion\endcsname{+0.9}
\expandafter\def\csname res@R10-R68@emotionp\endcsname{0.03}
\expandafter\def\csname res@R10-R68@helpsteer2\endcsname{$-$0.4}
\expandafter\def\csname res@R10-R68@helpsteer2p\endcsname{0.3}
\expandafter\def\csname res@R10-R68@id\endcsname{$-$0.31}
\expandafter\def\csname res@R10-R68@idhi\endcsname{$-$0.07}
\expandafter\def\csname res@R10-R68@idlo\endcsname{$-$0.55}
\expandafter\def\csname res@R10-R68@kev_transfer\endcsname{+1.6}
\expandafter\def\csname res@R10-R68@kev_transferp\endcsname{0.2}
\expandafter\def\csname res@R10-R68@massive_en\endcsname{$-$0.1}
\expandafter\def\csname res@R10-R68@massive_enp\endcsname{0.7}
\expandafter\def\csname res@R10-R68@mnli\endcsname{$-$0.7}
\expandafter\def\csname res@R10-R68@mnlip\endcsname{0.02}
\expandafter\def\csname res@R10-R68@oasst1\endcsname{$-$0.0}
\expandafter\def\csname res@R10-R68@oasst1p\endcsname{1}
\expandafter\def\csname res@R10-R68@ood\endcsname{+1.41}
\expandafter\def\csname res@R10-R68@oodhi\endcsname{+2.25}
\expandafter\def\csname res@R10-R68@oodlo\endcsname{+0.54}
\expandafter\def\csname res@R10-R68@procedural_typed_decisions\endcsname{$-$2.6}
\expandafter\def\csname res@R10-R68@procedural_typed_decisionsp\endcsname{6e-69}
\expandafter\def\csname res@R10-R68@qasc\endcsname{$-$1.0}
\expandafter\def\csname res@R10-R68@qascp\endcsname{0.4}
\expandafter\def\csname res@R10-R68@snli\endcsname{$-$0.2}
\expandafter\def\csname res@R10-R68@snlip\endcsname{0.4}
\expandafter\def\csname res@R10-R68@squad2\endcsname{$-$0.5}
\expandafter\def\csname res@R10-R68@squad2p\endcsname{0.09}
\expandafter\def\csname res@R10-R68@sst5\endcsname{+0.2}
\expandafter\def\csname res@R10-R68@sst5p\endcsname{0.9}
\expandafter\def\csname res@R10-R68@tydiqa\endcsname{+0.3}
\expandafter\def\csname res@R10-R68@tydiqap\endcsname{0.7}
\expandafter\def\csname res@R10-R68@typed_decisions\endcsname{+0.2}
\expandafter\def\csname res@R10-R68@typed_decisionsp\endcsname{0.8}
\expandafter\def\csname res@R10@ag_news\endcsname{79.8}
\expandafter\def\csname res@R10@banking77\endcsname{93.2}
\expandafter\def\csname res@R10@belebele_en\endcsname{76.1}
\expandafter\def\csname res@R10@boolq\endcsname{84.1}
\expandafter\def\csname res@R10@clinc150\endcsname{91.3}
\expandafter\def\csname res@R10@cosmosqa\endcsname{74.0}
\expandafter\def\csname res@R10@dbpedia14\endcsname{95.8}
\expandafter\def\csname res@R10@emotion\endcsname{93.5}
\expandafter\def\csname res@R10@helpsteer2\endcsname{60.1}
\expandafter\def\csname res@R10@id\endcsname{81.6}
\expandafter\def\csname res@R10@idece\endcsname{0.033}
\expandafter\def\csname res@R10@kev_transfer\endcsname{57.7}
\expandafter\def\csname res@R10@massive_en\endcsname{91.0}
\expandafter\def\csname res@R10@mnli\endcsname{83.4}
\expandafter\def\csname res@R10@oasst1\endcsname{78.8}
\expandafter\def\csname res@R10@ood\endcsname{79.4}
\expandafter\def\csname res@R10@oodece\endcsname{0.063}
\expandafter\def\csname res@R10@peakgb\endcsname{6.3}
\expandafter\def\csname res@R10@procedural_typed_decisions\endcsname{79.3}
\expandafter\def\csname res@R10@qasc\endcsname{87.5}
\expandafter\def\csname res@R10@snli\endcsname{87.2}
\expandafter\def\csname res@R10@squad2\endcsname{83.9}
\expandafter\def\csname res@R10@sst5\endcsname{56.6}
\expandafter\def\csname res@R10@sstep\endcsname{3.17}
\expandafter\def\csname res@R10@thead\endcsname{1.05}
\expandafter\def\csname res@R10@tydiqa\endcsname{91.7}
\expandafter\def\csname res@R10@typed_decisions\endcsname{76.6}
\expandafter\def\csname res@R13-R07@ag_news\endcsname{+0.1}
\expandafter\def\csname res@R13-R07@ag_newsp\endcsname{0.7}
\expandafter\def\csname res@R13-R07@banking77\endcsname{+0.2}
\expandafter\def\csname res@R13-R07@banking77p\endcsname{0.6}
\expandafter\def\csname res@R13-R07@belebele_en\endcsname{$-$0.3}
\expandafter\def\csname res@R13-R07@belebele_enp\endcsname{0.9}
\expandafter\def\csname res@R13-R07@boolq\endcsname{+0.6}
\expandafter\def\csname res@R13-R07@boolqp\endcsname{0.3}
\expandafter\def\csname res@R13-R07@clinc150\endcsname{+0.3}
\expandafter\def\csname res@R13-R07@clinc150p\endcsname{0.3}
\expandafter\def\csname res@R13-R07@cosmosqa\endcsname{$-$0.2}
\expandafter\def\csname res@R13-R07@cosmosqap\endcsname{0.8}
\expandafter\def\csname res@R13-R07@dbpedia14\endcsname{+0.6}
\expandafter\def\csname res@R13-R07@dbpedia14p\endcsname{7e-18}
\expandafter\def\csname res@R13-R07@emotion\endcsname{$-$0.4}
\expandafter\def\csname res@R13-R07@emotionp\endcsname{0.3}
\expandafter\def\csname res@R13-R07@helpsteer2\endcsname{+0.1}
\expandafter\def\csname res@R13-R07@helpsteer2p\endcsname{0.9}
\expandafter\def\csname res@R13-R07@id\endcsname{+0.21}
\expandafter\def\csname res@R13-R07@idhi\endcsname{+0.45}
\expandafter\def\csname res@R13-R07@idlo\endcsname{$-$0.04}
\expandafter\def\csname res@R13-R07@kev_transfer\endcsname{$-$0.8}
\expandafter\def\csname res@R13-R07@kev_transferp\endcsname{0.6}
\expandafter\def\csname res@R13-R07@massive_en\endcsname{+0.2}
\expandafter\def\csname res@R13-R07@massive_enp\endcsname{0.4}
\expandafter\def\csname res@R13-R07@mnli\endcsname{+0.1}
\expandafter\def\csname res@R13-R07@mnlip\endcsname{0.7}
\expandafter\def\csname res@R13-R07@oasst1\endcsname{+0.2}
\expandafter\def\csname res@R13-R07@oasst1p\endcsname{0.2}
\expandafter\def\csname res@R13-R07@ood\endcsname{$-$0.23}
\expandafter\def\csname res@R13-R07@oodhi\endcsname{+0.57}
\expandafter\def\csname res@R13-R07@oodlo\endcsname{$-$1.03}
\expandafter\def\csname res@R13-R07@procedural_typed_decisions\endcsname{+0.8}
\expandafter\def\csname res@R13-R07@procedural_typed_decisionsp\endcsname{2e-08}
\expandafter\def\csname res@R13-R07@qasc\endcsname{$-$0.8}
\expandafter\def\csname res@R13-R07@qascp\endcsname{0.5}
\expandafter\def\csname res@R13-R07@snli\endcsname{$-$0.2}
\expandafter\def\csname res@R13-R07@snlip\endcsname{0.3}
\expandafter\def\csname res@R13-R07@squad2\endcsname{$-$0.3}
\expandafter\def\csname res@R13-R07@squad2p\endcsname{0.3}
\expandafter\def\csname res@R13-R07@sst5\endcsname{$-$0.2}
\expandafter\def\csname res@R13-R07@sst5p\endcsname{0.9}
\expandafter\def\csname res@R13-R07@tydiqa\endcsname{+0.4}
\expandafter\def\csname res@R13-R07@tydiqap\endcsname{0.7}
\expandafter\def\csname res@R13-R07@typed_decisions\endcsname{+1.8}
\expandafter\def\csname res@R13-R07@typed_decisionsp\endcsname{0.01}
\expandafter\def\csname res@R13-R68@ag_news\endcsname{$-$1.4}
\expandafter\def\csname res@R13-R68@ag_newsp\endcsname{3e-07}
\expandafter\def\csname res@R13-R68@banking77\endcsname{+0.2}
\expandafter\def\csname res@R13-R68@banking77p\endcsname{0.7}
\expandafter\def\csname res@R13-R68@belebele_en\endcsname{$-$1.0}
\expandafter\def\csname res@R13-R68@belebele_enp\endcsname{0.5}
\expandafter\def\csname res@R13-R68@boolq\endcsname{$-$0.7}
\expandafter\def\csname res@R13-R68@boolqp\endcsname{0.3}
\expandafter\def\csname res@R13-R68@clinc150\endcsname{+0.0}
\expandafter\def\csname res@R13-R68@clinc150p\endcsname{1}
\expandafter\def\csname res@R13-R68@cosmosqa\endcsname{$-$0.3}
\expandafter\def\csname res@R13-R68@cosmosqap\endcsname{0.8}
\expandafter\def\csname res@R13-R68@dbpedia14\endcsname{$-$1.5}
\expandafter\def\csname res@R13-R68@dbpedia14p\endcsname{1e-116}
\expandafter\def\csname res@R13-R68@emotion\endcsname{+0.2}
\expandafter\def\csname res@R13-R68@emotionp\endcsname{0.7}
\expandafter\def\csname res@R13-R68@helpsteer2\endcsname{$-$0.3}
\expandafter\def\csname res@R13-R68@helpsteer2p\endcsname{0.4}
\expandafter\def\csname res@R13-R68@id\endcsname{+0.00}
\expandafter\def\csname res@R13-R68@idhi\endcsname{+0.24}
\expandafter\def\csname res@R13-R68@idlo\endcsname{$-$0.23}
\expandafter\def\csname res@R13-R68@kev_transfer\endcsname{$-$1.5}
\expandafter\def\csname res@R13-R68@kev_transferp\endcsname{0.2}
\expandafter\def\csname res@R13-R68@massive_en\endcsname{+0.2}
\expandafter\def\csname res@R13-R68@massive_enp\endcsname{0.4}
\expandafter\def\csname res@R13-R68@mnli\endcsname{+0.1}
\expandafter\def\csname res@R13-R68@mnlip\endcsname{0.7}
\expandafter\def\csname res@R13-R68@oasst1\endcsname{+0.2}
\expandafter\def\csname res@R13-R68@oasst1p\endcsname{0.1}
\expandafter\def\csname res@R13-R68@ood\endcsname{$-$1.46}
\expandafter\def\csname res@R13-R68@oodhi\endcsname{$-$0.66}
\expandafter\def\csname res@R13-R68@oodlo\endcsname{$-$2.27}
\expandafter\def\csname res@R13-R68@procedural_typed_decisions\endcsname{+0.4}
\expandafter\def\csname res@R13-R68@procedural_typed_decisionsp\endcsname{0.01}
\expandafter\def\csname res@R13-R68@qasc\endcsname{$-$1.8}
\expandafter\def\csname res@R13-R68@qascp\endcsname{0.1}
\expandafter\def\csname res@R13-R68@snli\endcsname{+0.3}
\expandafter\def\csname res@R13-R68@snlip\endcsname{0.2}
\expandafter\def\csname res@R13-R68@squad2\endcsname{$-$0.0}
\expandafter\def\csname res@R13-R68@squad2p\endcsname{0.9}
\expandafter\def\csname res@R13-R68@sst5\endcsname{$-$0.2}
\expandafter\def\csname res@R13-R68@sst5p\endcsname{0.8}
\expandafter\def\csname res@R13-R68@tydiqa\endcsname{$-$0.3}
\expandafter\def\csname res@R13-R68@tydiqap\endcsname{0.7}
\expandafter\def\csname res@R13-R68@typed_decisions\endcsname{+0.3}
\expandafter\def\csname res@R13-R68@typed_decisionsp\endcsname{0.7}
\expandafter\def\csname res@R13@ag_news\endcsname{80.6}
\expandafter\def\csname res@R13@banking77\endcsname{93.0}
\expandafter\def\csname res@R13@belebele_en\endcsname{65.3}
\expandafter\def\csname res@R13@boolq\endcsname{84.1}
\expandafter\def\csname res@R13@clinc150\endcsname{92.7}
\expandafter\def\csname res@R13@cosmosqa\endcsname{74.0}
\expandafter\def\csname res@R13@dbpedia14\endcsname{95.4}
\expandafter\def\csname res@R13@emotion\endcsname{92.8}
\expandafter\def\csname res@R13@helpsteer2\endcsname{60.1}
\expandafter\def\csname res@R13@id\endcsname{82.0}
\expandafter\def\csname res@R13@idece\endcsname{0.034}
\expandafter\def\csname res@R13@kev_transfer\endcsname{54.6}
\expandafter\def\csname res@R13@massive_en\endcsname{91.3}
\expandafter\def\csname res@R13@mnli\endcsname{84.2}
\expandafter\def\csname res@R13@oasst1\endcsname{79.0}
\expandafter\def\csname res@R13@ood\endcsname{76.5}
\expandafter\def\csname res@R13@oodece\endcsname{0.070}
\expandafter\def\csname res@R13@peakgb\endcsname{6.3}
\expandafter\def\csname res@R13@procedural_typed_decisions\endcsname{82.3}
\expandafter\def\csname res@R13@qasc\endcsname{86.6}
\expandafter\def\csname res@R13@snli\endcsname{87.7}
\expandafter\def\csname res@R13@squad2\endcsname{84.4}
\expandafter\def\csname res@R13@sst5\endcsname{56.2}
\expandafter\def\csname res@R13@sstep\endcsname{3.22}
\expandafter\def\csname res@R13@thead\endcsname{1.08}
\expandafter\def\csname res@R13@tydiqa\endcsname{91.1}
\expandafter\def\csname res@R13@typed_decisions\endcsname{76.6}
\expandafter\def\csname res@R16-R07@ag_news\endcsname{$-$3.1}
\expandafter\def\csname res@R16-R07@ag_newsp\endcsname{2e-22}
\expandafter\def\csname res@R16-R07@banking77\endcsname{+0.3}
\expandafter\def\csname res@R16-R07@banking77p\endcsname{0.5}
\expandafter\def\csname res@R16-R07@belebele_en\endcsname{+2.4}
\expandafter\def\csname res@R16-R07@belebele_enp\endcsname{0.1}
\expandafter\def\csname res@R16-R07@boolq\endcsname{+0.9}
\expandafter\def\csname res@R16-R07@boolqp\endcsname{0.2}
\expandafter\def\csname res@R16-R07@clinc150\endcsname{$-$0.5}
\expandafter\def\csname res@R16-R07@clinc150p\endcsname{0.09}
\expandafter\def\csname res@R16-R07@cosmosqa\endcsname{$-$1.3}
\expandafter\def\csname res@R16-R07@cosmosqap\endcsname{0.1}
\expandafter\def\csname res@R16-R07@dbpedia14\endcsname{$-$0.3}
\expandafter\def\csname res@R16-R07@dbpedia14p\endcsname{0.0008}
\expandafter\def\csname res@R16-R07@emotion\endcsname{$-$0.4}
\expandafter\def\csname res@R16-R07@emotionp\endcsname{0.4}
\expandafter\def\csname res@R16-R07@helpsteer2\endcsname{$-$0.2}
\expandafter\def\csname res@R16-R07@helpsteer2p\endcsname{0.7}
\expandafter\def\csname res@R16-R07@id\endcsname{$-$0.40}
\expandafter\def\csname res@R16-R07@idhi\endcsname{$-$0.15}
\expandafter\def\csname res@R16-R07@idlo\endcsname{$-$0.66}
\expandafter\def\csname res@R16-R07@kev_transfer\endcsname{+0.9}
\expandafter\def\csname res@R16-R07@kev_transferp\endcsname{0.6}
\expandafter\def\csname res@R16-R07@massive_en\endcsname{+0.2}
\expandafter\def\csname res@R16-R07@massive_enp\endcsname{0.4}
\expandafter\def\csname res@R16-R07@mnli\endcsname{$-$0.4}
\expandafter\def\csname res@R16-R07@mnlip\endcsname{0.2}
\expandafter\def\csname res@R16-R07@oasst1\endcsname{+0.0}
\expandafter\def\csname res@R16-R07@oasst1p\endcsname{0.9}
\expandafter\def\csname res@R16-R07@ood\endcsname{$-$0.37}
\expandafter\def\csname res@R16-R07@oodhi\endcsname{+0.51}
\expandafter\def\csname res@R16-R07@oodlo\endcsname{$-$1.28}
\expandafter\def\csname res@R16-R07@procedural_typed_decisions\endcsname{$-$3.6}
\expandafter\def\csname res@R16-R07@procedural_typed_decisionsp\endcsname{2e-132}
\expandafter\def\csname res@R16-R07@qasc\endcsname{$-$1.7}
\expandafter\def\csname res@R16-R07@qascp\endcsname{0.2}
\expandafter\def\csname res@R16-R07@snli\endcsname{$-$0.3}
\expandafter\def\csname res@R16-R07@snlip\endcsname{0.3}
\expandafter\def\csname res@R16-R07@squad2\endcsname{$-$1.1}
\expandafter\def\csname res@R16-R07@squad2p\endcsname{0.0008}
\expandafter\def\csname res@R16-R07@sst5\endcsname{$-$1.0}
\expandafter\def\csname res@R16-R07@sst5p\endcsname{0.3}
\expandafter\def\csname res@R16-R07@tydiqa\endcsname{+0.3}
\expandafter\def\csname res@R16-R07@tydiqap\endcsname{0.8}
\expandafter\def\csname res@R16-R07@typed_decisions\endcsname{+1.1}
\expandafter\def\csname res@R16-R07@typed_decisionsp\endcsname{0.1}
\expandafter\def\csname res@R16-R10@ag_news\endcsname{$-$2.4}
\expandafter\def\csname res@R16-R10@ag_newsp\endcsname{8e-15}
\expandafter\def\csname res@R16-R10@banking77\endcsname{$-$0.1}
\expandafter\def\csname res@R16-R10@banking77p\endcsname{0.8}
\expandafter\def\csname res@R16-R10@belebele_en\endcsname{$-$8.0}
\expandafter\def\csname res@R16-R10@belebele_enp\endcsname{2e-08}
\expandafter\def\csname res@R16-R10@boolq\endcsname{+0.3}
\expandafter\def\csname res@R16-R10@boolqp\endcsname{0.6}
\expandafter\def\csname res@R16-R10@clinc150\endcsname{+0.6}
\expandafter\def\csname res@R16-R10@clinc150p\endcsname{0.06}
\expandafter\def\csname res@R16-R10@cosmosqa\endcsname{$-$1.2}
\expandafter\def\csname res@R16-R10@cosmosqap\endcsname{0.1}
\expandafter\def\csname res@R16-R10@dbpedia14\endcsname{$-$1.2}
\expandafter\def\csname res@R16-R10@dbpedia14p\endcsname{1e-70}
\expandafter\def\csname res@R16-R10@emotion\endcsname{$-$0.7}
\expandafter\def\csname res@R16-R10@emotionp\endcsname{0.1}
\expandafter\def\csname res@R16-R10@helpsteer2\endcsname{$-$0.2}
\expandafter\def\csname res@R16-R10@helpsteer2p\endcsname{0.6}
\expandafter\def\csname res@R16-R10@id\endcsname{$-$0.30}
\expandafter\def\csname res@R16-R10@idhi\endcsname{$-$0.05}
\expandafter\def\csname res@R16-R10@idlo\endcsname{$-$0.52}
\expandafter\def\csname res@R16-R10@kev_transfer\endcsname{$-$1.5}
\expandafter\def\csname res@R16-R10@kev_transferp\endcsname{0.3}
\expandafter\def\csname res@R16-R10@massive_en\endcsname{+0.3}
\expandafter\def\csname res@R16-R10@massive_enp\endcsname{0.2}
\expandafter\def\csname res@R16-R10@mnli\endcsname{+0.3}
\expandafter\def\csname res@R16-R10@mnlip\endcsname{0.3}
\expandafter\def\csname res@R16-R10@oasst1\endcsname{+0.0}
\expandafter\def\csname res@R16-R10@oasst1p\endcsname{0.8}
\expandafter\def\csname res@R16-R10@ood\endcsname{$-$3.01}
\expandafter\def\csname res@R16-R10@oodhi\endcsname{$-$2.10}
\expandafter\def\csname res@R16-R10@oodlo\endcsname{$-$3.86}
\expandafter\def\csname res@R16-R10@procedural_typed_decisions\endcsname{$-$1.4}
\expandafter\def\csname res@R16-R10@procedural_typed_decisionsp\endcsname{4e-23}
\expandafter\def\csname res@R16-R10@qasc\endcsname{$-$1.8}
\expandafter\def\csname res@R16-R10@qascp\endcsname{0.1}
\expandafter\def\csname res@R16-R10@snli\endcsname{+0.5}
\expandafter\def\csname res@R16-R10@snlip\endcsname{0.07}
\expandafter\def\csname res@R16-R10@squad2\endcsname{$-$0.3}
\expandafter\def\csname res@R16-R10@squad2p\endcsname{0.4}
\expandafter\def\csname res@R16-R10@sst5\endcsname{$-$1.3}
\expandafter\def\csname res@R16-R10@sst5p\endcsname{0.1}
\expandafter\def\csname res@R16-R10@tydiqa\endcsname{$-$0.7}
\expandafter\def\csname res@R16-R10@tydiqap\endcsname{0.3}
\expandafter\def\csname res@R16-R10@typed_decisions\endcsname{$-$0.5}
\expandafter\def\csname res@R16-R10@typed_decisionsp\endcsname{0.4}
\expandafter\def\csname res@R16-R68@ag_news\endcsname{$-$4.7}
\expandafter\def\csname res@R16-R68@ag_newsp\endcsname{3e-41}
\expandafter\def\csname res@R16-R68@banking77\endcsname{+0.2}
\expandafter\def\csname res@R16-R68@banking77p\endcsname{0.6}
\expandafter\def\csname res@R16-R68@belebele_en\endcsname{+1.8}
\expandafter\def\csname res@R16-R68@belebele_enp\endcsname{0.3}
\expandafter\def\csname res@R16-R68@boolq\endcsname{$-$0.4}
\expandafter\def\csname res@R16-R68@boolqp\endcsname{0.6}
\expandafter\def\csname res@R16-R68@clinc150\endcsname{$-$0.8}
\expandafter\def\csname res@R16-R68@clinc150p\endcsname{0.009}
\expandafter\def\csname res@R16-R68@cosmosqa\endcsname{$-$1.3}
\expandafter\def\csname res@R16-R68@cosmosqap\endcsname{0.1}
\expandafter\def\csname res@R16-R68@dbpedia14\endcsname{$-$2.4}
\expandafter\def\csname res@R16-R68@dbpedia14p\endcsname{4e-212}
\expandafter\def\csname res@R16-R68@emotion\endcsname{+0.2}
\expandafter\def\csname res@R16-R68@emotionp\endcsname{0.7}
\expandafter\def\csname res@R16-R68@helpsteer2\endcsname{$-$0.6}
\expandafter\def\csname res@R16-R68@helpsteer2p\endcsname{0.1}
\expandafter\def\csname res@R16-R68@id\endcsname{$-$0.61}
\expandafter\def\csname res@R16-R68@idhi\endcsname{$-$0.35}
\expandafter\def\csname res@R16-R68@idlo\endcsname{$-$0.85}
\expandafter\def\csname res@R16-R68@kev_transfer\endcsname{+0.1}
\expandafter\def\csname res@R16-R68@kev_transferp\endcsname{1}
\expandafter\def\csname res@R16-R68@massive_en\endcsname{+0.2}
\expandafter\def\csname res@R16-R68@massive_enp\endcsname{0.4}
\expandafter\def\csname res@R16-R68@mnli\endcsname{$-$0.4}
\expandafter\def\csname res@R16-R68@mnlip\endcsname{0.2}
\expandafter\def\csname res@R16-R68@oasst1\endcsname{+0.0}
\expandafter\def\csname res@R16-R68@oasst1p\endcsname{0.8}
\expandafter\def\csname res@R16-R68@ood\endcsname{$-$1.60}
\expandafter\def\csname res@R16-R68@oodhi\endcsname{$-$0.71}
\expandafter\def\csname res@R16-R68@oodlo\endcsname{$-$2.49}
\expandafter\def\csname res@R16-R68@procedural_typed_decisions\endcsname{$-$4.0}
\expandafter\def\csname res@R16-R68@procedural_typed_decisionsp\endcsname{4e-151}
\expandafter\def\csname res@R16-R68@qasc\endcsname{$-$2.8}
\expandafter\def\csname res@R16-R68@qascp\endcsname{0.01}
\expandafter\def\csname res@R16-R68@snli\endcsname{+0.3}
\expandafter\def\csname res@R16-R68@snlip\endcsname{0.3}
\expandafter\def\csname res@R16-R68@squad2\endcsname{$-$0.8}
\expandafter\def\csname res@R16-R68@squad2p\endcsname{0.01}
\expandafter\def\csname res@R16-R68@sst5\endcsname{$-$1.1}
\expandafter\def\csname res@R16-R68@sst5p\endcsname{0.2}
\expandafter\def\csname res@R16-R68@tydiqa\endcsname{$-$0.4}
\expandafter\def\csname res@R16-R68@tydiqap\endcsname{0.7}
\expandafter\def\csname res@R16-R68@typed_decisions\endcsname{$-$0.3}
\expandafter\def\csname res@R16-R68@typed_decisionsp\endcsname{0.7}
\expandafter\def\csname res@R16@ag_news\endcsname{77.4}
\expandafter\def\csname res@R16@banking77\endcsname{93.1}
\expandafter\def\csname res@R16@belebele_en\endcsname{68.1}
\expandafter\def\csname res@R16@boolq\endcsname{84.4}
\expandafter\def\csname res@R16@clinc150\endcsname{91.9}
\expandafter\def\csname res@R16@cosmosqa\endcsname{72.9}
\expandafter\def\csname res@R16@dbpedia14\endcsname{94.6}
\expandafter\def\csname res@R16@emotion\endcsname{92.8}
\expandafter\def\csname res@R16@helpsteer2\endcsname{59.9}
\expandafter\def\csname res@R16@id\endcsname{81.3}
\expandafter\def\csname res@R16@idece\endcsname{0.033}
\expandafter\def\csname res@R16@kev_transfer\endcsname{56.2}
\expandafter\def\csname res@R16@massive_en\endcsname{91.3}
\expandafter\def\csname res@R16@mnli\endcsname{83.7}
\expandafter\def\csname res@R16@oasst1\endcsname{78.9}
\expandafter\def\csname res@R16@ood\endcsname{76.4}
\expandafter\def\csname res@R16@oodece\endcsname{0.057}
\expandafter\def\csname res@R16@peakgb\endcsname{6.3}
\expandafter\def\csname res@R16@procedural_typed_decisions\endcsname{77.9}
\expandafter\def\csname res@R16@qasc\endcsname{85.6}
\expandafter\def\csname res@R16@snli\endcsname{87.7}
\expandafter\def\csname res@R16@squad2\endcsname{83.6}
\expandafter\def\csname res@R16@sst5\endcsname{55.3}
\expandafter\def\csname res@R16@sstep\endcsname{3.18}
\expandafter\def\csname res@R16@thead\endcsname{1.05}
\expandafter\def\csname res@R16@tydiqa\endcsname{91.0}
\expandafter\def\csname res@R16@typed_decisions\endcsname{76.0}
\expandafter\def\csname res@R19-R07@ag_news\endcsname{$-$1.9}
\expandafter\def\csname res@R19-R07@ag_newsp\endcsname{1e-09}
\expandafter\def\csname res@R19-R07@banking77\endcsname{$-$0.1}
\expandafter\def\csname res@R19-R07@banking77p\endcsname{0.8}
\expandafter\def\csname res@R19-R07@belebele_en\endcsname{$-$0.8}
\expandafter\def\csname res@R19-R07@belebele_enp\endcsname{0.6}
\expandafter\def\csname res@R19-R07@boolq\endcsname{+0.6}
\expandafter\def\csname res@R19-R07@boolqp\endcsname{0.3}
\expandafter\def\csname res@R19-R07@clinc150\endcsname{$-$0.3}
\expandafter\def\csname res@R19-R07@clinc150p\endcsname{0.4}
\expandafter\def\csname res@R19-R07@cosmosqa\endcsname{$-$1.3}
\expandafter\def\csname res@R19-R07@cosmosqap\endcsname{0.1}
\expandafter\def\csname res@R19-R07@dbpedia14\endcsname{$-$0.0}
\expandafter\def\csname res@R19-R07@dbpedia14p\endcsname{0.6}
\expandafter\def\csname res@R19-R07@emotion\endcsname{$-$0.2}
\expandafter\def\csname res@R19-R07@emotionp\endcsname{0.7}
\expandafter\def\csname res@R19-R07@helpsteer2\endcsname{$-$1.3}
\expandafter\def\csname res@R19-R07@helpsteer2p\endcsname{0.01}
\expandafter\def\csname res@R19-R07@id\endcsname{$-$0.03}
\expandafter\def\csname res@R19-R07@idhi\endcsname{+0.23}
\expandafter\def\csname res@R19-R07@idlo\endcsname{$-$0.31}
\expandafter\def\csname res@R19-R07@kev_transfer\endcsname{+1.2}
\expandafter\def\csname res@R19-R07@kev_transferp\endcsname{0.3}
\expandafter\def\csname res@R19-R07@massive_en\endcsname{$-$0.5}
\expandafter\def\csname res@R19-R07@massive_enp\endcsname{0.06}
\expandafter\def\csname res@R19-R07@mnli\endcsname{$-$0.2}
\expandafter\def\csname res@R19-R07@mnlip\endcsname{0.5}
\expandafter\def\csname res@R19-R07@oasst1\endcsname{+0.0}
\expandafter\def\csname res@R19-R07@oasst1p\endcsname{0.9}
\expandafter\def\csname res@R19-R07@ood\endcsname{$-$0.38}
\expandafter\def\csname res@R19-R07@oodhi\endcsname{+0.42}
\expandafter\def\csname res@R19-R07@oodlo\endcsname{$-$1.19}
\expandafter\def\csname res@R19-R07@procedural_typed_decisions\endcsname{+2.0}
\expandafter\def\csname res@R19-R07@procedural_typed_decisionsp\endcsname{1e-40}
\expandafter\def\csname res@R19-R07@qasc\endcsname{$-$0.4}
\expandafter\def\csname res@R19-R07@qascp\endcsname{0.7}
\expandafter\def\csname res@R19-R07@snli\endcsname{$-$0.1}
\expandafter\def\csname res@R19-R07@snlip\endcsname{0.8}
\expandafter\def\csname res@R19-R07@squad2\endcsname{$-$0.4}
\expandafter\def\csname res@R19-R07@squad2p\endcsname{0.2}
\expandafter\def\csname res@R19-R07@sst5\endcsname{$-$1.2}
\expandafter\def\csname res@R19-R07@sst5p\endcsname{0.2}
\expandafter\def\csname res@R19-R07@tydiqa\endcsname{+0.6}
\expandafter\def\csname res@R19-R07@tydiqap\endcsname{0.5}
\expandafter\def\csname res@R19-R07@typed_decisions\endcsname{+1.8}
\expandafter\def\csname res@R19-R07@typed_decisionsp\endcsname{0.02}
\expandafter\def\csname res@R19@ag_news\endcsname{78.6}
\expandafter\def\csname res@R19@banking77\endcsname{92.7}
\expandafter\def\csname res@R19@belebele_en\endcsname{64.9}
\expandafter\def\csname res@R19@boolq\endcsname{84.1}
\expandafter\def\csname res@R19@clinc150\endcsname{92.1}
\expandafter\def\csname res@R19@cosmosqa\endcsname{72.8}
\expandafter\def\csname res@R19@dbpedia14\endcsname{94.8}
\expandafter\def\csname res@R19@emotion\endcsname{93.0}
\expandafter\def\csname res@R19@helpsteer2\endcsname{58.8}
\expandafter\def\csname res@R19@id\endcsname{81.7}
\expandafter\def\csname res@R19@idece\endcsname{0.036}
\expandafter\def\csname res@R19@kev_transfer\endcsname{56.6}
\expandafter\def\csname res@R19@massive_en\endcsname{90.6}
\expandafter\def\csname res@R19@mnli\endcsname{83.9}
\expandafter\def\csname res@R19@oasst1\endcsname{78.9}
\expandafter\def\csname res@R19@ood\endcsname{76.4}
\expandafter\def\csname res@R19@oodece\endcsname{0.075}
\expandafter\def\csname res@R19@peakgb\endcsname{6.3}
\expandafter\def\csname res@R19@procedural_typed_decisions\endcsname{83.5}
\expandafter\def\csname res@R19@qasc\endcsname{86.9}
\expandafter\def\csname res@R19@snli\endcsname{87.9}
\expandafter\def\csname res@R19@squad2\endcsname{84.3}
\expandafter\def\csname res@R19@sst5\endcsname{55.2}
\expandafter\def\csname res@R19@sstep\endcsname{7.74}
\expandafter\def\csname res@R19@thead\endcsname{1.11}
\expandafter\def\csname res@R19@tydiqa\endcsname{91.3}
\expandafter\def\csname res@R19@typed_decisions\endcsname{76.8}
\expandafter\def\csname res@R22@ag_news\endcsname{82.1}
\expandafter\def\csname res@R22@banking77\endcsname{93.4}
\expandafter\def\csname res@R22@belebele_en\endcsname{76.4}
\expandafter\def\csname res@R22@boolq\endcsname{88.2}
\expandafter\def\csname res@R22@clinc150\endcsname{93.8}
\expandafter\def\csname res@R22@cosmosqa\endcsname{82.1}
\expandafter\def\csname res@R22@dbpedia14\endcsname{97.9}
\expandafter\def\csname res@R22@emotion\endcsname{93.7}
\expandafter\def\csname res@R22@helpsteer2\endcsname{60.2}
\expandafter\def\csname res@R22@id\endcsname{84.3}
\expandafter\def\csname res@R22@idece\endcsname{0.031}
\expandafter\def\csname res@R22@kev_transfer\endcsname{65.3}
\expandafter\def\csname res@R22@massive_en\endcsname{91.9}
\expandafter\def\csname res@R22@mnli\endcsname{87.1}
\expandafter\def\csname res@R22@oasst1\endcsname{79.1}
\expandafter\def\csname res@R22@ood\endcsname{83.1}
\expandafter\def\csname res@R22@oodece\endcsname{0.051}
\expandafter\def\csname res@R22@peakgb\endcsname{12.0}
\expandafter\def\csname res@R22@procedural_typed_decisions\endcsname{86.9}
\expandafter\def\csname res@R22@qasc\endcsname{93.8}
\expandafter\def\csname res@R22@snli\endcsname{90.4}
\expandafter\def\csname res@R22@squad2\endcsname{88.4}
\expandafter\def\csname res@R22@sst5\endcsname{59.5}
\expandafter\def\csname res@R22@sstep\endcsname{3.77}
\expandafter\def\csname res@R22@thead\endcsname{1.09}
\expandafter\def\csname res@R22@tydiqa\endcsname{92.1}
\expandafter\def\csname res@R22@typed_decisions\endcsname{78.0}
\expandafter\def\csname res@R25-R22@ag_news\endcsname{$-$1.3}
\expandafter\def\csname res@R25-R22@ag_newsp\endcsname{2e-07}
\expandafter\def\csname res@R25-R22@banking77\endcsname{+0.2}
\expandafter\def\csname res@R25-R22@banking77p\endcsname{0.6}
\expandafter\def\csname res@R25-R22@belebele_en\endcsname{+0.9}
\expandafter\def\csname res@R25-R22@belebele_enp\endcsname{0.5}
\expandafter\def\csname res@R25-R22@boolq\endcsname{+1.0}
\expandafter\def\csname res@R25-R22@boolqp\endcsname{0.04}
\expandafter\def\csname res@R25-R22@clinc150\endcsname{$-$0.5}
\expandafter\def\csname res@R25-R22@clinc150p\endcsname{0.02}
\expandafter\def\csname res@R25-R22@cosmosqa\endcsname{$-$0.2}
\expandafter\def\csname res@R25-R22@cosmosqap\endcsname{0.8}
\expandafter\def\csname res@R25-R22@dbpedia14\endcsname{$-$0.1}
\expandafter\def\csname res@R25-R22@dbpedia14p\endcsname{0.004}
\expandafter\def\csname res@R25-R22@emotion\endcsname{$-$0.4}
\expandafter\def\csname res@R25-R22@emotionp\endcsname{0.5}
\expandafter\def\csname res@R25-R22@helpsteer2\endcsname{+0.4}
\expandafter\def\csname res@R25-R22@helpsteer2p\endcsname{0.3}
\expandafter\def\csname res@R25-R22@id\endcsname{+0.14}
\expandafter\def\csname res@R25-R22@idhi\endcsname{+0.36}
\expandafter\def\csname res@R25-R22@idlo\endcsname{$-$0.07}
\expandafter\def\csname res@R25-R22@kev_transfer\endcsname{$-$1.1}
\expandafter\def\csname res@R25-R22@kev_transferp\endcsname{0.3}
\expandafter\def\csname res@R25-R22@massive_en\endcsname{$-$0.4}
\expandafter\def\csname res@R25-R22@massive_enp\endcsname{0.06}
\expandafter\def\csname res@R25-R22@mnli\endcsname{+0.4}
\expandafter\def\csname res@R25-R22@mnlip\endcsname{0.09}
\expandafter\def\csname res@R25-R22@oasst1\endcsname{$-$0.0}
\expandafter\def\csname res@R25-R22@oasst1p\endcsname{0.9}
\expandafter\def\csname res@R25-R22@ood\endcsname{$-$0.80}
\expandafter\def\csname res@R25-R22@oodhi\endcsname{$-$0.15}
\expandafter\def\csname res@R25-R22@oodlo\endcsname{$-$1.42}
\expandafter\def\csname res@R25-R22@procedural_typed_decisions\endcsname{$-$0.5}
\expandafter\def\csname res@R25-R22@procedural_typed_decisionsp\endcsname{7e-05}
\expandafter\def\csname res@R25-R22@qasc\endcsname{$-$2.3}
\expandafter\def\csname res@R25-R22@qascp\endcsname{0.005}
\expandafter\def\csname res@R25-R22@snli\endcsname{$-$0.3}
\expandafter\def\csname res@R25-R22@snlip\endcsname{0.2}
\expandafter\def\csname res@R25-R22@squad2\endcsname{+0.2}
\expandafter\def\csname res@R25-R22@squad2p\endcsname{0.5}
\expandafter\def\csname res@R25-R22@sst5\endcsname{+1.1}
\expandafter\def\csname res@R25-R22@sst5p\endcsname{0.1}
\expandafter\def\csname res@R25-R22@tydiqa\endcsname{+0.5}
\expandafter\def\csname res@R25-R22@tydiqap\endcsname{0.5}
\expandafter\def\csname res@R25-R22@typed_decisions\endcsname{+0.5}
\expandafter\def\csname res@R25-R22@typed_decisionsp\endcsname{0.5}
\expandafter\def\csname res@R25@ag_news\endcsname{80.7}
\expandafter\def\csname res@R25@banking77\endcsname{93.6}
\expandafter\def\csname res@R25@belebele_en\endcsname{77.3}
\expandafter\def\csname res@R25@boolq\endcsname{89.2}
\expandafter\def\csname res@R25@clinc150\endcsname{93.3}
\expandafter\def\csname res@R25@cosmosqa\endcsname{81.9}
\expandafter\def\csname res@R25@dbpedia14\endcsname{97.7}
\expandafter\def\csname res@R25@emotion\endcsname{93.3}
\expandafter\def\csname res@R25@helpsteer2\endcsname{60.6}
\expandafter\def\csname res@R25@id\endcsname{84.5}
\expandafter\def\csname res@R25@idece\endcsname{0.032}
\expandafter\def\csname res@R25@kev_transfer\endcsname{64.1}
\expandafter\def\csname res@R25@massive_en\endcsname{91.5}
\expandafter\def\csname res@R25@mnli\endcsname{87.5}
\expandafter\def\csname res@R25@oasst1\endcsname{79.1}
\expandafter\def\csname res@R25@ood\endcsname{82.3}
\expandafter\def\csname res@R25@oodece\endcsname{0.056}
\expandafter\def\csname res@R25@peakgb\endcsname{12.5}
\expandafter\def\csname res@R25@procedural_typed_decisions\endcsname{86.5}
\expandafter\def\csname res@R25@qasc\endcsname{91.6}
\expandafter\def\csname res@R25@snli\endcsname{90.1}
\expandafter\def\csname res@R25@squad2\endcsname{88.6}
\expandafter\def\csname res@R25@sst5\endcsname{60.6}
\expandafter\def\csname res@R25@sstep\endcsname{4.67}
\expandafter\def\csname res@R25@thead\endcsname{1.11}
\expandafter\def\csname res@R25@tydiqa\endcsname{92.6}
\expandafter\def\csname res@R25@typed_decisions\endcsname{78.5}
\expandafter\def\csname res@R26@ag_news\endcsname{86.6}
\expandafter\def\csname res@R26@banking77\endcsname{93.8}
\expandafter\def\csname res@R26@belebele_en\endcsname{90.8}
\expandafter\def\csname res@R26@boolq\endcsname{90.6}
\expandafter\def\csname res@R26@clinc150\endcsname{94.4}
\expandafter\def\csname res@R26@cosmosqa\endcsname{87.4}
\expandafter\def\csname res@R26@dbpedia14\endcsname{98.5}
\expandafter\def\csname res@R26@emotion\endcsname{93.4}
\expandafter\def\csname res@R26@helpsteer2\endcsname{61.3}
\expandafter\def\csname res@R26@id\endcsname{85.9}
\expandafter\def\csname res@R26@idece\endcsname{0.030}
\expandafter\def\csname res@R26@kev_transfer\endcsname{72.5}
\expandafter\def\csname res@R26@massive_en\endcsname{92.3}
\expandafter\def\csname res@R26@mnli\endcsname{89.2}
\expandafter\def\csname res@R26@oasst1\endcsname{79.1}
\expandafter\def\csname res@R26@ood\endcsname{88.9}
\expandafter\def\csname res@R26@oodece\endcsname{0.026}
\expandafter\def\csname res@R26@peakgb\endcsname{21.0}
\expandafter\def\csname res@R26@procedural_typed_decisions\endcsname{90.5}
\expandafter\def\csname res@R26@qasc\endcsname{96.1}
\expandafter\def\csname res@R26@snli\endcsname{91.4}
\expandafter\def\csname res@R26@squad2\endcsname{91.0}
\expandafter\def\csname res@R26@sst5\endcsname{61.6}
\expandafter\def\csname res@R26@sstep\endcsname{5.53}
\expandafter\def\csname res@R26@thead\endcsname{1.09}
\expandafter\def\csname res@R26@tydiqa\endcsname{92.9}
\expandafter\def\csname res@R26@typed_decisions\endcsname{79.5}
\expandafter\def\csname res@R29b-R26@ag_news\endcsname{$-$0.6}
\expandafter\def\csname res@R29b-R26@ag_newsp\endcsname{0.005}
\expandafter\def\csname res@R29b-R26@banking77\endcsname{$-$0.3}
\expandafter\def\csname res@R29b-R26@banking77p\endcsname{0.4}
\expandafter\def\csname res@R29b-R26@belebele_en\endcsname{+0.1}
\expandafter\def\csname res@R29b-R26@belebele_enp\endcsname{1}
\expandafter\def\csname res@R29b-R26@boolq\endcsname{$-$0.2}
\expandafter\def\csname res@R29b-R26@boolqp\endcsname{0.6}
\expandafter\def\csname res@R29b-R26@clinc150\endcsname{$-$0.1}
\expandafter\def\csname res@R29b-R26@clinc150p\endcsname{0.8}
\expandafter\def\csname res@R29b-R26@cosmosqa\endcsname{+0.9}
\expandafter\def\csname res@R29b-R26@cosmosqap\endcsname{0.1}
\expandafter\def\csname res@R29b-R26@dbpedia14\endcsname{+0.0}
\expandafter\def\csname res@R29b-R26@dbpedia14p\endcsname{0.5}
\expandafter\def\csname res@R29b-R26@emotion\endcsname{$-$0.6}
\expandafter\def\csname res@R29b-R26@emotionp\endcsname{0.2}
\expandafter\def\csname res@R29b-R26@helpsteer2\endcsname{$-$0.1}
\expandafter\def\csname res@R29b-R26@helpsteer2p\endcsname{0.8}
\expandafter\def\csname res@R29b-R26@id\endcsname{+0.05}
\expandafter\def\csname res@R29b-R26@idhi\endcsname{+0.27}
\expandafter\def\csname res@R29b-R26@idlo\endcsname{$-$0.15}
\expandafter\def\csname res@R29b-R26@kev_transfer\endcsname{+2.8}
\expandafter\def\csname res@R29b-R26@kev_transferp\endcsname{0.01}
\expandafter\def\csname res@R29b-R26@massive_en\endcsname{$-$0.4}
\expandafter\def\csname res@R29b-R26@massive_enp\endcsname{0.05}
\expandafter\def\csname res@R29b-R26@mnli\endcsname{+0.1}
\expandafter\def\csname res@R29b-R26@mnlip\endcsname{0.8}
\expandafter\def\csname res@R29b-R26@oasst1\endcsname{+0.3}
\expandafter\def\csname res@R29b-R26@oasst1p\endcsname{0.03}
\expandafter\def\csname res@R29b-R26@ood\endcsname{+0.43}
\expandafter\def\csname res@R29b-R26@oodhi\endcsname{+1.00}
\expandafter\def\csname res@R29b-R26@oodlo\endcsname{$-$0.17}
\expandafter\def\csname res@R29b-R26@procedural_typed_decisions\endcsname{$-$0.3}
\expandafter\def\csname res@R29b-R26@procedural_typed_decisionsp\endcsname{0.004}
\expandafter\def\csname res@R29b-R26@qasc\endcsname{$-$0.1}
\expandafter\def\csname res@R29b-R26@qascp\endcsname{1}
\expandafter\def\csname res@R29b-R26@snli\endcsname{+0.3}
\expandafter\def\csname res@R29b-R26@snlip\endcsname{0.09}
\expandafter\def\csname res@R29b-R26@squad2\endcsname{+0.1}
\expandafter\def\csname res@R29b-R26@squad2p\endcsname{0.6}
\expandafter\def\csname res@R29b-R26@sst5\endcsname{+0.5}
\expandafter\def\csname res@R29b-R26@sst5p\endcsname{0.6}
\expandafter\def\csname res@R29b-R26@tydiqa\endcsname{+0.0}
\expandafter\def\csname res@R29b-R26@tydiqap\endcsname{1}
\expandafter\def\csname res@R29b-R26@typed_decisions\endcsname{+0.7}
\expandafter\def\csname res@R29b-R26@typed_decisionsp\endcsname{0.4}
\expandafter\def\csname res@R29b@ag_news\endcsname{86.0}
\expandafter\def\csname res@R29b@banking77\endcsname{93.5}
\expandafter\def\csname res@R29b@belebele_en\endcsname{90.9}
\expandafter\def\csname res@R29b@boolq\endcsname{90.4}
\expandafter\def\csname res@R29b@clinc150\endcsname{94.3}
\expandafter\def\csname res@R29b@cosmosqa\endcsname{88.3}
\expandafter\def\csname res@R29b@dbpedia14\endcsname{98.5}
\expandafter\def\csname res@R29b@emotion\endcsname{92.8}
\expandafter\def\csname res@R29b@helpsteer2\endcsname{61.1}
\expandafter\def\csname res@R29b@id\endcsname{85.9}
\expandafter\def\csname res@R29b@idece\endcsname{0.031}
\expandafter\def\csname res@R29b@kev_transfer\endcsname{75.2}
\expandafter\def\csname res@R29b@massive_en\endcsname{91.9}
\expandafter\def\csname res@R29b@mnli\endcsname{89.3}
\expandafter\def\csname res@R29b@oasst1\endcsname{79.4}
\expandafter\def\csname res@R29b@ood\endcsname{89.3}
\expandafter\def\csname res@R29b@oodece\endcsname{0.030}
\expandafter\def\csname res@R29b@peakgb\endcsname{21.5}
\expandafter\def\csname res@R29b@procedural_typed_decisions\endcsname{90.1}
\expandafter\def\csname res@R29b@qasc\endcsname{96.0}
\expandafter\def\csname res@R29b@snli\endcsname{91.7}
\expandafter\def\csname res@R29b@squad2\endcsname{91.2}
\expandafter\def\csname res@R29b@sst5\endcsname{62.1}
\expandafter\def\csname res@R29b@sstep\endcsname{6.80}
\expandafter\def\csname res@R29b@thead\endcsname{1.08}
\expandafter\def\csname res@R29b@tydiqa\endcsname{92.9}
\expandafter\def\csname res@R29b@typed_decisions\endcsname{80.2}
\expandafter\def\csname res@R30b@ag_news\endcsname{86.8}
\expandafter\def\csname res@R30b@banking77\endcsname{92.8}
\expandafter\def\csname res@R30b@belebele_en\endcsname{88.6}
\expandafter\def\csname res@R30b@boolq\endcsname{91.3}
\expandafter\def\csname res@R30b@clinc150\endcsname{94.1}
\expandafter\def\csname res@R30b@cosmosqa\endcsname{88.7}
\expandafter\def\csname res@R30b@dbpedia14\endcsname{98.6}
\expandafter\def\csname res@R30b@emotion\endcsname{92.5}
\expandafter\def\csname res@R30b@helpsteer2\endcsname{61.4}
\expandafter\def\csname res@R30b@id\endcsname{85.8}
\expandafter\def\csname res@R30b@idece\endcsname{0.028}
\expandafter\def\csname res@R30b@kev_transfer\endcsname{78.1}
\expandafter\def\csname res@R30b@massive_en\endcsname{91.6}
\expandafter\def\csname res@R30b@mnli\endcsname{90.3}
\expandafter\def\csname res@R30b@oasst1\endcsname{79.1}
\expandafter\def\csname res@R30b@ood\endcsname{89.8}
\expandafter\def\csname res@R30b@oodece\endcsname{0.029}
\expandafter\def\csname res@R30b@peakgb\endcsname{34.1}
\expandafter\def\csname res@R30b@procedural_typed_decisions\endcsname{87.8}
\expandafter\def\csname res@R30b@qasc\endcsname{97.0}
\expandafter\def\csname res@R30b@snli\endcsname{92.0}
\expandafter\def\csname res@R30b@squad2\endcsname{91.8}
\expandafter\def\csname res@R30b@sst5\endcsname{61.8}
\expandafter\def\csname res@R30b@sstep\endcsname{6.59}
\expandafter\def\csname res@R30b@thead\endcsname{1.03}
\expandafter\def\csname res@R30b@tydiqa\endcsname{93.7}
\expandafter\def\csname res@R30b@typed_decisions\endcsname{78.3}
\expandafter\def\csname res@R31-R30b@ag_news\endcsname{$-$0.9}
\expandafter\def\csname res@R31-R30b@ag_newsp\endcsname{5e-09}
\expandafter\def\csname res@R31-R30b@banking77\endcsname{+0.2}
\expandafter\def\csname res@R31-R30b@banking77p\endcsname{0.6}
\expandafter\def\csname res@R31-R30b@belebele_en\endcsname{+0.7}
\expandafter\def\csname res@R31-R30b@belebele_enp\endcsname{0.5}
\expandafter\def\csname res@R31-R30b@boolq\endcsname{$-$0.4}
\expandafter\def\csname res@R31-R30b@boolqp\endcsname{0.3}
\expandafter\def\csname res@R31-R30b@clinc150\endcsname{+0.1}
\expandafter\def\csname res@R31-R30b@clinc150p\endcsname{0.9}
\expandafter\def\csname res@R31-R30b@cosmosqa\endcsname{+0.1}
\expandafter\def\csname res@R31-R30b@cosmosqap\endcsname{0.9}
\expandafter\def\csname res@R31-R30b@dbpedia14\endcsname{+0.0}
\expandafter\def\csname res@R31-R30b@dbpedia14p\endcsname{0.6}
\expandafter\def\csname res@R31-R30b@emotion\endcsname{+0.1}
\expandafter\def\csname res@R31-R30b@emotionp\endcsname{0.8}
\expandafter\def\csname res@R31-R30b@helpsteer2\endcsname{$-$0.8}
\expandafter\def\csname res@R31-R30b@helpsteer2p\endcsname{0.01}
\expandafter\def\csname res@R31-R30b@id\endcsname{+0.07}
\expandafter\def\csname res@R31-R30b@idhi\endcsname{+0.25}
\expandafter\def\csname res@R31-R30b@idlo\endcsname{$-$0.11}
\expandafter\def\csname res@R31-R30b@kev_transfer\endcsname{$-$3.4}
\expandafter\def\csname res@R31-R30b@kev_transferp\endcsname{0.0008}
\expandafter\def\csname res@R31-R30b@massive_en\endcsname{+0.5}
\expandafter\def\csname res@R31-R30b@massive_enp\endcsname{0.01}
\expandafter\def\csname res@R31-R30b@mnli\endcsname{+0.0}
\expandafter\def\csname res@R31-R30b@mnlip\endcsname{1}
\expandafter\def\csname res@R31-R30b@oasst1\endcsname{$-$0.1}
\expandafter\def\csname res@R31-R30b@oasst1p\endcsname{0.7}
\expandafter\def\csname res@R31-R30b@ood\endcsname{$-$0.80}
\expandafter\def\csname res@R31-R30b@oodhi\endcsname{$-$0.26}
\expandafter\def\csname res@R31-R30b@oodlo\endcsname{$-$1.33}
\expandafter\def\csname res@R31-R30b@procedural_typed_decisions\endcsname{+0.5}
\expandafter\def\csname res@R31-R30b@procedural_typed_decisionsp\endcsname{2e-05}
\expandafter\def\csname res@R31-R30b@qasc\endcsname{$-$0.3}
\expandafter\def\csname res@R31-R30b@qascp\endcsname{0.6}
\expandafter\def\csname res@R31-R30b@snli\endcsname{$-$0.2}
\expandafter\def\csname res@R31-R30b@snlip\endcsname{0.3}
\expandafter\def\csname res@R31-R30b@squad2\endcsname{+0.2}
\expandafter\def\csname res@R31-R30b@squad2p\endcsname{0.2}
\expandafter\def\csname res@R31-R30b@sst5\endcsname{+0.6}
\expandafter\def\csname res@R31-R30b@sst5p\endcsname{0.4}
\expandafter\def\csname res@R31-R30b@tydiqa\endcsname{$-$0.3}
\expandafter\def\csname res@R31-R30b@tydiqap\endcsname{0.6}
\expandafter\def\csname res@R31-R30b@typed_decisions\endcsname{+0.6}
\expandafter\def\csname res@R31-R30b@typed_decisionsp\endcsname{0.4}
\expandafter\def\csname res@R31@ag_news\endcsname{85.9}
\expandafter\def\csname res@R31@banking77\endcsname{92.9}
\expandafter\def\csname res@R31@belebele_en\endcsname{89.2}
\expandafter\def\csname res@R31@boolq\endcsname{90.9}
\expandafter\def\csname res@R31@clinc150\endcsname{94.2}
\expandafter\def\csname res@R31@cosmosqa\endcsname{88.7}
\expandafter\def\csname res@R31@dbpedia14\endcsname{98.6}
\expandafter\def\csname res@R31@emotion\endcsname{92.7}
\expandafter\def\csname res@R31@helpsteer2\endcsname{60.6}
\expandafter\def\csname res@R31@id\endcsname{85.9}
\expandafter\def\csname res@R31@idece\endcsname{0.024}
\expandafter\def\csname res@R31@kev_transfer\endcsname{74.7}
\expandafter\def\csname res@R31@massive_en\endcsname{92.2}
\expandafter\def\csname res@R31@mnli\endcsname{90.4}
\expandafter\def\csname res@R31@oasst1\endcsname{79.1}
\expandafter\def\csname res@R31@ood\endcsname{89.0}
\expandafter\def\csname res@R31@oodece\endcsname{0.030}
\expandafter\def\csname res@R31@peakgb\endcsname{35.9}
\expandafter\def\csname res@R31@procedural_typed_decisions\endcsname{88.3}
\expandafter\def\csname res@R31@qasc\endcsname{96.7}
\expandafter\def\csname res@R31@snli\endcsname{91.8}
\expandafter\def\csname res@R31@squad2\endcsname{92.1}
\expandafter\def\csname res@R31@sst5\endcsname{62.4}
\expandafter\def\csname res@R31@sstep\endcsname{7.82}
\expandafter\def\csname res@R31@thead\endcsname{1.02}
\expandafter\def\csname res@R31@tydiqa\endcsname{93.4}
\expandafter\def\csname res@R31@typed_decisions\endcsname{78.9}
\expandafter\def\csname res@R32@ag_news\endcsname{87.0}
\expandafter\def\csname res@R32@banking77\endcsname{93.5}
\expandafter\def\csname res@R32@belebele_en\endcsname{90.8}
\expandafter\def\csname res@R32@boolq\endcsname{92.5}
\expandafter\def\csname res@R32@clinc150\endcsname{95.4}
\expandafter\def\csname res@R32@cosmosqa\endcsname{91.1}
\expandafter\def\csname res@R32@dbpedia14\endcsname{98.7}
\expandafter\def\csname res@R32@emotion\endcsname{92.9}
\expandafter\def\csname res@R32@helpsteer2\endcsname{61.3}
\expandafter\def\csname res@R32@id\endcsname{86.6}
\expandafter\def\csname res@R32@idece\endcsname{0.025}
\expandafter\def\csname res@R32@kev_transfer\endcsname{81.2}
\expandafter\def\csname res@R32@massive_en\endcsname{91.6}
\expandafter\def\csname res@R32@mnli\endcsname{90.1}
\expandafter\def\csname res@R32@oasst1\endcsname{79.3}
\expandafter\def\csname res@R32@ood\endcsname{90.8}
\expandafter\def\csname res@R32@oodece\endcsname{0.023}
\expandafter\def\csname res@R32@peakgb\endcsname{53.0}
\expandafter\def\csname res@R32@procedural_typed_decisions\endcsname{90.3}
\expandafter\def\csname res@R32@qasc\endcsname{96.4}
\expandafter\def\csname res@R32@snli\endcsname{92.1}
\expandafter\def\csname res@R32@squad2\endcsname{92.4}
\expandafter\def\csname res@R32@sst5\endcsname{62.6}
\expandafter\def\csname res@R32@sstep\endcsname{9.09}
\expandafter\def\csname res@R32@thead\endcsname{1.00}
\expandafter\def\csname res@R32@tydiqa\endcsname{93.7}
\expandafter\def\csname res@R32@typed_decisions\endcsname{79.7}
\expandafter\def\csname res@R33b-R32@ag_news\endcsname{+0.3}
\expandafter\def\csname res@R33b-R32@ag_newsp\endcsname{0.06}
\expandafter\def\csname res@R33b-R32@banking77\endcsname{$-$0.3}
\expandafter\def\csname res@R33b-R32@banking77p\endcsname{0.4}
\expandafter\def\csname res@R33b-R32@belebele_en\endcsname{+2.1}
\expandafter\def\csname res@R33b-R32@belebele_enp\endcsname{0.003}
\expandafter\def\csname res@R33b-R32@boolq\endcsname{$-$0.1}
\expandafter\def\csname res@R33b-R32@boolqp\endcsname{0.8}
\expandafter\def\csname res@R33b-R32@clinc150\endcsname{$-$0.7}
\expandafter\def\csname res@R33b-R32@clinc150p\endcsname{0.0005}
\expandafter\def\csname res@R33b-R32@cosmosqa\endcsname{$-$0.5}
\expandafter\def\csname res@R33b-R32@cosmosqap\endcsname{0.4}
\expandafter\def\csname res@R33b-R32@dbpedia14\endcsname{$-$0.0}
\expandafter\def\csname res@R33b-R32@dbpedia14p\endcsname{0.1}
\expandafter\def\csname res@R33b-R32@emotion\endcsname{$-$0.1}
\expandafter\def\csname res@R33b-R32@emotionp\endcsname{0.9}
\expandafter\def\csname res@R33b-R32@helpsteer2\endcsname{$-$0.2}
\expandafter\def\csname res@R33b-R32@helpsteer2p\endcsname{0.5}
\expandafter\def\csname res@R33b-R32@id\endcsname{$-$0.17}
\expandafter\def\csname res@R33b-R32@idhi\endcsname{$-$0.00}
\expandafter\def\csname res@R33b-R32@idlo\endcsname{$-$0.34}
\expandafter\def\csname res@R33b-R32@kev_transfer\endcsname{$-$1.0}
\expandafter\def\csname res@R33b-R32@kev_transferp\endcsname{0.3}
\expandafter\def\csname res@R33b-R32@massive_en\endcsname{+0.0}
\expandafter\def\csname res@R33b-R32@massive_enp\endcsname{0.9}
\expandafter\def\csname res@R33b-R32@mnli\endcsname{+0.0}
\expandafter\def\csname res@R33b-R32@mnlip\endcsname{1}
\expandafter\def\csname res@R33b-R32@oasst1\endcsname{$-$0.0}
\expandafter\def\csname res@R33b-R32@oasst1p\endcsname{0.8}
\expandafter\def\csname res@R33b-R32@ood\endcsname{+0.59}
\expandafter\def\csname res@R33b-R32@oodhi\endcsname{+1.05}
\expandafter\def\csname res@R33b-R32@oodlo\endcsname{+0.13}
\expandafter\def\csname res@R33b-R32@procedural_typed_decisions\endcsname{$-$0.4}
\expandafter\def\csname res@R33b-R32@procedural_typed_decisionsp\endcsname{0.0008}
\expandafter\def\csname res@R33b-R32@qasc\endcsname{+1.5}
\expandafter\def\csname res@R33b-R32@qascp\endcsname{0.0005}
\expandafter\def\csname res@R33b-R32@snli\endcsname{+0.3}
\expandafter\def\csname res@R33b-R32@snlip\endcsname{0.1}
\expandafter\def\csname res@R33b-R32@squad2\endcsname{+0.0}
\expandafter\def\csname res@R33b-R32@squad2p\endcsname{1}
\expandafter\def\csname res@R33b-R32@sst5\endcsname{+0.4}
\expandafter\def\csname res@R33b-R32@sst5p\endcsname{0.6}
\expandafter\def\csname res@R33b-R32@tydiqa\endcsname{$-$0.2}
\expandafter\def\csname res@R33b-R32@tydiqap\endcsname{0.8}
\expandafter\def\csname res@R33b-R32@typed_decisions\endcsname{$-$0.8}
\expandafter\def\csname res@R33b-R32@typed_decisionsp\endcsname{0.2}
\expandafter\def\csname res@R33b@ag_news\endcsname{87.4}
\expandafter\def\csname res@R33b@banking77\endcsname{93.2}
\expandafter\def\csname res@R33b@belebele_en\endcsname{92.9}
\expandafter\def\csname res@R33b@boolq\endcsname{92.4}
\expandafter\def\csname res@R33b@clinc150\endcsname{94.8}
\expandafter\def\csname res@R33b@cosmosqa\endcsname{90.6}
\expandafter\def\csname res@R33b@dbpedia14\endcsname{98.6}
\expandafter\def\csname res@R33b@emotion\endcsname{92.8}
\expandafter\def\csname res@R33b@helpsteer2\endcsname{61.1}
\expandafter\def\csname res@R33b@id\endcsname{86.4}
\expandafter\def\csname res@R33b@idece\endcsname{0.024}
\expandafter\def\csname res@R33b@kev_transfer\endcsname{80.2}
\expandafter\def\csname res@R33b@massive_en\endcsname{91.6}
\expandafter\def\csname res@R33b@mnli\endcsname{90.1}
\expandafter\def\csname res@R33b@oasst1\endcsname{79.3}
\expandafter\def\csname res@R33b@ood\endcsname{91.4}
\expandafter\def\csname res@R33b@oodece\endcsname{0.024}
\expandafter\def\csname res@R33b@peakgb\endcsname{54.8}
\expandafter\def\csname res@R33b@procedural_typed_decisions\endcsname{89.9}
\expandafter\def\csname res@R33b@qasc\endcsname{97.9}
\expandafter\def\csname res@R33b@snli\endcsname{92.3}
\expandafter\def\csname res@R33b@squad2\endcsname{92.4}
\expandafter\def\csname res@R33b@sst5\endcsname{63.0}
\expandafter\def\csname res@R33b@sstep\endcsname{10.35}
\expandafter\def\csname res@R33b@thead\endcsname{1.00}
\expandafter\def\csname res@R33b@tydiqa\endcsname{93.5}
\expandafter\def\csname res@R33b@typed_decisions\endcsname{78.8}
\expandafter\def\csname res@R38-R26@ag_news\endcsname{+1.1}
\expandafter\def\csname res@R38-R26@ag_newsp\endcsname{6e-07}
\expandafter\def\csname res@R38-R26@banking77\endcsname{+0.0}
\expandafter\def\csname res@R38-R26@banking77p\endcsname{1}
\expandafter\def\csname res@R38-R26@belebele_en\endcsname{+0.8}
\expandafter\def\csname res@R38-R26@belebele_enp\endcsname{0.4}
\expandafter\def\csname res@R38-R26@boolq\endcsname{+0.1}
\expandafter\def\csname res@R38-R26@boolqp\endcsname{0.9}
\expandafter\def\csname res@R38-R26@clinc150\endcsname{+0.3}
\expandafter\def\csname res@R38-R26@clinc150p\endcsname{0.2}
\expandafter\def\csname res@R38-R26@cosmosqa\endcsname{+0.3}
\expandafter\def\csname res@R38-R26@cosmosqap\endcsname{0.6}
\expandafter\def\csname res@R38-R26@dbpedia14\endcsname{$-$0.1}
\expandafter\def\csname res@R38-R26@dbpedia14p\endcsname{0.0002}
\expandafter\def\csname res@R38-R26@emotion\endcsname{$-$0.1}
\expandafter\def\csname res@R38-R26@emotionp\endcsname{1}
\expandafter\def\csname res@R38-R26@helpsteer2\endcsname{+0.0}
\expandafter\def\csname res@R38-R26@helpsteer2p\endcsname{1}
\expandafter\def\csname res@R38-R26@id\endcsname{+0.06}
\expandafter\def\csname res@R38-R26@idhi\endcsname{+0.27}
\expandafter\def\csname res@R38-R26@idlo\endcsname{$-$0.16}
\expandafter\def\csname res@R38-R26@kev_transfer\endcsname{+0.9}
\expandafter\def\csname res@R38-R26@kev_transferp\endcsname{0.4}
\expandafter\def\csname res@R38-R26@massive_en\endcsname{$-$0.4}
\expandafter\def\csname res@R38-R26@massive_enp\endcsname{0.08}
\expandafter\def\csname res@R38-R26@mnli\endcsname{$-$0.2}
\expandafter\def\csname res@R38-R26@mnlip\endcsname{0.4}
\expandafter\def\csname res@R38-R26@oasst1\endcsname{+0.1}
\expandafter\def\csname res@R38-R26@oasst1p\endcsname{0.7}
\expandafter\def\csname res@R38-R26@ood\endcsname{+0.53}
\expandafter\def\csname res@R38-R26@oodhi\endcsname{+1.06}
\expandafter\def\csname res@R38-R26@oodlo\endcsname{$-$0.03}
\expandafter\def\csname res@R38-R26@procedural_typed_decisions\endcsname{+0.2}
\expandafter\def\csname res@R38-R26@procedural_typed_decisionsp\endcsname{0.2}
\expandafter\def\csname res@R38-R26@qasc\endcsname{+0.0}
\expandafter\def\csname res@R38-R26@qascp\endcsname{1}
\expandafter\def\csname res@R38-R26@snli\endcsname{+0.1}
\expandafter\def\csname res@R38-R26@snlip\endcsname{0.5}
\expandafter\def\csname res@R38-R26@squad2\endcsname{+0.3}
\expandafter\def\csname res@R38-R26@squad2p\endcsname{0.2}
\expandafter\def\csname res@R38-R26@sst5\endcsname{$-$0.1}
\expandafter\def\csname res@R38-R26@sst5p\endcsname{0.9}
\expandafter\def\csname res@R38-R26@tydiqa\endcsname{$-$0.1}
\expandafter\def\csname res@R38-R26@tydiqap\endcsname{1}
\expandafter\def\csname res@R38-R26@typed_decisions\endcsname{+0.2}
\expandafter\def\csname res@R38-R26@typed_decisionsp\endcsname{0.8}
\expandafter\def\csname res@R38@ag_news\endcsname{87.8}
\expandafter\def\csname res@R38@banking77\endcsname{93.8}
\expandafter\def\csname res@R38@belebele_en\endcsname{91.6}
\expandafter\def\csname res@R38@boolq\endcsname{90.7}
\expandafter\def\csname res@R38@clinc150\endcsname{94.7}
\expandafter\def\csname res@R38@cosmosqa\endcsname{87.7}
\expandafter\def\csname res@R38@dbpedia14\endcsname{98.3}
\expandafter\def\csname res@R38@emotion\endcsname{93.3}
\expandafter\def\csname res@R38@helpsteer2\endcsname{61.3}
\expandafter\def\csname res@R38@id\endcsname{85.9}
\expandafter\def\csname res@R38@idece\endcsname{0.031}
\expandafter\def\csname res@R38@kev_transfer\endcsname{73.3}
\expandafter\def\csname res@R38@massive_en\endcsname{91.9}
\expandafter\def\csname res@R38@mnli\endcsname{89.0}
\expandafter\def\csname res@R38@oasst1\endcsname{79.1}
\expandafter\def\csname res@R38@ood\endcsname{89.4}
\expandafter\def\csname res@R38@oodece\endcsname{0.029}
\expandafter\def\csname res@R38@peakgb\endcsname{21.0}
\expandafter\def\csname res@R38@procedural_typed_decisions\endcsname{90.6}
\expandafter\def\csname res@R38@qasc\endcsname{96.1}
\expandafter\def\csname res@R38@snli\endcsname{91.5}
\expandafter\def\csname res@R38@squad2\endcsname{91.4}
\expandafter\def\csname res@R38@sst5\endcsname{61.5}
\expandafter\def\csname res@R38@sstep\endcsname{5.51}
\expandafter\def\csname res@R38@thead\endcsname{1.10}
\expandafter\def\csname res@R38@tydiqa\endcsname{92.8}
\expandafter\def\csname res@R38@typed_decisions\endcsname{79.8}
\expandafter\def\csname res@R41-R26@ag_news\endcsname{+0.9}
\expandafter\def\csname res@R41-R26@ag_newsp\endcsname{0.0002}
\expandafter\def\csname res@R41-R26@banking77\endcsname{+0.4}
\expandafter\def\csname res@R41-R26@banking77p\endcsname{0.2}
\expandafter\def\csname res@R41-R26@belebele_en\endcsname{+0.7}
\expandafter\def\csname res@R41-R26@belebele_enp\endcsname{0.5}
\expandafter\def\csname res@R41-R26@boolq\endcsname{+0.0}
\expandafter\def\csname res@R41-R26@boolqp\endcsname{1}
\expandafter\def\csname res@R41-R26@clinc150\endcsname{+0.5}
\expandafter\def\csname res@R41-R26@clinc150p\endcsname{0.02}
\expandafter\def\csname res@R41-R26@cosmosqa\endcsname{$-$0.3}
\expandafter\def\csname res@R41-R26@cosmosqap\endcsname{0.7}
\expandafter\def\csname res@R41-R26@dbpedia14\endcsname{$-$0.0}
\expandafter\def\csname res@R41-R26@dbpedia14p\endcsname{0.6}
\expandafter\def\csname res@R41-R26@emotion\endcsname{$-$0.4}
\expandafter\def\csname res@R41-R26@emotionp\endcsname{0.4}
\expandafter\def\csname res@R41-R26@helpsteer2\endcsname{$-$0.6}
\expandafter\def\csname res@R41-R26@helpsteer2p\endcsname{0.1}
\expandafter\def\csname res@R41-R26@id\endcsname{+0.05}
\expandafter\def\csname res@R41-R26@idhi\endcsname{+0.27}
\expandafter\def\csname res@R41-R26@idlo\endcsname{$-$0.17}
\expandafter\def\csname res@R41-R26@kev_transfer\endcsname{+0.1}
\expandafter\def\csname res@R41-R26@kev_transferp\endcsname{1}
\expandafter\def\csname res@R41-R26@massive_en\endcsname{$-$0.2}
\expandafter\def\csname res@R41-R26@massive_enp\endcsname{0.5}
\expandafter\def\csname res@R41-R26@mnli\endcsname{+0.2}
\expandafter\def\csname res@R41-R26@mnlip\endcsname{0.5}
\expandafter\def\csname res@R41-R26@oasst1\endcsname{+0.0}
\expandafter\def\csname res@R41-R26@oasst1p\endcsname{0.9}
\expandafter\def\csname res@R41-R26@ood\endcsname{+0.09}
\expandafter\def\csname res@R41-R26@oodhi\endcsname{+0.69}
\expandafter\def\csname res@R41-R26@oodlo\endcsname{$-$0.61}
\expandafter\def\csname res@R41-R26@procedural_typed_decisions\endcsname{+0.3}
\expandafter\def\csname res@R41-R26@procedural_typed_decisionsp\endcsname{0.004}
\expandafter\def\csname res@R41-R26@qasc\endcsname{$-$1.2}
\expandafter\def\csname res@R41-R26@qascp\endcsname{0.1}
\expandafter\def\csname res@R41-R26@snli\endcsname{+0.3}
\expandafter\def\csname res@R41-R26@snlip\endcsname{0.1}
\expandafter\def\csname res@R41-R26@squad2\endcsname{+0.1}
\expandafter\def\csname res@R41-R26@squad2p\endcsname{0.8}
\expandafter\def\csname res@R41-R26@sst5\endcsname{+0.3}
\expandafter\def\csname res@R41-R26@sst5p\endcsname{0.7}
\expandafter\def\csname res@R41-R26@tydiqa\endcsname{$-$0.2}
\expandafter\def\csname res@R41-R26@tydiqap\endcsname{0.9}
\expandafter\def\csname res@R41-R26@typed_decisions\endcsname{+0.2}
\expandafter\def\csname res@R41-R26@typed_decisionsp\endcsname{0.8}
\expandafter\def\csname res@R41-R38@ag_news\endcsname{$-$0.3}
\expandafter\def\csname res@R41-R38@ag_newsp\endcsname{0.2}
\expandafter\def\csname res@R41-R38@banking77\endcsname{+0.4}
\expandafter\def\csname res@R41-R38@banking77p\endcsname{0.3}
\expandafter\def\csname res@R41-R38@belebele_en\endcsname{$-$0.1}
\expandafter\def\csname res@R41-R38@belebele_enp\endcsname{1}
\expandafter\def\csname res@R41-R38@boolq\endcsname{$-$0.1}
\expandafter\def\csname res@R41-R38@boolqp\endcsname{0.9}
\expandafter\def\csname res@R41-R38@clinc150\endcsname{+0.2}
\expandafter\def\csname res@R41-R38@clinc150p\endcsname{0.3}
\expandafter\def\csname res@R41-R38@cosmosqa\endcsname{$-$0.6}
\expandafter\def\csname res@R41-R38@cosmosqap\endcsname{0.4}
\expandafter\def\csname res@R41-R38@dbpedia14\endcsname{+0.1}
\expandafter\def\csname res@R41-R38@dbpedia14p\endcsname{0.002}
\expandafter\def\csname res@R41-R38@emotion\endcsname{$-$0.4}
\expandafter\def\csname res@R41-R38@emotionp\endcsname{0.4}
\expandafter\def\csname res@R41-R38@helpsteer2\endcsname{$-$0.6}
\expandafter\def\csname res@R41-R38@helpsteer2p\endcsname{0.1}
\expandafter\def\csname res@R41-R38@id\endcsname{$-$0.01}
\expandafter\def\csname res@R41-R38@idhi\endcsname{+0.19}
\expandafter\def\csname res@R41-R38@idlo\endcsname{$-$0.21}
\expandafter\def\csname res@R41-R38@kev_transfer\endcsname{$-$0.8}
\expandafter\def\csname res@R41-R38@kev_transferp\endcsname{0.6}
\expandafter\def\csname res@R41-R38@massive_en\endcsname{+0.2}
\expandafter\def\csname res@R41-R38@massive_enp\endcsname{0.3}
\expandafter\def\csname res@R41-R38@mnli\endcsname{+0.4}
\expandafter\def\csname res@R41-R38@mnlip\endcsname{0.08}
\expandafter\def\csname res@R41-R38@oasst1\endcsname{$-$0.0}
\expandafter\def\csname res@R41-R38@oasst1p\endcsname{0.9}
\expandafter\def\csname res@R41-R38@ood\endcsname{$-$0.45}
\expandafter\def\csname res@R41-R38@oodhi\endcsname{+0.15}
\expandafter\def\csname res@R41-R38@oodlo\endcsname{$-$1.07}
\expandafter\def\csname res@R41-R38@procedural_typed_decisions\endcsname{+0.2}
\expandafter\def\csname res@R41-R38@procedural_typed_decisionsp\endcsname{0.1}
\expandafter\def\csname res@R41-R38@qasc\endcsname{$-$1.2}
\expandafter\def\csname res@R41-R38@qascp\endcsname{0.05}
\expandafter\def\csname res@R41-R38@snli\endcsname{+0.2}
\expandafter\def\csname res@R41-R38@snlip\endcsname{0.4}
\expandafter\def\csname res@R41-R38@squad2\endcsname{$-$0.3}
\expandafter\def\csname res@R41-R38@squad2p\endcsname{0.2}
\expandafter\def\csname res@R41-R38@sst5\endcsname{+0.5}
\expandafter\def\csname res@R41-R38@sst5p\endcsname{0.6}
\expandafter\def\csname res@R41-R38@tydiqa\endcsname{$-$0.1}
\expandafter\def\csname res@R41-R38@tydiqap\endcsname{1}
\expandafter\def\csname res@R41-R38@typed_decisions\endcsname{+0.0}
\expandafter\def\csname res@R41-R38@typed_decisionsp\endcsname{1}
\expandafter\def\csname res@R41@ag_news\endcsname{87.5}
\expandafter\def\csname res@R41@banking77\endcsname{94.2}
\expandafter\def\csname res@R41@belebele_en\endcsname{91.4}
\expandafter\def\csname res@R41@boolq\endcsname{90.7}
\expandafter\def\csname res@R41@clinc150\endcsname{94.9}
\expandafter\def\csname res@R41@cosmosqa\endcsname{87.1}
\expandafter\def\csname res@R41@dbpedia14\endcsname{98.4}
\expandafter\def\csname res@R41@emotion\endcsname{93.0}
\expandafter\def\csname res@R41@helpsteer2\endcsname{60.6}
\expandafter\def\csname res@R41@id\endcsname{85.9}
\expandafter\def\csname res@R41@idece\endcsname{0.030}
\expandafter\def\csname res@R41@kev_transfer\endcsname{72.6}
\expandafter\def\csname res@R41@massive_en\endcsname{92.1}
\expandafter\def\csname res@R41@mnli\endcsname{89.4}
\expandafter\def\csname res@R41@oasst1\endcsname{79.1}
\expandafter\def\csname res@R41@ood\endcsname{89.0}
\expandafter\def\csname res@R41@oodece\endcsname{0.028}
\expandafter\def\csname res@R41@peakgb\endcsname{21.0}
\expandafter\def\csname res@R41@procedural_typed_decisions\endcsname{90.8}
\expandafter\def\csname res@R41@qasc\endcsname{94.9}
\expandafter\def\csname res@R41@snli\endcsname{91.7}
\expandafter\def\csname res@R41@squad2\endcsname{91.1}
\expandafter\def\csname res@R41@sst5\endcsname{61.9}
\expandafter\def\csname res@R41@sstep\endcsname{5.51}
\expandafter\def\csname res@R41@thead\endcsname{1.08}
\expandafter\def\csname res@R41@tydiqa\endcsname{92.6}
\expandafter\def\csname res@R41@typed_decisions\endcsname{79.8}
\expandafter\def\csname res@R46-R07@ag_news\endcsname{+0.3}
\expandafter\def\csname res@R46-R07@ag_newsp\endcsname{0.4}
\expandafter\def\csname res@R46-R07@banking77\endcsname{+0.3}
\expandafter\def\csname res@R46-R07@banking77p\endcsname{0.4}
\expandafter\def\csname res@R46-R07@belebele_en\endcsname{$-$0.8}
\expandafter\def\csname res@R46-R07@belebele_enp\endcsname{0.6}
\expandafter\def\csname res@R46-R07@boolq\endcsname{+0.5}
\expandafter\def\csname res@R46-R07@boolqp\endcsname{0.5}
\expandafter\def\csname res@R46-R07@clinc150\endcsname{+0.3}
\expandafter\def\csname res@R46-R07@clinc150p\endcsname{0.2}
\expandafter\def\csname res@R46-R07@cosmosqa\endcsname{+1.5}
\expandafter\def\csname res@R46-R07@cosmosqap\endcsname{0.06}
\expandafter\def\csname res@R46-R07@dbpedia14\endcsname{+2.5}
\expandafter\def\csname res@R46-R07@dbpedia14p\endcsname{7e-259}
\expandafter\def\csname res@R46-R07@emotion\endcsname{+0.1}
\expandafter\def\csname res@R46-R07@emotionp\endcsname{0.9}
\expandafter\def\csname res@R46-R07@helpsteer2\endcsname{$-$0.5}
\expandafter\def\csname res@R46-R07@helpsteer2p\endcsname{0.2}
\expandafter\def\csname res@R46-R07@id\endcsname{$-$0.02}
\expandafter\def\csname res@R46-R07@idhi\endcsname{+0.24}
\expandafter\def\csname res@R46-R07@idlo\endcsname{$-$0.28}
\expandafter\def\csname res@R46-R07@kev_transfer\endcsname{$-$0.6}
\expandafter\def\csname res@R46-R07@kev_transferp\endcsname{0.7}
\expandafter\def\csname res@R46-R07@massive_en\endcsname{$-$0.1}
\expandafter\def\csname res@R46-R07@massive_enp\endcsname{0.6}
\expandafter\def\csname res@R46-R07@mnli\endcsname{$-$0.3}
\expandafter\def\csname res@R46-R07@mnlip\endcsname{0.3}
\expandafter\def\csname res@R46-R07@oasst1\endcsname{$-$0.0}
\expandafter\def\csname res@R46-R07@oasst1p\endcsname{0.8}
\expandafter\def\csname res@R46-R07@ood\endcsname{+0.93}
\expandafter\def\csname res@R46-R07@oodhi\endcsname{+1.82}
\expandafter\def\csname res@R46-R07@oodlo\endcsname{+0.09}
\expandafter\def\csname res@R46-R07@procedural_typed_decisions\endcsname{$-$2.9}
\expandafter\def\csname res@R46-R07@procedural_typed_decisionsp\endcsname{2e-82}
\expandafter\def\csname res@R46-R07@qasc\endcsname{+3.2}
\expandafter\def\csname res@R46-R07@qascp\endcsname{0.002}
\expandafter\def\csname res@R46-R07@snli\endcsname{+0.0}
\expandafter\def\csname res@R46-R07@snlip\endcsname{0.9}
\expandafter\def\csname res@R46-R07@squad2\endcsname{$-$0.2}
\expandafter\def\csname res@R46-R07@squad2p\endcsname{0.6}
\expandafter\def\csname res@R46-R07@sst5\endcsname{+0.0}
\expandafter\def\csname res@R46-R07@sst5p\endcsname{1}
\expandafter\def\csname res@R46-R07@tydiqa\endcsname{+0.9}
\expandafter\def\csname res@R46-R07@tydiqap\endcsname{0.3}
\expandafter\def\csname res@R46-R07@typed_decisions\endcsname{+0.1}
\expandafter\def\csname res@R46-R07@typed_decisionsp\endcsname{1}
\expandafter\def\csname res@R46-R68@ag_news\endcsname{$-$1.3}
\expandafter\def\csname res@R46-R68@ag_newsp\endcsname{2e-05}
\expandafter\def\csname res@R46-R68@banking77\endcsname{+0.3}
\expandafter\def\csname res@R46-R68@banking77p\endcsname{0.4}
\expandafter\def\csname res@R46-R68@belebele_en\endcsname{$-$1.4}
\expandafter\def\csname res@R46-R68@belebele_enp\endcsname{0.4}
\expandafter\def\csname res@R46-R68@boolq\endcsname{$-$0.8}
\expandafter\def\csname res@R46-R68@boolqp\endcsname{0.2}
\expandafter\def\csname res@R46-R68@clinc150\endcsname{+0.1}
\expandafter\def\csname res@R46-R68@clinc150p\endcsname{0.9}
\expandafter\def\csname res@R46-R68@cosmosqa\endcsname{+1.5}
\expandafter\def\csname res@R46-R68@cosmosqap\endcsname{0.08}
\expandafter\def\csname res@R46-R68@dbpedia14\endcsname{+0.5}
\expandafter\def\csname res@R46-R68@dbpedia14p\endcsname{2e-16}
\expandafter\def\csname res@R46-R68@emotion\endcsname{+0.8}
\expandafter\def\csname res@R46-R68@emotionp\endcsname{0.09}
\expandafter\def\csname res@R46-R68@helpsteer2\endcsname{$-$0.9}
\expandafter\def\csname res@R46-R68@helpsteer2p\endcsname{0.02}
\expandafter\def\csname res@R46-R68@id\endcsname{$-$0.23}
\expandafter\def\csname res@R46-R68@idhi\endcsname{+0.04}
\expandafter\def\csname res@R46-R68@idlo\endcsname{$-$0.47}
\expandafter\def\csname res@R46-R68@kev_transfer\endcsname{$-$1.3}
\expandafter\def\csname res@R46-R68@kev_transferp\endcsname{0.3}
\expandafter\def\csname res@R46-R68@massive_en\endcsname{$-$0.1}
\expandafter\def\csname res@R46-R68@massive_enp\endcsname{0.7}
\expandafter\def\csname res@R46-R68@mnli\endcsname{$-$0.3}
\expandafter\def\csname res@R46-R68@mnlip\endcsname{0.3}
\expandafter\def\csname res@R46-R68@oasst1\endcsname{$-$0.0}
\expandafter\def\csname res@R46-R68@oasst1p\endcsname{0.9}
\expandafter\def\csname res@R46-R68@ood\endcsname{$-$0.29}
\expandafter\def\csname res@R46-R68@oodhi\endcsname{+0.62}
\expandafter\def\csname res@R46-R68@oodlo\endcsname{$-$1.15}
\expandafter\def\csname res@R46-R68@procedural_typed_decisions\endcsname{$-$3.3}
\expandafter\def\csname res@R46-R68@procedural_typed_decisionsp\endcsname{4e-100}
\expandafter\def\csname res@R46-R68@qasc\endcsname{+2.2}
\expandafter\def\csname res@R46-R68@qascp\endcsname{0.04}
\expandafter\def\csname res@R46-R68@snli\endcsname{+0.6}
\expandafter\def\csname res@R46-R68@snlip\endcsname{0.03}
\expandafter\def\csname res@R46-R68@squad2\endcsname{+0.1}
\expandafter\def\csname res@R46-R68@squad2p\endcsname{0.7}
\expandafter\def\csname res@R46-R68@sst5\endcsname{$-$0.0}
\expandafter\def\csname res@R46-R68@sst5p\endcsname{1}
\expandafter\def\csname res@R46-R68@tydiqa\endcsname{+0.2}
\expandafter\def\csname res@R46-R68@tydiqap\endcsname{0.9}
\expandafter\def\csname res@R46-R68@typed_decisions\endcsname{$-$1.4}
\expandafter\def\csname res@R46-R68@typed_decisionsp\endcsname{0.07}
\expandafter\def\csname res@R46@ag_news\endcsname{80.8}
\expandafter\def\csname res@R46@banking77\endcsname{93.1}
\expandafter\def\csname res@R46@belebele_en\endcsname{64.9}
\expandafter\def\csname res@R46@boolq\endcsname{84.0}
\expandafter\def\csname res@R46@clinc150\endcsname{92.7}
\expandafter\def\csname res@R46@cosmosqa\endcsname{75.7}
\expandafter\def\csname res@R46@dbpedia14\endcsname{97.4}
\expandafter\def\csname res@R46@emotion\endcsname{93.3}
\expandafter\def\csname res@R46@helpsteer2\endcsname{59.5}
\expandafter\def\csname res@R46@id\endcsname{81.7}
\expandafter\def\csname res@R46@idece\endcsname{0.034}
\expandafter\def\csname res@R46@kev_transfer\endcsname{54.9}
\expandafter\def\csname res@R46@massive_en\endcsname{91.0}
\expandafter\def\csname res@R46@mnli\endcsname{83.7}
\expandafter\def\csname res@R46@oasst1\endcsname{78.8}
\expandafter\def\csname res@R46@ood\endcsname{77.7}
\expandafter\def\csname res@R46@oodece\endcsname{0.073}
\expandafter\def\csname res@R46@peakgb\endcsname{6.3}
\expandafter\def\csname res@R46@procedural_typed_decisions\endcsname{78.6}
\expandafter\def\csname res@R46@qasc\endcsname{90.6}
\expandafter\def\csname res@R46@snli\endcsname{88.0}
\expandafter\def\csname res@R46@squad2\endcsname{84.6}
\expandafter\def\csname res@R46@sst5\endcsname{56.3}
\expandafter\def\csname res@R46@sstep\endcsname{3.22}
\expandafter\def\csname res@R46@thead\endcsname{1.06}
\expandafter\def\csname res@R46@tydiqa\endcsname{91.6}
\expandafter\def\csname res@R46@typed_decisions\endcsname{75.0}
\expandafter\def\csname res@R49-R22@ag_news\endcsname{+0.2}
\expandafter\def\csname res@R49-R22@ag_newsp\endcsname{0.5}
\expandafter\def\csname res@R49-R22@banking77\endcsname{$-$0.2}
\expandafter\def\csname res@R49-R22@banking77p\endcsname{0.7}
\expandafter\def\csname res@R49-R22@belebele_en\endcsname{+0.2}
\expandafter\def\csname res@R49-R22@belebele_enp\endcsname{0.9}
\expandafter\def\csname res@R49-R22@boolq\endcsname{$-$0.4}
\expandafter\def\csname res@R49-R22@boolqp\endcsname{0.5}
\expandafter\def\csname res@R49-R22@clinc150\endcsname{+0.3}
\expandafter\def\csname res@R49-R22@clinc150p\endcsname{0.2}
\expandafter\def\csname res@R49-R22@cosmosqa\endcsname{+0.7}
\expandafter\def\csname res@R49-R22@cosmosqap\endcsname{0.3}
\expandafter\def\csname res@R49-R22@dbpedia14\endcsname{+0.1}
\expandafter\def\csname res@R49-R22@dbpedia14p\endcsname{0.003}
\expandafter\def\csname res@R49-R22@emotion\endcsname{$-$0.3}
\expandafter\def\csname res@R49-R22@emotionp\endcsname{0.5}
\expandafter\def\csname res@R49-R22@helpsteer2\endcsname{$-$0.3}
\expandafter\def\csname res@R49-R22@helpsteer2p\endcsname{0.4}
\expandafter\def\csname res@R49-R22@id\endcsname{$-$0.03}
\expandafter\def\csname res@R49-R22@idhi\endcsname{+0.19}
\expandafter\def\csname res@R49-R22@idlo\endcsname{$-$0.26}
\expandafter\def\csname res@R49-R22@kev_transfer\endcsname{$-$2.3}
\expandafter\def\csname res@R49-R22@kev_transferp\endcsname{0.05}
\expandafter\def\csname res@R49-R22@massive_en\endcsname{$-$0.2}
\expandafter\def\csname res@R49-R22@massive_enp\endcsname{0.5}
\expandafter\def\csname res@R49-R22@mnli\endcsname{$-$0.2}
\expandafter\def\csname res@R49-R22@mnlip\endcsname{0.5}
\expandafter\def\csname res@R49-R22@oasst1\endcsname{$-$0.0}
\expandafter\def\csname res@R49-R22@oasst1p\endcsname{1}
\expandafter\def\csname res@R49-R22@ood\endcsname{$-$0.36}
\expandafter\def\csname res@R49-R22@oodhi\endcsname{+0.34}
\expandafter\def\csname res@R49-R22@oodlo\endcsname{$-$1.02}
\expandafter\def\csname res@R49-R22@procedural_typed_decisions\endcsname{$-$1.7}
\expandafter\def\csname res@R49-R22@procedural_typed_decisionsp\endcsname{2e-41}
\expandafter\def\csname res@R49-R22@qasc\endcsname{$-$0.1}
\expandafter\def\csname res@R49-R22@qascp\endcsname{1}
\expandafter\def\csname res@R49-R22@snli\endcsname{$-$0.0}
\expandafter\def\csname res@R49-R22@snlip\endcsname{1}
\expandafter\def\csname res@R49-R22@squad2\endcsname{$-$0.7}
\expandafter\def\csname res@R49-R22@squad2p\endcsname{0.009}
\expandafter\def\csname res@R49-R22@sst5\endcsname{+1.1}
\expandafter\def\csname res@R49-R22@sst5p\endcsname{0.2}
\expandafter\def\csname res@R49-R22@tydiqa\endcsname{+0.4}
\expandafter\def\csname res@R49-R22@tydiqap\endcsname{0.7}
\expandafter\def\csname res@R49-R22@typed_decisions\endcsname{+0.9}
\expandafter\def\csname res@R49-R22@typed_decisionsp\endcsname{0.2}
\expandafter\def\csname res@R49@ag_news\endcsname{82.3}
\expandafter\def\csname res@R49@banking77\endcsname{93.2}
\expandafter\def\csname res@R49@belebele_en\endcsname{76.7}
\expandafter\def\csname res@R49@boolq\endcsname{87.8}
\expandafter\def\csname res@R49@clinc150\endcsname{94.1}
\expandafter\def\csname res@R49@cosmosqa\endcsname{82.8}
\expandafter\def\csname res@R49@dbpedia14\endcsname{98.0}
\expandafter\def\csname res@R49@emotion\endcsname{93.4}
\expandafter\def\csname res@R49@helpsteer2\endcsname{59.9}
\expandafter\def\csname res@R49@id\endcsname{84.3}
\expandafter\def\csname res@R49@idece\endcsname{0.030}
\expandafter\def\csname res@R49@kev_transfer\endcsname{63.0}
\expandafter\def\csname res@R49@massive_en\endcsname{91.7}
\expandafter\def\csname res@R49@mnli\endcsname{86.9}
\expandafter\def\csname res@R49@oasst1\endcsname{79.1}
\expandafter\def\csname res@R49@ood\endcsname{82.7}
\expandafter\def\csname res@R49@oodece\endcsname{0.050}
\expandafter\def\csname res@R49@peakgb\endcsname{12.0}
\expandafter\def\csname res@R49@procedural_typed_decisions\endcsname{85.2}
\expandafter\def\csname res@R49@qasc\endcsname{93.7}
\expandafter\def\csname res@R49@snli\endcsname{90.3}
\expandafter\def\csname res@R49@squad2\endcsname{87.7}
\expandafter\def\csname res@R49@sst5\endcsname{60.6}
\expandafter\def\csname res@R49@sstep\endcsname{3.65}
\expandafter\def\csname res@R49@thead\endcsname{1.09}
\expandafter\def\csname res@R49@tydiqa\endcsname{92.5}
\expandafter\def\csname res@R49@typed_decisions\endcsname{78.9}
\expandafter\def\csname res@R50@ag_news\endcsname{33.4}
\expandafter\def\csname res@R50@banking77\endcsname{85.7}
\expandafter\def\csname res@R50@belebele_en\endcsname{23.8}
\expandafter\def\csname res@R50@boolq\endcsname{64.5}
\expandafter\def\csname res@R50@clinc150\endcsname{78.0}
\expandafter\def\csname res@R50@cosmosqa\endcsname{42.2}
\expandafter\def\csname res@R50@dbpedia14\endcsname{31.5}
\expandafter\def\csname res@R50@emotion\endcsname{91.8}
\expandafter\def\csname res@R50@helpsteer2\endcsname{57.5}
\expandafter\def\csname res@R50@id\endcsname{63.5}
\expandafter\def\csname res@R50@idece\endcsname{0.036}
\expandafter\def\csname res@R50@kev_transfer\endcsname{45.3}
\expandafter\def\csname res@R50@massive_en\endcsname{84.3}
\expandafter\def\csname res@R50@mnli\endcsname{48.0}
\expandafter\def\csname res@R50@oasst1\endcsname{77.8}
\expandafter\def\csname res@R50@ood\endcsname{36.1}
\expandafter\def\csname res@R50@oodece\endcsname{0.118}
\expandafter\def\csname res@R50@peakgb\endcsname{19.7}
\expandafter\def\csname res@R50@procedural_typed_decisions\endcsname{46.5}
\expandafter\def\csname res@R50@qasc\endcsname{46.7}
\expandafter\def\csname res@R50@snli\endcsname{49.7}
\expandafter\def\csname res@R50@squad2\endcsname{51.9}
\expandafter\def\csname res@R50@sst5\endcsname{43.7}
\expandafter\def\csname res@R50@sstep\endcsname{5.52}
\expandafter\def\csname res@R50@thead\endcsname{1.07}
\expandafter\def\csname res@R50@tydiqa\endcsname{76.4}
\expandafter\def\csname res@R50@typed_decisions\endcsname{54.8}
\expandafter\def\csname res@R50b-R26@ag_news\endcsname{$-$0.5}
\expandafter\def\csname res@R50b-R26@ag_newsp\endcsname{0.07}
\expandafter\def\csname res@R50b-R26@banking77\endcsname{$-$0.1}
\expandafter\def\csname res@R50b-R26@banking77p\endcsname{0.7}
\expandafter\def\csname res@R50b-R26@belebele_en\endcsname{$-$0.6}
\expandafter\def\csname res@R50b-R26@belebele_enp\endcsname{0.6}
\expandafter\def\csname res@R50b-R26@boolq\endcsname{$-$0.5}
\expandafter\def\csname res@R50b-R26@boolqp\endcsname{0.3}
\expandafter\def\csname res@R50b-R26@clinc150\endcsname{+0.3}
\expandafter\def\csname res@R50b-R26@clinc150p\endcsname{0.3}
\expandafter\def\csname res@R50b-R26@cosmosqa\endcsname{+1.0}
\expandafter\def\csname res@R50b-R26@cosmosqap\endcsname{0.1}
\expandafter\def\csname res@R50b-R26@dbpedia14\endcsname{$-$0.1}
\expandafter\def\csname res@R50b-R26@dbpedia14p\endcsname{0.04}
\expandafter\def\csname res@R50b-R26@emotion\endcsname{$-$1.0}
\expandafter\def\csname res@R50b-R26@emotionp\endcsname{0.03}
\expandafter\def\csname res@R50b-R26@helpsteer2\endcsname{$-$0.1}
\expandafter\def\csname res@R50b-R26@helpsteer2p\endcsname{0.9}
\expandafter\def\csname res@R50b-R26@id\endcsname{+0.05}
\expandafter\def\csname res@R50b-R26@idhi\endcsname{+0.27}
\expandafter\def\csname res@R50b-R26@idlo\endcsname{$-$0.17}
\expandafter\def\csname res@R50b-R26@kev_transfer\endcsname{+2.3}
\expandafter\def\csname res@R50b-R26@kev_transferp\endcsname{0.05}
\expandafter\def\csname res@R50b-R26@massive_en\endcsname{$-$0.5}
\expandafter\def\csname res@R50b-R26@massive_enp\endcsname{0.02}
\expandafter\def\csname res@R50b-R26@mnli\endcsname{+0.2}
\expandafter\def\csname res@R50b-R26@mnlip\endcsname{0.5}
\expandafter\def\csname res@R50b-R26@oasst1\endcsname{+0.2}
\expandafter\def\csname res@R50b-R26@oasst1p\endcsname{0.2}
\expandafter\def\csname res@R50b-R26@ood\endcsname{+0.35}
\expandafter\def\csname res@R50b-R26@oodhi\endcsname{+0.94}
\expandafter\def\csname res@R50b-R26@oodlo\endcsname{$-$0.23}
\expandafter\def\csname res@R50b-R26@procedural_typed_decisions\endcsname{$-$1.1}
\expandafter\def\csname res@R50b-R26@procedural_typed_decisionsp\endcsname{2e-24}
\expandafter\def\csname res@R50b-R26@qasc\endcsname{+0.5}
\expandafter\def\csname res@R50b-R26@qascp\endcsname{0.4}
\expandafter\def\csname res@R50b-R26@snli\endcsname{+0.8}
\expandafter\def\csname res@R50b-R26@snlip\endcsname{3e-05}
\expandafter\def\csname res@R50b-R26@squad2\endcsname{+0.3}
\expandafter\def\csname res@R50b-R26@squad2p\endcsname{0.3}
\expandafter\def\csname res@R50b-R26@sst5\endcsname{+1.9}
\expandafter\def\csname res@R50b-R26@sst5p\endcsname{0.02}
\expandafter\def\csname res@R50b-R26@tydiqa\endcsname{+0.2}
\expandafter\def\csname res@R50b-R26@tydiqap\endcsname{0.9}
\expandafter\def\csname res@R50b-R26@typed_decisions\endcsname{$-$0.6}
\expandafter\def\csname res@R50b-R26@typed_decisionsp\endcsname{0.4}
\expandafter\def\csname res@R50b@ag_news\endcsname{86.1}
\expandafter\def\csname res@R50b@banking77\endcsname{93.7}
\expandafter\def\csname res@R50b@belebele_en\endcsname{90.2}
\expandafter\def\csname res@R50b@boolq\endcsname{90.1}
\expandafter\def\csname res@R50b@clinc150\endcsname{94.7}
\expandafter\def\csname res@R50b@cosmosqa\endcsname{88.4}
\expandafter\def\csname res@R50b@dbpedia14\endcsname{98.4}
\expandafter\def\csname res@R50b@emotion\endcsname{92.4}
\expandafter\def\csname res@R50b@helpsteer2\endcsname{61.2}
\expandafter\def\csname res@R50b@id\endcsname{85.9}
\expandafter\def\csname res@R50b@idece\endcsname{0.028}
\expandafter\def\csname res@R50b@kev_transfer\endcsname{74.8}
\expandafter\def\csname res@R50b@massive_en\endcsname{91.8}
\expandafter\def\csname res@R50b@mnli\endcsname{89.4}
\expandafter\def\csname res@R50b@oasst1\endcsname{79.3}
\expandafter\def\csname res@R50b@ood\endcsname{89.2}
\expandafter\def\csname res@R50b@oodece\endcsname{0.029}
\expandafter\def\csname res@R50b@peakgb\endcsname{21.0}
\expandafter\def\csname res@R50b@procedural_typed_decisions\endcsname{89.3}
\expandafter\def\csname res@R50b@qasc\endcsname{96.7}
\expandafter\def\csname res@R50b@snli\endcsname{92.2}
\expandafter\def\csname res@R50b@squad2\endcsname{91.3}
\expandafter\def\csname res@R50b@sst5\endcsname{63.5}
\expandafter\def\csname res@R50b@sstep\endcsname{5.43}
\expandafter\def\csname res@R50b@thead\endcsname{1.08}
\expandafter\def\csname res@R50b@tydiqa\endcsname{93.1}
\expandafter\def\csname res@R50b@typed_decisions\endcsname{78.9}
\expandafter\def\csname res@R53-R30b@ag_news\endcsname{$-$0.3}
\expandafter\def\csname res@R53-R30b@ag_newsp\endcsname{0.05}
\expandafter\def\csname res@R53-R30b@banking77\endcsname{+0.5}
\expandafter\def\csname res@R53-R30b@banking77p\endcsname{0.07}
\expandafter\def\csname res@R53-R30b@belebele_en\endcsname{$-$2.7}
\expandafter\def\csname res@R53-R30b@belebele_enp\endcsname{0.004}
\expandafter\def\csname res@R53-R30b@boolq\endcsname{+0.1}
\expandafter\def\csname res@R53-R30b@boolqp\endcsname{0.9}
\expandafter\def\csname res@R53-R30b@clinc150\endcsname{+0.3}
\expandafter\def\csname res@R53-R30b@clinc150p\endcsname{0.3}
\expandafter\def\csname res@R53-R30b@cosmosqa\endcsname{+1.5}
\expandafter\def\csname res@R53-R30b@cosmosqap\endcsname{0.009}
\expandafter\def\csname res@R53-R30b@dbpedia14\endcsname{$-$0.1}
\expandafter\def\csname res@R53-R30b@dbpedia14p\endcsname{0.06}
\expandafter\def\csname res@R53-R30b@emotion\endcsname{$-$0.2}
\expandafter\def\csname res@R53-R30b@emotionp\endcsname{0.6}
\expandafter\def\csname res@R53-R30b@helpsteer2\endcsname{$-$0.8}
\expandafter\def\csname res@R53-R30b@helpsteer2p\endcsname{0.03}
\expandafter\def\csname res@R53-R30b@id\endcsname{+0.09}
\expandafter\def\csname res@R53-R30b@idhi\endcsname{+0.28}
\expandafter\def\csname res@R53-R30b@idlo\endcsname{$-$0.12}
\expandafter\def\csname res@R53-R30b@kev_transfer\endcsname{$-$3.8}
\expandafter\def\csname res@R53-R30b@kev_transferp\endcsname{0.0003}
\expandafter\def\csname res@R53-R30b@massive_en\endcsname{+0.2}
\expandafter\def\csname res@R53-R30b@massive_enp\endcsname{0.3}
\expandafter\def\csname res@R53-R30b@mnli\endcsname{+0.0}
\expandafter\def\csname res@R53-R30b@mnlip\endcsname{0.9}
\expandafter\def\csname res@R53-R30b@oasst1\endcsname{$-$0.0}
\expandafter\def\csname res@R53-R30b@oasst1p\endcsname{0.9}
\expandafter\def\csname res@R53-R30b@ood\endcsname{$-$1.33}
\expandafter\def\csname res@R53-R30b@oodhi\endcsname{$-$0.81}
\expandafter\def\csname res@R53-R30b@oodlo\endcsname{$-$1.88}
\expandafter\def\csname res@R53-R30b@procedural_typed_decisions\endcsname{$-$0.6}
\expandafter\def\csname res@R53-R30b@procedural_typed_decisionsp\endcsname{4e-06}
\expandafter\def\csname res@R53-R30b@qasc\endcsname{+0.2}
\expandafter\def\csname res@R53-R30b@qascp\endcsname{0.8}
\expandafter\def\csname res@R53-R30b@snli\endcsname{$-$0.1}
\expandafter\def\csname res@R53-R30b@snlip\endcsname{0.8}
\expandafter\def\csname res@R53-R30b@squad2\endcsname{$-$0.0}
\expandafter\def\csname res@R53-R30b@squad2p\endcsname{1}
\expandafter\def\csname res@R53-R30b@sst5\endcsname{+1.0}
\expandafter\def\csname res@R53-R30b@sst5p\endcsname{0.1}
\expandafter\def\csname res@R53-R30b@tydiqa\endcsname{$-$0.5}
\expandafter\def\csname res@R53-R30b@tydiqap\endcsname{0.4}
\expandafter\def\csname res@R53-R30b@typed_decisions\endcsname{+0.0}
\expandafter\def\csname res@R53-R30b@typed_decisionsp\endcsname{1}
\expandafter\def\csname res@R53@ag_news\endcsname{86.5}
\expandafter\def\csname res@R53@banking77\endcsname{93.3}
\expandafter\def\csname res@R53@belebele_en\endcsname{85.9}
\expandafter\def\csname res@R53@boolq\endcsname{91.4}
\expandafter\def\csname res@R53@clinc150\endcsname{94.4}
\expandafter\def\csname res@R53@cosmosqa\endcsname{90.1}
\expandafter\def\csname res@R53@dbpedia14\endcsname{98.5}
\expandafter\def\csname res@R53@emotion\endcsname{92.3}
\expandafter\def\csname res@R53@helpsteer2\endcsname{60.6}
\expandafter\def\csname res@R53@id\endcsname{85.9}
\expandafter\def\csname res@R53@idece\endcsname{0.025}
\expandafter\def\csname res@R53@kev_transfer\endcsname{74.3}
\expandafter\def\csname res@R53@massive_en\endcsname{91.8}
\expandafter\def\csname res@R53@mnli\endcsname{90.4}
\expandafter\def\csname res@R53@oasst1\endcsname{79.1}
\expandafter\def\csname res@R53@ood\endcsname{88.5}
\expandafter\def\csname res@R53@oodece\endcsname{0.035}
\expandafter\def\csname res@R53@peakgb\endcsname{34.1}
\expandafter\def\csname res@R53@procedural_typed_decisions\endcsname{87.3}
\expandafter\def\csname res@R53@qasc\endcsname{97.2}
\expandafter\def\csname res@R53@snli\endcsname{91.9}
\expandafter\def\csname res@R53@squad2\endcsname{91.8}
\expandafter\def\csname res@R53@sst5\endcsname{62.8}
\expandafter\def\csname res@R53@sstep\endcsname{6.51}
\expandafter\def\csname res@R53@thead\endcsname{1.01}
\expandafter\def\csname res@R53@tydiqa\endcsname{93.2}
\expandafter\def\csname res@R53@typed_decisions\endcsname{78.3}
\expandafter\def\csname res@R54-R32@ag_news\endcsname{$-$0.5}
\expandafter\def\csname res@R54-R32@ag_newsp\endcsname{0.01}
\expandafter\def\csname res@R54-R32@banking77\endcsname{$-$0.2}
\expandafter\def\csname res@R54-R32@banking77p\endcsname{0.6}
\expandafter\def\csname res@R54-R32@belebele_en\endcsname{+1.3}
\expandafter\def\csname res@R54-R32@belebele_enp\endcsname{0.1}
\expandafter\def\csname res@R54-R32@boolq\endcsname{+0.2}
\expandafter\def\csname res@R54-R32@boolqp\endcsname{0.6}
\expandafter\def\csname res@R54-R32@clinc150\endcsname{$-$0.6}
\expandafter\def\csname res@R54-R32@clinc150p\endcsname{0.001}
\expandafter\def\csname res@R54-R32@cosmosqa\endcsname{+0.0}
\expandafter\def\csname res@R54-R32@cosmosqap\endcsname{1}
\expandafter\def\csname res@R54-R32@dbpedia14\endcsname{$-$0.3}
\expandafter\def\csname res@R54-R32@dbpedia14p\endcsname{7e-20}
\expandafter\def\csname res@R54-R32@emotion\endcsname{$-$0.4}
\expandafter\def\csname res@R54-R32@emotionp\endcsname{0.3}
\expandafter\def\csname res@R54-R32@helpsteer2\endcsname{$-$0.2}
\expandafter\def\csname res@R54-R32@helpsteer2p\endcsname{0.5}
\expandafter\def\csname res@R54-R32@id\endcsname{$-$0.18}
\expandafter\def\csname res@R54-R32@idhi\endcsname{$-$0.00}
\expandafter\def\csname res@R54-R32@idlo\endcsname{$-$0.37}
\expandafter\def\csname res@R54-R32@kev_transfer\endcsname{$-$0.9}
\expandafter\def\csname res@R54-R32@kev_transferp\endcsname{0.4}
\expandafter\def\csname res@R54-R32@massive_en\endcsname{+0.3}
\expandafter\def\csname res@R54-R32@massive_enp\endcsname{0.1}
\expandafter\def\csname res@R54-R32@mnli\endcsname{+0.4}
\expandafter\def\csname res@R54-R32@mnlip\endcsname{0.04}
\expandafter\def\csname res@R54-R32@oasst1\endcsname{$-$0.1}
\expandafter\def\csname res@R54-R32@oasst1p\endcsname{0.6}
\expandafter\def\csname res@R54-R32@ood\endcsname{+0.10}
\expandafter\def\csname res@R54-R32@oodhi\endcsname{+0.61}
\expandafter\def\csname res@R54-R32@oodlo\endcsname{$-$0.42}
\expandafter\def\csname res@R54-R32@procedural_typed_decisions\endcsname{$-$3.3}
\expandafter\def\csname res@R54-R32@procedural_typed_decisionsp\endcsname{7e-148}
\expandafter\def\csname res@R54-R32@qasc\endcsname{+0.8}
\expandafter\def\csname res@R54-R32@qascp\endcsname{0.1}
\expandafter\def\csname res@R54-R32@snli\endcsname{+0.5}
\expandafter\def\csname res@R54-R32@snlip\endcsname{0.007}
\expandafter\def\csname res@R54-R32@squad2\endcsname{$-$0.1}
\expandafter\def\csname res@R54-R32@squad2p\endcsname{0.6}
\expandafter\def\csname res@R54-R32@sst5\endcsname{+0.8}
\expandafter\def\csname res@R54-R32@sst5p\endcsname{0.2}
\expandafter\def\csname res@R54-R32@tydiqa\endcsname{$-$0.1}
\expandafter\def\csname res@R54-R32@tydiqap\endcsname{1}
\expandafter\def\csname res@R54-R32@typed_decisions\endcsname{+0.0}
\expandafter\def\csname res@R54-R32@typed_decisionsp\endcsname{1}
\expandafter\def\csname res@R54@ag_news\endcsname{86.6}
\expandafter\def\csname res@R54@banking77\endcsname{93.3}
\expandafter\def\csname res@R54@belebele_en\endcsname{92.1}
\expandafter\def\csname res@R54@boolq\endcsname{92.7}
\expandafter\def\csname res@R54@clinc150\endcsname{94.8}
\expandafter\def\csname res@R54@cosmosqa\endcsname{91.1}
\expandafter\def\csname res@R54@dbpedia14\endcsname{98.4}
\expandafter\def\csname res@R54@emotion\endcsname{92.5}
\expandafter\def\csname res@R54@helpsteer2\endcsname{61.1}
\expandafter\def\csname res@R54@id\endcsname{86.4}
\expandafter\def\csname res@R54@idece\endcsname{0.025}
\expandafter\def\csname res@R54@kev_transfer\endcsname{80.3}
\expandafter\def\csname res@R54@massive_en\endcsname{91.9}
\expandafter\def\csname res@R54@mnli\endcsname{90.5}
\expandafter\def\csname res@R54@oasst1\endcsname{79.2}
\expandafter\def\csname res@R54@ood\endcsname{90.9}
\expandafter\def\csname res@R54@oodece\endcsname{0.026}
\expandafter\def\csname res@R54@peakgb\endcsname{53.0}
\expandafter\def\csname res@R54@procedural_typed_decisions\endcsname{87.0}
\expandafter\def\csname res@R54@qasc\endcsname{97.2}
\expandafter\def\csname res@R54@snli\endcsname{92.6}
\expandafter\def\csname res@R54@squad2\endcsname{92.3}
\expandafter\def\csname res@R54@sst5\endcsname{63.4}
\expandafter\def\csname res@R54@sstep\endcsname{9.03}
\expandafter\def\csname res@R54@thead\endcsname{1.02}
\expandafter\def\csname res@R54@tydiqa\endcsname{93.6}
\expandafter\def\csname res@R54@typed_decisions\endcsname{79.6}
\expandafter\def\csname res@R56@ag_news\endcsname{85.9}
\expandafter\def\csname res@R56@banking77\endcsname{94.0}
\expandafter\def\csname res@R56@belebele_en\endcsname{84.6}
\expandafter\def\csname res@R56@boolq\endcsname{89.0}
\expandafter\def\csname res@R56@clinc150\endcsname{94.4}
\expandafter\def\csname res@R56@cosmosqa\endcsname{84.3}
\expandafter\def\csname res@R56@dbpedia14\endcsname{95.9}
\expandafter\def\csname res@R56@emotion\endcsname{93.2}
\expandafter\def\csname res@R56@helpsteer2\endcsname{60.6}
\expandafter\def\csname res@R56@id\endcsname{85.0}
\expandafter\def\csname res@R56@idece\endcsname{0.029}
\expandafter\def\csname res@R56@kev_transfer\endcsname{71.0}
\expandafter\def\csname res@R56@massive_en\endcsname{91.7}
\expandafter\def\csname res@R56@mnli\endcsname{88.3}
\expandafter\def\csname res@R56@oasst1\endcsname{78.9}
\expandafter\def\csname res@R56@ood\endcsname{85.6}
\expandafter\def\csname res@R56@oodece\endcsname{0.034}
\expandafter\def\csname res@R56@peakgb\endcsname{21.1}
\expandafter\def\csname res@R56@procedural_typed_decisions\endcsname{89.5}
\expandafter\def\csname res@R56@qasc\endcsname{90.7}
\expandafter\def\csname res@R56@snli\endcsname{90.8}
\expandafter\def\csname res@R56@squad2\endcsname{89.5}
\expandafter\def\csname res@R56@sst5\endcsname{58.5}
\expandafter\def\csname res@R56@sstep\endcsname{5.62}
\expandafter\def\csname res@R56@thead\endcsname{1.07}
\expandafter\def\csname res@R56@tydiqa\endcsname{91.9}
\expandafter\def\csname res@R56@typed_decisions\endcsname{79.8}
\expandafter\def\csname res@R57@ag_news\endcsname{81.9}
\expandafter\def\csname res@R57@banking77\endcsname{94.1}
\expandafter\def\csname res@R57@belebele_en\endcsname{90.6}
\expandafter\def\csname res@R57@boolq\endcsname{91.7}
\expandafter\def\csname res@R57@clinc150\endcsname{95.2}
\expandafter\def\csname res@R57@cosmosqa\endcsname{88.9}
\expandafter\def\csname res@R57@dbpedia14\endcsname{98.0}
\expandafter\def\csname res@R57@emotion\endcsname{92.9}
\expandafter\def\csname res@R57@helpsteer2\endcsname{61.8}
\expandafter\def\csname res@R57@id\endcsname{86.4}
\expandafter\def\csname res@R57@idece\endcsname{0.026}
\expandafter\def\csname res@R57@kev_transfer\endcsname{74.1}
\expandafter\def\csname res@R57@massive_en\endcsname{92.1}
\expandafter\def\csname res@R57@mnli\endcsname{89.8}
\expandafter\def\csname res@R57@oasst1\endcsname{79.1}
\expandafter\def\csname res@R57@ood\endcsname{88.1}
\expandafter\def\csname res@R57@oodece\endcsname{0.035}
\expandafter\def\csname res@R57@peakgb\endcsname{31.5}
\expandafter\def\csname res@R57@procedural_typed_decisions\endcsname{91.3}
\expandafter\def\csname res@R57@qasc\endcsname{95.8}
\expandafter\def\csname res@R57@snli\endcsname{92.1}
\expandafter\def\csname res@R57@squad2\endcsname{91.6}
\expandafter\def\csname res@R57@sst5\endcsname{62.5}
\expandafter\def\csname res@R57@sstep\endcsname{7.65}
\expandafter\def\csname res@R57@thead\endcsname{1.08}
\expandafter\def\csname res@R57@tydiqa\endcsname{93.9}
\expandafter\def\csname res@R57@typed_decisions\endcsname{78.6}
\expandafter\def\csname res@R60-R57@ag_news\endcsname{+2.4}
\expandafter\def\csname res@R60-R57@ag_newsp\endcsname{3e-22}
\expandafter\def\csname res@R60-R57@banking77\endcsname{+0.1}
\expandafter\def\csname res@R60-R57@banking77p\endcsname{0.7}
\expandafter\def\csname res@R60-R57@belebele_en\endcsname{$-$0.8}
\expandafter\def\csname res@R60-R57@belebele_enp\endcsname{0.4}
\expandafter\def\csname res@R60-R57@boolq\endcsname{$-$0.5}
\expandafter\def\csname res@R60-R57@boolqp\endcsname{0.2}
\expandafter\def\csname res@R60-R57@clinc150\endcsname{$-$0.3}
\expandafter\def\csname res@R60-R57@clinc150p\endcsname{0.3}
\expandafter\def\csname res@R60-R57@cosmosqa\endcsname{+0.6}
\expandafter\def\csname res@R60-R57@cosmosqap\endcsname{0.3}
\expandafter\def\csname res@R60-R57@dbpedia14\endcsname{+0.1}
\expandafter\def\csname res@R60-R57@dbpedia14p\endcsname{0.06}
\expandafter\def\csname res@R60-R57@emotion\endcsname{+0.0}
\expandafter\def\csname res@R60-R57@emotionp\endcsname{1}
\expandafter\def\csname res@R60-R57@helpsteer2\endcsname{+0.0}
\expandafter\def\csname res@R60-R57@helpsteer2p\endcsname{1}
\expandafter\def\csname res@R60-R57@id\endcsname{$-$0.06}
\expandafter\def\csname res@R60-R57@idhi\endcsname{+0.14}
\expandafter\def\csname res@R60-R57@idlo\endcsname{$-$0.27}
\expandafter\def\csname res@R60-R57@kev_transfer\endcsname{+0.9}
\expandafter\def\csname res@R60-R57@kev_transferp\endcsname{0.4}
\expandafter\def\csname res@R60-R57@massive_en\endcsname{+0.3}
\expandafter\def\csname res@R60-R57@massive_enp\endcsname{0.2}
\expandafter\def\csname res@R60-R57@mnli\endcsname{$-$0.1}
\expandafter\def\csname res@R60-R57@mnlip\endcsname{0.8}
\expandafter\def\csname res@R60-R57@oasst1\endcsname{+0.1}
\expandafter\def\csname res@R60-R57@oasst1p\endcsname{0.7}
\expandafter\def\csname res@R60-R57@ood\endcsname{$-$0.08}
\expandafter\def\csname res@R60-R57@oodhi\endcsname{+0.50}
\expandafter\def\csname res@R60-R57@oodlo\endcsname{$-$0.64}
\expandafter\def\csname res@R60-R57@procedural_typed_decisions\endcsname{$-$0.1}
\expandafter\def\csname res@R60-R57@procedural_typed_decisionsp\endcsname{0.4}
\expandafter\def\csname res@R60-R57@qasc\endcsname{$-$2.9}
\expandafter\def\csname res@R60-R57@qascp\endcsname{3e-05}
\expandafter\def\csname res@R60-R57@snli\endcsname{$-$0.3}
\expandafter\def\csname res@R60-R57@snlip\endcsname{0.1}
\expandafter\def\csname res@R60-R57@squad2\endcsname{$-$0.1}
\expandafter\def\csname res@R60-R57@squad2p\endcsname{0.5}
\expandafter\def\csname res@R60-R57@sst5\endcsname{$-$0.9}
\expandafter\def\csname res@R60-R57@sst5p\endcsname{0.2}
\expandafter\def\csname res@R60-R57@tydiqa\endcsname{$-$0.4}
\expandafter\def\csname res@R60-R57@tydiqap\endcsname{0.6}
\expandafter\def\csname res@R60-R57@typed_decisions\endcsname{+0.8}
\expandafter\def\csname res@R60-R57@typed_decisionsp\endcsname{0.2}
\expandafter\def\csname res@R60@ag_news\endcsname{84.2}
\expandafter\def\csname res@R60@banking77\endcsname{94.3}
\expandafter\def\csname res@R60@belebele_en\endcsname{89.8}
\expandafter\def\csname res@R60@boolq\endcsname{91.2}
\expandafter\def\csname res@R60@clinc150\endcsname{94.9}
\expandafter\def\csname res@R60@cosmosqa\endcsname{89.6}
\expandafter\def\csname res@R60@dbpedia14\endcsname{98.1}
\expandafter\def\csname res@R60@emotion\endcsname{92.9}
\expandafter\def\csname res@R60@helpsteer2\endcsname{61.8}
\expandafter\def\csname res@R60@id\endcsname{86.3}
\expandafter\def\csname res@R60@idece\endcsname{0.029}
\expandafter\def\csname res@R60@kev_transfer\endcsname{75.0}
\expandafter\def\csname res@R60@massive_en\endcsname{92.3}
\expandafter\def\csname res@R60@mnli\endcsname{89.7}
\expandafter\def\csname res@R60@oasst1\endcsname{79.2}
\expandafter\def\csname res@R60@ood\endcsname{88.0}
\expandafter\def\csname res@R60@oodece\endcsname{0.031}
\expandafter\def\csname res@R60@peakgb\endcsname{32.0}
\expandafter\def\csname res@R60@procedural_typed_decisions\endcsname{91.2}
\expandafter\def\csname res@R60@qasc\endcsname{92.9}
\expandafter\def\csname res@R60@snli\endcsname{91.8}
\expandafter\def\csname res@R60@squad2\endcsname{91.5}
\expandafter\def\csname res@R60@sst5\endcsname{61.6}
\expandafter\def\csname res@R60@sstep\endcsname{9.08}
\expandafter\def\csname res@R60@thead\endcsname{1.11}
\expandafter\def\csname res@R60@tydiqa\endcsname{93.5}
\expandafter\def\csname res@R60@typed_decisions\endcsname{79.3}
\expandafter\def\csname res@R63c@ag_news\endcsname{86.9}
\expandafter\def\csname res@R63c@banking77\endcsname{91.1}
\expandafter\def\csname res@R63c@belebele_en\endcsname{89.8}
\expandafter\def\csname res@R63c@boolq\endcsname{92.0}
\expandafter\def\csname res@R63c@clinc150\endcsname{92.8}
\expandafter\def\csname res@R63c@cosmosqa\endcsname{89.7}
\expandafter\def\csname res@R63c@dbpedia14\endcsname{98.4}
\expandafter\def\csname res@R63c@emotion\endcsname{84.0}
\expandafter\def\csname res@R63c@helpsteer2\endcsname{61.0}
\expandafter\def\csname res@R63c@id\endcsname{85.0}
\expandafter\def\csname res@R63c@idece\endcsname{0.026}
\expandafter\def\csname res@R63c@kev_transfer\endcsname{81.3}
\expandafter\def\csname res@R63c@massive_en\endcsname{90.3}
\expandafter\def\csname res@R63c@mnli\endcsname{90.6}
\expandafter\def\csname res@R63c@oasst1\endcsname{79.1}
\expandafter\def\csname res@R63c@ood\endcsname{90.5}
\expandafter\def\csname res@R63c@oodece\endcsname{0.026}
\expandafter\def\csname res@R63c@peakgb\endcsname{47.6}
\expandafter\def\csname res@R63c@procedural_typed_decisions\endcsname{89.5}
\expandafter\def\csname res@R63c@qasc\endcsname{96.1}
\expandafter\def\csname res@R63c@snli\endcsname{92.4}
\expandafter\def\csname res@R63c@squad2\endcsname{91.6}
\expandafter\def\csname res@R63c@sst5\endcsname{61.4}
\expandafter\def\csname res@R63c@sstep\endcsname{12.55}
\expandafter\def\csname res@R63c@thead\endcsname{1.03}
\expandafter\def\csname res@R63c@tydiqa\endcsname{93.0}
\expandafter\def\csname res@R63c@typed_decisions\endcsname{76.7}
\expandafter\def\csname res@R66-R26@ag_news\endcsname{$-$3.5}
\expandafter\def\csname res@R66-R26@ag_newsp\endcsname{2e-40}
\expandafter\def\csname res@R66-R26@banking77\endcsname{$-$0.7}
\expandafter\def\csname res@R66-R26@banking77p\endcsname{0.04}
\expandafter\def\csname res@R66-R26@belebele_en\endcsname{$-$6.0}
\expandafter\def\csname res@R66-R26@belebele_enp\endcsname{8e-09}
\expandafter\def\csname res@R66-R26@boolq\endcsname{$-$0.2}
\expandafter\def\csname res@R66-R26@boolqp\endcsname{0.8}
\expandafter\def\csname res@R66-R26@clinc150\endcsname{+0.0}
\expandafter\def\csname res@R66-R26@clinc150p\endcsname{0.9}
\expandafter\def\csname res@R66-R26@cosmosqa\endcsname{$-$0.8}
\expandafter\def\csname res@R66-R26@cosmosqap\endcsname{0.2}
\expandafter\def\csname res@R66-R26@dbpedia14\endcsname{$-$0.1}
\expandafter\def\csname res@R66-R26@dbpedia14p\endcsname{0.0003}
\expandafter\def\csname res@R66-R26@emotion\endcsname{$-$1.1}
\expandafter\def\csname res@R66-R26@emotionp\endcsname{0.08}
\expandafter\def\csname res@R66-R26@helpsteer2\endcsname{$-$0.6}
\expandafter\def\csname res@R66-R26@helpsteer2p\endcsname{0.2}
\expandafter\def\csname res@R66-R26@id\endcsname{$-$0.66}
\expandafter\def\csname res@R66-R26@idhi\endcsname{$-$0.42}
\expandafter\def\csname res@R66-R26@idlo\endcsname{$-$0.88}
\expandafter\def\csname res@R66-R26@kev_transfer\endcsname{+1.3}
\expandafter\def\csname res@R66-R26@kev_transferp\endcsname{0.2}
\expandafter\def\csname res@R66-R26@massive_en\endcsname{$-$0.8}
\expandafter\def\csname res@R66-R26@massive_enp\endcsname{0.0007}
\expandafter\def\csname res@R66-R26@mnli\endcsname{$-$0.1}
\expandafter\def\csname res@R66-R26@mnlip\endcsname{0.6}
\expandafter\def\csname res@R66-R26@oasst1\endcsname{$-$0.2}
\expandafter\def\csname res@R66-R26@oasst1p\endcsname{0.3}
\expandafter\def\csname res@R66-R26@ood\endcsname{$-$1.68}
\expandafter\def\csname res@R66-R26@oodhi\endcsname{$-$1.06}
\expandafter\def\csname res@R66-R26@oodlo\endcsname{$-$2.31}
\expandafter\def\csname res@R66-R26@procedural_typed_decisions\endcsname{$-$3.0}
\expandafter\def\csname res@R66-R26@procedural_typed_decisionsp\endcsname{3e-123}
\expandafter\def\csname res@R66-R26@qasc\endcsname{$-$0.1}
\expandafter\def\csname res@R66-R26@qascp\endcsname{1}
\expandafter\def\csname res@R66-R26@snli\endcsname{$-$0.4}
\expandafter\def\csname res@R66-R26@snlip\endcsname{0.09}
\expandafter\def\csname res@R66-R26@squad2\endcsname{+0.0}
\expandafter\def\csname res@R66-R26@squad2p\endcsname{1}
\expandafter\def\csname res@R66-R26@sst5\endcsname{$-$0.2}
\expandafter\def\csname res@R66-R26@sst5p\endcsname{0.9}
\expandafter\def\csname res@R66-R26@tydiqa\endcsname{$-$0.3}
\expandafter\def\csname res@R66-R26@tydiqap\endcsname{0.7}
\expandafter\def\csname res@R66-R26@typed_decisions\endcsname{$-$1.7}
\expandafter\def\csname res@R66-R26@typed_decisionsp\endcsname{0.02}
\expandafter\def\csname res@R66@ag_news\endcsname{83.1}
\expandafter\def\csname res@R66@banking77\endcsname{93.1}
\expandafter\def\csname res@R66@belebele_en\endcsname{84.8}
\expandafter\def\csname res@R66@boolq\endcsname{90.5}
\expandafter\def\csname res@R66@clinc150\endcsname{94.4}
\expandafter\def\csname res@R66@cosmosqa\endcsname{86.6}
\expandafter\def\csname res@R66@dbpedia14\endcsname{98.3}
\expandafter\def\csname res@R66@emotion\endcsname{92.3}
\expandafter\def\csname res@R66@helpsteer2\endcsname{60.7}
\expandafter\def\csname res@R66@id\endcsname{85.2}
\expandafter\def\csname res@R66@idece\endcsname{0.025}
\expandafter\def\csname res@R66@kev_transfer\endcsname{73.8}
\expandafter\def\csname res@R66@massive_en\endcsname{91.5}
\expandafter\def\csname res@R66@mnli\endcsname{89.1}
\expandafter\def\csname res@R66@oasst1\endcsname{78.9}
\expandafter\def\csname res@R66@ood\endcsname{87.2}
\expandafter\def\csname res@R66@oodece\endcsname{0.033}
\expandafter\def\csname res@R66@peakgb\endcsname{21.0}
\expandafter\def\csname res@R66@procedural_typed_decisions\endcsname{87.5}
\expandafter\def\csname res@R66@qasc\endcsname{96.0}
\expandafter\def\csname res@R66@snli\endcsname{91.0}
\expandafter\def\csname res@R66@squad2\endcsname{91.1}
\expandafter\def\csname res@R66@sst5\endcsname{61.4}
\expandafter\def\csname res@R66@sstep\endcsname{5.47}
\expandafter\def\csname res@R66@thead\endcsname{1.02}
\expandafter\def\csname res@R66@tydiqa\endcsname{92.5}
\expandafter\def\csname res@R66@typed_decisions\endcsname{77.8}
\expandafter\def\csname res@R67-R10@ag_news\endcsname{$-$4.0}
\expandafter\def\csname res@R67-R10@ag_newsp\endcsname{8e-41}
\expandafter\def\csname res@R67-R10@banking77\endcsname{$-$0.7}
\expandafter\def\csname res@R67-R10@banking77p\endcsname{0.03}
\expandafter\def\csname res@R67-R10@belebele_en\endcsname{$-$6.9}
\expandafter\def\csname res@R67-R10@belebele_enp\endcsname{2e-06}
\expandafter\def\csname res@R67-R10@boolq\endcsname{+0.6}
\expandafter\def\csname res@R67-R10@boolqp\endcsname{0.4}
\expandafter\def\csname res@R67-R10@clinc150\endcsname{+1.0}
\expandafter\def\csname res@R67-R10@clinc150p\endcsname{0.0005}
\expandafter\def\csname res@R67-R10@cosmosqa\endcsname{$-$0.2}
\expandafter\def\csname res@R67-R10@cosmosqap\endcsname{0.8}
\expandafter\def\csname res@R67-R10@dbpedia14\endcsname{$-$0.9}
\expandafter\def\csname res@R67-R10@dbpedia14p\endcsname{1e-40}
\expandafter\def\csname res@R67-R10@emotion\endcsname{$-$0.6}
\expandafter\def\csname res@R67-R10@emotionp\endcsname{0.2}
\expandafter\def\csname res@R67-R10@helpsteer2\endcsname{$-$0.1}
\expandafter\def\csname res@R67-R10@helpsteer2p\endcsname{0.8}
\expandafter\def\csname res@R67-R10@id\endcsname{$-$0.02}
\expandafter\def\csname res@R67-R10@idhi\endcsname{+0.21}
\expandafter\def\csname res@R67-R10@idlo\endcsname{$-$0.26}
\expandafter\def\csname res@R67-R10@kev_transfer\endcsname{$-$1.3}
\expandafter\def\csname res@R67-R10@kev_transferp\endcsname{0.3}
\expandafter\def\csname res@R67-R10@massive_en\endcsname{+0.1}
\expandafter\def\csname res@R67-R10@massive_enp\endcsname{0.8}
\expandafter\def\csname res@R67-R10@mnli\endcsname{+0.2}
\expandafter\def\csname res@R67-R10@mnlip\endcsname{0.5}
\expandafter\def\csname res@R67-R10@oasst1\endcsname{+0.0}
\expandafter\def\csname res@R67-R10@oasst1p\endcsname{0.8}
\expandafter\def\csname res@R67-R10@ood\endcsname{$-$3.05}
\expandafter\def\csname res@R67-R10@oodhi\endcsname{$-$2.22}
\expandafter\def\csname res@R67-R10@oodlo\endcsname{$-$3.81}
\expandafter\def\csname res@R67-R10@procedural_typed_decisions\endcsname{$-$0.8}
\expandafter\def\csname res@R67-R10@procedural_typed_decisionsp\endcsname{9e-08}
\expandafter\def\csname res@R67-R10@qasc\endcsname{$-$2.1}
\expandafter\def\csname res@R67-R10@qascp\endcsname{0.04}
\expandafter\def\csname res@R67-R10@snli\endcsname{+0.3}
\expandafter\def\csname res@R67-R10@snlip\endcsname{0.3}
\expandafter\def\csname res@R67-R10@squad2\endcsname{$-$0.3}
\expandafter\def\csname res@R67-R10@squad2p\endcsname{0.4}
\expandafter\def\csname res@R67-R10@sst5\endcsname{+0.7}
\expandafter\def\csname res@R67-R10@sst5p\endcsname{0.4}
\expandafter\def\csname res@R67-R10@tydiqa\endcsname{$-$0.6}
\expandafter\def\csname res@R67-R10@tydiqap\endcsname{0.4}
\expandafter\def\csname res@R67-R10@typed_decisions\endcsname{+0.1}
\expandafter\def\csname res@R67-R10@typed_decisionsp\endcsname{1}
\expandafter\def\csname res@R67@ag_news\endcsname{75.8}
\expandafter\def\csname res@R67@banking77\endcsname{92.5}
\expandafter\def\csname res@R67@belebele_en\endcsname{69.2}
\expandafter\def\csname res@R67@boolq\endcsname{84.7}
\expandafter\def\csname res@R67@clinc150\endcsname{92.3}
\expandafter\def\csname res@R67@cosmosqa\endcsname{73.8}
\expandafter\def\csname res@R67@dbpedia14\endcsname{94.9}
\expandafter\def\csname res@R67@emotion\endcsname{92.9}
\expandafter\def\csname res@R67@helpsteer2\endcsname{60.0}
\expandafter\def\csname res@R67@id\endcsname{81.6}
\expandafter\def\csname res@R67@idece\endcsname{0.032}
\expandafter\def\csname res@R67@kev_transfer\endcsname{56.4}
\expandafter\def\csname res@R67@massive_en\endcsname{91.1}
\expandafter\def\csname res@R67@mnli\endcsname{83.6}
\expandafter\def\csname res@R67@oasst1\endcsname{78.9}
\expandafter\def\csname res@R67@ood\endcsname{76.3}
\expandafter\def\csname res@R67@oodece\endcsname{0.067}
\expandafter\def\csname res@R67@peakgb\endcsname{6.3}
\expandafter\def\csname res@R67@procedural_typed_decisions\endcsname{78.5}
\expandafter\def\csname res@R67@qasc\endcsname{85.4}
\expandafter\def\csname res@R67@snli\endcsname{87.5}
\expandafter\def\csname res@R67@squad2\endcsname{83.6}
\expandafter\def\csname res@R67@sst5\endcsname{57.3}
\expandafter\def\csname res@R67@sstep\endcsname{3.02}
\expandafter\def\csname res@R67@thead\endcsname{1.05}
\expandafter\def\csname res@R67@tydiqa\endcsname{91.1}
\expandafter\def\csname res@R67@typed_decisions\endcsname{76.6}
\expandafter\def\csname res@R68-R07@ag_news\endcsname{+1.6}
\expandafter\def\csname res@R68-R07@ag_newsp\endcsname{4e-08}
\expandafter\def\csname res@R68-R07@banking77\endcsname{+0.0}
\expandafter\def\csname res@R68-R07@banking77p\endcsname{1}
\expandafter\def\csname res@R68-R07@belebele_en\endcsname{+0.7}
\expandafter\def\csname res@R68-R07@belebele_enp\endcsname{0.6}
\expandafter\def\csname res@R68-R07@boolq\endcsname{+1.3}
\expandafter\def\csname res@R68-R07@boolqp\endcsname{0.04}
\expandafter\def\csname res@R68-R07@clinc150\endcsname{+0.3}
\expandafter\def\csname res@R68-R07@clinc150p\endcsname{0.3}
\expandafter\def\csname res@R68-R07@cosmosqa\endcsname{+0.0}
\expandafter\def\csname res@R68-R07@cosmosqap\endcsname{1}
\expandafter\def\csname res@R68-R07@dbpedia14\endcsname{+2.1}
\expandafter\def\csname res@R68-R07@dbpedia14p\endcsname{1e-202}
\expandafter\def\csname res@R68-R07@emotion\endcsname{$-$0.7}
\expandafter\def\csname res@R68-R07@emotionp\endcsname{0.1}
\expandafter\def\csname res@R68-R07@helpsteer2\endcsname{+0.4}
\expandafter\def\csname res@R68-R07@helpsteer2p\endcsname{0.2}
\expandafter\def\csname res@R68-R07@id\endcsname{+0.21}
\expandafter\def\csname res@R68-R07@idhi\endcsname{+0.44}
\expandafter\def\csname res@R68-R07@idlo\endcsname{$-$0.03}
\expandafter\def\csname res@R68-R07@kev_transfer\endcsname{+0.8}
\expandafter\def\csname res@R68-R07@kev_transferp\endcsname{0.6}
\expandafter\def\csname res@R68-R07@massive_en\endcsname{$-$0.0}
\expandafter\def\csname res@R68-R07@massive_enp\endcsname{1}
\expandafter\def\csname res@R68-R07@mnli\endcsname{$-$0.0}
\expandafter\def\csname res@R68-R07@mnlip\endcsname{1}
\expandafter\def\csname res@R68-R07@oasst1\endcsname{$-$0.0}
\expandafter\def\csname res@R68-R07@oasst1p\endcsname{1}
\expandafter\def\csname res@R68-R07@ood\endcsname{+1.23}
\expandafter\def\csname res@R68-R07@oodhi\endcsname{+1.99}
\expandafter\def\csname res@R68-R07@oodlo\endcsname{+0.47}
\expandafter\def\csname res@R68-R07@procedural_typed_decisions\endcsname{+0.4}
\expandafter\def\csname res@R68-R07@procedural_typed_decisionsp\endcsname{0.004}
\expandafter\def\csname res@R68-R07@qasc\endcsname{+1.1}
\expandafter\def\csname res@R68-R07@qascp\endcsname{0.3}
\expandafter\def\csname res@R68-R07@snli\endcsname{$-$0.5}
\expandafter\def\csname res@R68-R07@snlip\endcsname{0.03}
\expandafter\def\csname res@R68-R07@squad2\endcsname{$-$0.3}
\expandafter\def\csname res@R68-R07@squad2p\endcsname{0.3}
\expandafter\def\csname res@R68-R07@sst5\endcsname{+0.0}
\expandafter\def\csname res@R68-R07@sst5p\endcsname{1}
\expandafter\def\csname res@R68-R07@tydiqa\endcsname{+0.7}
\expandafter\def\csname res@R68-R07@tydiqap\endcsname{0.4}
\expandafter\def\csname res@R68-R07@typed_decisions\endcsname{+1.5}
\expandafter\def\csname res@R68-R07@typed_decisionsp\endcsname{0.04}
\expandafter\def\csname res@R68@ag_news\endcsname{82.1}
\expandafter\def\csname res@R68@banking77\endcsname{92.8}
\expandafter\def\csname res@R68@belebele_en\endcsname{66.3}
\expandafter\def\csname res@R68@boolq\endcsname{84.8}
\expandafter\def\csname res@R68@clinc150\endcsname{92.7}
\expandafter\def\csname res@R68@cosmosqa\endcsname{74.2}
\expandafter\def\csname res@R68@dbpedia14\endcsname{96.9}
\expandafter\def\csname res@R68@emotion\endcsname{92.5}
\expandafter\def\csname res@R68@helpsteer2\endcsname{60.4}
\expandafter\def\csname res@R68@id\endcsname{82.0}
\expandafter\def\csname res@R68@idece\endcsname{0.034}
\expandafter\def\csname res@R68@kev_transfer\endcsname{56.1}
\expandafter\def\csname res@R68@massive_en\endcsname{91.1}
\expandafter\def\csname res@R68@mnli\endcsname{84.0}
\expandafter\def\csname res@R68@oasst1\endcsname{78.8}
\expandafter\def\csname res@R68@ood\endcsname{78.0}
\expandafter\def\csname res@R68@oodece\endcsname{0.069}
\expandafter\def\csname res@R68@peakgb\endcsname{6.3}
\expandafter\def\csname res@R68@procedural_typed_decisions\endcsname{81.9}
\expandafter\def\csname res@R68@qasc\endcsname{88.4}
\expandafter\def\csname res@R68@snli\endcsname{87.4}
\expandafter\def\csname res@R68@squad2\endcsname{84.4}
\expandafter\def\csname res@R68@sst5\endcsname{56.4}
\expandafter\def\csname res@R68@sstep\endcsname{3.17}
\expandafter\def\csname res@R68@thead\endcsname{1.08}
\expandafter\def\csname res@R68@tydiqa\endcsname{91.4}
\expandafter\def\csname res@R68@typed_decisions\endcsname{76.3}
\expandafter\def\csname res@R69-R68@ag_news\endcsname{$-$3.7}
\expandafter\def\csname res@R69-R68@ag_newsp\endcsname{7e-31}
\expandafter\def\csname res@R69-R68@banking77\endcsname{$-$0.1}
\expandafter\def\csname res@R69-R68@banking77p\endcsname{0.7}
\expandafter\def\csname res@R69-R68@belebele_en\endcsname{+1.9}
\expandafter\def\csname res@R69-R68@belebele_enp\endcsname{0.2}
\expandafter\def\csname res@R69-R68@boolq\endcsname{$-$0.1}
\expandafter\def\csname res@R69-R68@boolqp\endcsname{0.9}
\expandafter\def\csname res@R69-R68@clinc150\endcsname{$-$0.2}
\expandafter\def\csname res@R69-R68@clinc150p\endcsname{0.5}
\expandafter\def\csname res@R69-R68@cosmosqa\endcsname{$-$0.8}
\expandafter\def\csname res@R69-R68@cosmosqap\endcsname{0.3}
\expandafter\def\csname res@R69-R68@dbpedia14\endcsname{$-$3.1}
\expandafter\def\csname res@R69-R68@dbpedia14p\endcsname{0}
\expandafter\def\csname res@R69-R68@emotion\endcsname{+0.4}
\expandafter\def\csname res@R69-R68@emotionp\endcsname{0.6}
\expandafter\def\csname res@R69-R68@helpsteer2\endcsname{$-$1.2}
\expandafter\def\csname res@R69-R68@helpsteer2p\endcsname{0.02}
\expandafter\def\csname res@R69-R68@id\endcsname{$-$0.08}
\expandafter\def\csname res@R69-R68@idhi\endcsname{+0.18}
\expandafter\def\csname res@R69-R68@idlo\endcsname{$-$0.33}
\expandafter\def\csname res@R69-R68@kev_transfer\endcsname{+4.6}
\expandafter\def\csname res@R69-R68@kev_transferp\endcsname{0.0003}
\expandafter\def\csname res@R69-R68@massive_en\endcsname{$-$0.4}
\expandafter\def\csname res@R69-R68@massive_enp\endcsname{0.2}
\expandafter\def\csname res@R69-R68@mnli\endcsname{$-$0.1}
\expandafter\def\csname res@R69-R68@mnlip\endcsname{0.7}
\expandafter\def\csname res@R69-R68@oasst1\endcsname{+0.2}
\expandafter\def\csname res@R69-R68@oasst1p\endcsname{0.3}
\expandafter\def\csname res@R69-R68@ood\endcsname{$-$0.55}
\expandafter\def\csname res@R69-R68@oodhi\endcsname{+0.32}
\expandafter\def\csname res@R69-R68@oodlo\endcsname{$-$1.35}
\expandafter\def\csname res@R69-R68@procedural_typed_decisions\endcsname{+1.0}
\expandafter\def\csname res@R69-R68@procedural_typed_decisionsp\endcsname{3e-10}
\expandafter\def\csname res@R69-R68@qasc\endcsname{$-$2.4}
\expandafter\def\csname res@R69-R68@qascp\endcsname{0.04}
\expandafter\def\csname res@R69-R68@snli\endcsname{+0.4}
\expandafter\def\csname res@R69-R68@snlip\endcsname{0.1}
\expandafter\def\csname res@R69-R68@squad2\endcsname{+0.5}
\expandafter\def\csname res@R69-R68@squad2p\endcsname{0.1}
\expandafter\def\csname res@R69-R68@sst5\endcsname{$-$0.3}
\expandafter\def\csname res@R69-R68@sst5p\endcsname{0.8}
\expandafter\def\csname res@R69-R68@tydiqa\endcsname{$-$0.2}
\expandafter\def\csname res@R69-R68@tydiqap\endcsname{0.9}
\expandafter\def\csname res@R69-R68@typed_decisions\endcsname{$-$0.1}
\expandafter\def\csname res@R69-R68@typed_decisionsp\endcsname{0.9}
\expandafter\def\csname res@R69@ag_news\endcsname{78.3}
\expandafter\def\csname res@R69@banking77\endcsname{92.7}
\expandafter\def\csname res@R69@belebele_en\endcsname{68.2}
\expandafter\def\csname res@R69@boolq\endcsname{84.7}
\expandafter\def\csname res@R69@clinc150\endcsname{92.4}
\expandafter\def\csname res@R69@cosmosqa\endcsname{73.4}
\expandafter\def\csname res@R69@dbpedia14\endcsname{93.8}
\expandafter\def\csname res@R69@emotion\endcsname{92.9}
\expandafter\def\csname res@R69@helpsteer2\endcsname{59.3}
\expandafter\def\csname res@R69@id\endcsname{81.9}
\expandafter\def\csname res@R69@idece\endcsname{0.034}
\expandafter\def\csname res@R69@kev_transfer\endcsname{60.7}
\expandafter\def\csname res@R69@massive_en\endcsname{90.7}
\expandafter\def\csname res@R69@mnli\endcsname{83.9}
\expandafter\def\csname res@R69@oasst1\endcsname{79.0}
\expandafter\def\csname res@R69@ood\endcsname{77.4}
\expandafter\def\csname res@R69@oodece\endcsname{0.058}
\expandafter\def\csname res@R69@peakgb\endcsname{6.3}
\expandafter\def\csname res@R69@procedural_typed_decisions\endcsname{82.9}
\expandafter\def\csname res@R69@qasc\endcsname{86.1}
\expandafter\def\csname res@R69@snli\endcsname{87.8}
\expandafter\def\csname res@R69@squad2\endcsname{84.9}
\expandafter\def\csname res@R69@sst5\endcsname{56.1}
\expandafter\def\csname res@R69@sstep\endcsname{7.98}
\expandafter\def\csname res@R69@thead\endcsname{1.12}
\expandafter\def\csname res@R69@tydiqa\endcsname{91.2}
\expandafter\def\csname res@R69@typed_decisions\endcsname{76.2}
\expandafter\def\csname res@all@runs\endcsname{31}
\expandafter\def\csname res@all@trainh\endcsname{353}
\expandafter\def\csname res@cost@candidate@inferms\endcsname{9.9}
\expandafter\def\csname res@cost@candidate@trainh\endcsname{17.4}
\expandafter\def\csname res@cost@clef@inferms\endcsname{13.6}
\expandafter\def\csname res@cost@clef@trainh\endcsname{8.7}
\expandafter\def\csname res@cost@cosine@inferms\endcsname{6.4}
\expandafter\def\csname res@cost@cosine@trainh\endcsname{6.9}
\expandafter\def\csname res@cost@cosine_perq@inferms\endcsname{9.9}
\expandafter\def\csname res@cost@cosine_perq@trainh\endcsname{17.4}
\expandafter\def\csname res@cost@d3@inferms\endcsname{9.4}
\expandafter\def\csname res@cost@d3@trainh\endcsname{17.6}
\expandafter\def\csname res@cost@nimble@inferms\endcsname{9.4}
\expandafter\def\csname res@cost@nimble@trainh\endcsname{17.8}
\expandafter\def\csname res@cost@q3-0.6b@lat13@joint\endcsname{30}
\expandafter\def\csname res@cost@q3-0.6b@lat13@perq\endcsname{42}
\expandafter\def\csname res@cost@q3-0.6b@lat1@joint\endcsname{29}
\expandafter\def\csname res@cost@q3-0.6b@lat1@perq\endcsname{29}
\expandafter\def\csname res@cost@q3-0.6b@lat2@joint\endcsname{29}
\expandafter\def\csname res@cost@q3-0.6b@lat2@perq\endcsname{32}
\expandafter\def\csname res@cost@q3-0.6b@lat4@joint\endcsname{29}
\expandafter\def\csname res@cost@q3-0.6b@lat4@perq\endcsname{32}
\expandafter\def\csname res@cost@q3-0.6b@lat8@joint\endcsname{29}
\expandafter\def\csname res@cost@q3-0.6b@lat8@perq\endcsname{33}
\expandafter\def\csname res@cost@q3-4b@lat13@joint\endcsname{40}
\expandafter\def\csname res@cost@q3-4b@lat13@perq\endcsname{136}
\expandafter\def\csname res@cost@q3-4b@lat1@joint\endcsname{38}
\expandafter\def\csname res@cost@q3-4b@lat1@perq\endcsname{38}
\expandafter\def\csname res@cost@q3-4b@lat2@joint\endcsname{38}
\expandafter\def\csname res@cost@q3-4b@lat2@perq\endcsname{42}
\expandafter\def\csname res@cost@q3-4b@lat4@joint\endcsname{38}
\expandafter\def\csname res@cost@q3-4b@lat4@perq\endcsname{46}
\expandafter\def\csname res@cost@q3-4b@lat8@joint\endcsname{39}
\expandafter\def\csname res@cost@q3-4b@lat8@perq\endcsname{86}
\expandafter\def\csname res@cost@qpr\endcsname{2.5}
\expandafter\def\csname res@cost@stepx\endcsname{2.53}
\expandafter\def\csname res@eff@q3-0.6b@candidate@headpct\endcsname{0.0038\%}
\expandafter\def\csname res@eff@q3-0.6b@candidate@p50\endcsname{29.3}
\expandafter\def\csname res@eff@q3-0.6b@candidate@qps\endcsname{254}
\expandafter\def\csname res@eff@q3-0.6b@candidate@rowms\endcsname{9.9}
\expandafter\def\csname res@eff@q3-0.6b@candidate@step\endcsname{7.30}
\expandafter\def\csname res@eff@q3-0.6b@clef@headpct\endcsname{0.7\%}
\expandafter\def\csname res@eff@q3-0.6b@clef@p50\endcsname{35.1}
\expandafter\def\csname res@eff@q3-0.6b@clef@qps\endcsname{186}
\expandafter\def\csname res@eff@q3-0.6b@clef@rowms\endcsname{13.6}
\expandafter\def\csname res@eff@q3-0.6b@clef@step\endcsname{3.65}
\expandafter\def\csname res@eff@q3-0.6b@cosine@headpct\endcsname{0.0018\%}
\expandafter\def\csname res@eff@q3-0.6b@cosine@p50\endcsname{29.1}
\expandafter\def\csname res@eff@q3-0.6b@cosine@qps\endcsname{398}
\expandafter\def\csname res@eff@q3-0.6b@cosine@rowms\endcsname{6.4}
\expandafter\def\csname res@eff@q3-0.6b@cosine@step\endcsname{2.89}
\expandafter\def\csname res@eff@q3-0.6b@cosineloop@step\endcsname{3.02}
\expandafter\def\csname res@eff@q3-0.6b@cosineloop@stepperq\endcsname{7.39}
\expandafter\def\csname res@eff@q3-0.6b@d3@headpct\endcsname{$<$0.001\%}
\expandafter\def\csname res@eff@q3-0.6b@d3@p50\endcsname{28.6}
\expandafter\def\csname res@eff@q3-0.6b@d3@qps\endcsname{269}
\expandafter\def\csname res@eff@q3-0.6b@d3@rowms\endcsname{9.4}
\expandafter\def\csname res@eff@q3-0.6b@d3@step\endcsname{7.37}
\expandafter\def\csname res@eff@q3-0.6b@fastdiff\endcsname{2e-07}
\expandafter\def\csname res@eff@q3-0.6b@joint@flopx\endcsname{1.00}
\expandafter\def\csname res@eff@q3-0.6b@joint@p50\endcsname{29.1}
\expandafter\def\csname res@eff@q3-0.6b@joint@p95\endcsname{29.8}
\expandafter\def\csname res@eff@q3-0.6b@joint@qps\endcsname{398}
\expandafter\def\csname res@eff@q3-0.6b@joint@step\endcsname{2.89}
\expandafter\def\csname res@eff@q3-0.6b@nimble@headpct\endcsname{$<$0.001\%}
\expandafter\def\csname res@eff@q3-0.6b@nimble@p50\endcsname{28.7}
\expandafter\def\csname res@eff@q3-0.6b@nimble@qps\endcsname{269}
\expandafter\def\csname res@eff@q3-0.6b@nimble@rowms\endcsname{9.4}
\expandafter\def\csname res@eff@q3-0.6b@nimble@step\endcsname{7.45}
\expandafter\def\csname res@eff@q3-0.6b@perq@flopx\endcsname{1.74}
\expandafter\def\csname res@eff@q3-0.6b@perq@p50\endcsname{29.0}
\expandafter\def\csname res@eff@q3-0.6b@perq@p95\endcsname{38.2}
\expandafter\def\csname res@eff@q3-0.6b@perq@qps\endcsname{255}
\expandafter\def\csname res@eff@q3-0.6b@perq@step\endcsname{7.31}
\expandafter\def\csname res@eff@q3-0.6b@sweep13@joint\endcsname{30}
\expandafter\def\csname res@eff@q3-0.6b@sweep13@perq\endcsname{42}
\expandafter\def\csname res@eff@q3-0.6b@sweep1@joint\endcsname{29}
\expandafter\def\csname res@eff@q3-0.6b@sweep1@perq\endcsname{29}
\expandafter\def\csname res@eff@q3-0.6b@sweep2@joint\endcsname{29}
\expandafter\def\csname res@eff@q3-0.6b@sweep2@perq\endcsname{32}
\expandafter\def\csname res@eff@q3-0.6b@sweep4@joint\endcsname{29}
\expandafter\def\csname res@eff@q3-0.6b@sweep4@perq\endcsname{32}
\expandafter\def\csname res@eff@q3-0.6b@sweep8@joint\endcsname{29}
\expandafter\def\csname res@eff@q3-0.6b@sweep8@perq\endcsname{33}
\expandafter\def\csname res@eff@q3-4b@candidate@headpct\endcsname{0.0013\%}
\expandafter\def\csname res@eff@q3-4b@candidate@p50\endcsname{38.0}
\expandafter\def\csname res@eff@q3-4b@candidate@qps\endcsname{75}
\expandafter\def\csname res@eff@q3-4b@candidate@rowms\endcsname{33.7}
\expandafter\def\csname res@eff@q3-4b@candidate@step\endcsname{10.52}
\expandafter\def\csname res@eff@q3-4b@clef@headpct\endcsname{0.12\%}
\expandafter\def\csname res@eff@q3-4b@clef@p50\endcsname{44.8}
\expandafter\def\csname res@eff@q3-4b@clef@qps\endcsname{89}
\expandafter\def\csname res@eff@q3-4b@clef@rowms\endcsname{28.3}
\expandafter\def\csname res@eff@q3-4b@clef@step\endcsname{5.85}
\expandafter\def\csname res@eff@q3-4b@cosine@headpct\endcsname{$<$0.001\%}
\expandafter\def\csname res@eff@q3-4b@cosine@p50\endcsname{38.4}
\expandafter\def\csname res@eff@q3-4b@cosine@qps\endcsname{121}
\expandafter\def\csname res@eff@q3-4b@cosine@rowms\endcsname{20.8}
\expandafter\def\csname res@eff@q3-4b@cosine@step\endcsname{5.11}
\expandafter\def\csname res@eff@q3-4b@cosineloop@step\endcsname{5.13}
\expandafter\def\csname res@eff@q3-4b@cosineloop@stepperq\endcsname{10.50}
\expandafter\def\csname res@eff@q3-4b@d3@headpct\endcsname{$<$0.001\%}
\expandafter\def\csname res@eff@q3-4b@d3@p50\endcsname{39.0}
\expandafter\def\csname res@eff@q3-4b@d3@qps\endcsname{78}
\expandafter\def\csname res@eff@q3-4b@d3@rowms\endcsname{32.3}
\expandafter\def\csname res@eff@q3-4b@d3@step\endcsname{9.97}
\expandafter\def\csname res@eff@q3-4b@fastdiff\endcsname{2e-07}
\expandafter\def\csname res@eff@q3-4b@joint@flopx\endcsname{1.00}
\expandafter\def\csname res@eff@q3-4b@joint@p50\endcsname{38.4}
\expandafter\def\csname res@eff@q3-4b@joint@p95\endcsname{54.5}
\expandafter\def\csname res@eff@q3-4b@joint@qps\endcsname{121}
\expandafter\def\csname res@eff@q3-4b@joint@step\endcsname{5.11}
\expandafter\def\csname res@eff@q3-4b@nimble@headpct\endcsname{$<$0.001\%}
\expandafter\def\csname res@eff@q3-4b@nimble@p50\endcsname{37.3}
\expandafter\def\csname res@eff@q3-4b@nimble@qps\endcsname{78}
\expandafter\def\csname res@eff@q3-4b@nimble@rowms\endcsname{32.3}
\expandafter\def\csname res@eff@q3-4b@nimble@step\endcsname{9.94}
\expandafter\def\csname res@eff@q3-4b@perq@flopx\endcsname{1.85}
\expandafter\def\csname res@eff@q3-4b@perq@p50\endcsname{38.6}
\expandafter\def\csname res@eff@q3-4b@perq@p95\endcsname{121.9}
\expandafter\def\csname res@eff@q3-4b@perq@qps\endcsname{75}
\expandafter\def\csname res@eff@q3-4b@perq@step\endcsname{10.56}
\expandafter\def\csname res@eff@q3-4b@sweep13@joint\endcsname{40}
\expandafter\def\csname res@eff@q3-4b@sweep13@perq\endcsname{136}
\expandafter\def\csname res@eff@q3-4b@sweep1@joint\endcsname{38}
\expandafter\def\csname res@eff@q3-4b@sweep1@perq\endcsname{38}
\expandafter\def\csname res@eff@q3-4b@sweep2@joint\endcsname{38}
\expandafter\def\csname res@eff@q3-4b@sweep2@perq\endcsname{42}
\expandafter\def\csname res@eff@q3-4b@sweep4@joint\endcsname{38}
\expandafter\def\csname res@eff@q3-4b@sweep4@perq\endcsname{46}
\expandafter\def\csname res@eff@q3-4b@sweep8@joint\endcsname{39}
\expandafter\def\csname res@eff@q3-4b@sweep8@perq\endcsname{86}
\expandafter\def\csname res@eff@tokperq\endcsname{1.77}
\expandafter\def\csname res@eff@tokperq@helpsteer2\endcsname{2.9}
\expandafter\def\csname res@eff@tokperq@oasst1\endcsname{3.8}
\expandafter\def\csname res@eff@tokq\endcsname{15,165}

\expandafter\def\csname res@R93@ag_news\endcsname{83.5}
\expandafter\def\csname res@R93@banking77\endcsname{92.8}
\expandafter\def\csname res@R93@belebele_en\endcsname{61.6}
\expandafter\def\csname res@R93@boolq\endcsname{84.7}
\expandafter\def\csname res@R93@clinc150\endcsname{92.4}
\expandafter\def\csname res@R93@cosmosqa\endcsname{74.1}
\expandafter\def\csname res@R93@dbpedia14\endcsname{95.8}
\expandafter\def\csname res@R93@emotion\endcsname{92.6}
\expandafter\def\csname res@R93@helpsteer2\endcsname{60.0}
\expandafter\def\csname res@R93@id\endcsname{81.7}
\expandafter\def\csname res@R93@idece\endcsname{0.032}
\expandafter\def\csname res@R93@kev_transfer\endcsname{54.9}
\expandafter\def\csname res@R93@massive_en\endcsname{91.1}
\expandafter\def\csname res@R93@mnli\endcsname{84.0}
\expandafter\def\csname res@R93@oasst1\endcsname{78.9}
\expandafter\def\csname res@R93@ood\endcsname{77.4}
\expandafter\def\csname res@R93@oodece\endcsname{0.074}
\expandafter\def\csname res@R93@peakgb\endcsname{6.3}
\expandafter\def\csname res@R93@procedural_typed_decisions\endcsname{78.3}
\expandafter\def\csname res@R93@qasc\endcsname{91.3}
\expandafter\def\csname res@R93@snli\endcsname{87.7}
\expandafter\def\csname res@R93@squad2\endcsname{83.4}
\expandafter\def\csname res@R93@sst5\endcsname{57.6}
\expandafter\def\csname res@R93@sstep\endcsname{3.32}
\expandafter\def\csname res@R93@thead\endcsname{1.07}
\expandafter\def\csname res@R93@tydiqa\endcsname{91.4}
\expandafter\def\csname res@R93@typed_decisions\endcsname{75.9}
\expandafter\def\csname res@R94@ag_news\endcsname{79.8}
\expandafter\def\csname res@R94@banking77\endcsname{93.1}
\expandafter\def\csname res@R94@belebele_en\endcsname{66.1}
\expandafter\def\csname res@R94@boolq\endcsname{84.1}
\expandafter\def\csname res@R94@clinc150\endcsname{92.1}
\expandafter\def\csname res@R94@cosmosqa\endcsname{74.4}
\expandafter\def\csname res@R94@dbpedia14\endcsname{93.5}
\expandafter\def\csname res@R94@emotion\endcsname{93.0}
\expandafter\def\csname res@R94@helpsteer2\endcsname{59.6}
\expandafter\def\csname res@R94@id\endcsname{81.6}
\expandafter\def\csname res@R94@idece\endcsname{0.033}
\expandafter\def\csname res@R94@kev_transfer\endcsname{56.4}
\expandafter\def\csname res@R94@massive_en\endcsname{91.0}
\expandafter\def\csname res@R94@mnli\endcsname{83.7}
\expandafter\def\csname res@R94@oasst1\endcsname{78.9}
\expandafter\def\csname res@R94@ood\endcsname{75.9}
\expandafter\def\csname res@R94@oodece\endcsname{0.065}
\expandafter\def\csname res@R94@peakgb\endcsname{6.3}
\expandafter\def\csname res@R94@procedural_typed_decisions\endcsname{81.0}
\expandafter\def\csname res@R94@qasc\endcsname{83.6}
\expandafter\def\csname res@R94@snli\endcsname{88.0}
\expandafter\def\csname res@R94@squad2\endcsname{84.2}
\expandafter\def\csname res@R94@sst5\endcsname{54.8}
\expandafter\def\csname res@R94@sstep\endcsname{3.34}
\expandafter\def\csname res@R94@thead\endcsname{1.08}
\expandafter\def\csname res@R94@tydiqa\endcsname{90.3}
\expandafter\def\csname res@R94@typed_decisions\endcsname{76.2}
\expandafter\def\csname res@R97@ag_news\endcsname{83.5}
\expandafter\def\csname res@R97@banking77\endcsname{94.0}
\expandafter\def\csname res@R97@belebele_en\endcsname{87.3}
\expandafter\def\csname res@R97@boolq\endcsname{90.3}
\expandafter\def\csname res@R97@clinc150\endcsname{94.9}
\expandafter\def\csname res@R97@cosmosqa\endcsname{88.0}
\expandafter\def\csname res@R97@dbpedia14\endcsname{97.8}
\expandafter\def\csname res@R97@emotion\endcsname{93.7}
\expandafter\def\csname res@R97@helpsteer2\endcsname{60.8}
\expandafter\def\csname res@R97@id\endcsname{85.9}
\expandafter\def\csname res@R97@idece\endcsname{0.029}
\expandafter\def\csname res@R97@kev_transfer\endcsname{69.7}
\expandafter\def\csname res@R97@massive_en\endcsname{92.2}
\expandafter\def\csname res@R97@mnli\endcsname{89.2}
\expandafter\def\csname res@R97@oasst1\endcsname{79.2}
\expandafter\def\csname res@R97@ood\endcsname{86.9}
\expandafter\def\csname res@R97@oodece\endcsname{0.042}
\expandafter\def\csname res@R97@peakgb\endcsname{21.0}
\expandafter\def\csname res@R97@procedural_typed_decisions\endcsname{89.5}
\expandafter\def\csname res@R97@qasc\endcsname{96.2}
\expandafter\def\csname res@R97@snli\endcsname{91.7}
\expandafter\def\csname res@R97@squad2\endcsname{90.8}
\expandafter\def\csname res@R97@sst5\endcsname{62.1}
\expandafter\def\csname res@R97@sstep\endcsname{6.12}
\expandafter\def\csname res@R97@thead\endcsname{1.07}
\expandafter\def\csname res@R97@tydiqa\endcsname{92.9}
\expandafter\def\csname res@R97@typed_decisions\endcsname{79.6}
\expandafter\def\csname res@R100@ag_news\endcsname{86.0}
\expandafter\def\csname res@R100@banking77\endcsname{94.1}
\expandafter\def\csname res@R100@belebele_en\endcsname{88.0}
\expandafter\def\csname res@R100@boolq\endcsname{90.6}
\expandafter\def\csname res@R100@clinc150\endcsname{94.1}
\expandafter\def\csname res@R100@cosmosqa\endcsname{86.0}
\expandafter\def\csname res@R100@dbpedia14\endcsname{98.0}
\expandafter\def\csname res@R100@emotion\endcsname{93.1}
\expandafter\def\csname res@R100@helpsteer2\endcsname{61.4}
\expandafter\def\csname res@R100@id\endcsname{85.7}
\expandafter\def\csname res@R100@idece\endcsname{0.031}
\expandafter\def\csname res@R100@kev_transfer\endcsname{71.2}
\expandafter\def\csname res@R100@massive_en\endcsname{92.1}
\expandafter\def\csname res@R100@mnli\endcsname{89.3}
\expandafter\def\csname res@R100@oasst1\endcsname{79.1}
\expandafter\def\csname res@R100@ood\endcsname{87.6}
\expandafter\def\csname res@R100@oodece\endcsname{0.037}
\expandafter\def\csname res@R100@peakgb\endcsname{16.9}
\expandafter\def\csname res@R100@procedural_typed_decisions\endcsname{89.9}
\expandafter\def\csname res@R100@qasc\endcsname{94.6}
\expandafter\def\csname res@R100@snli\endcsname{91.6}
\expandafter\def\csname res@R100@squad2\endcsname{90.9}
\expandafter\def\csname res@R100@sst5\endcsname{61.0}
\expandafter\def\csname res@R100@sstep\endcsname{2.77}
\expandafter\def\csname res@R100@thead\endcsname{1.08}
\expandafter\def\csname res@R100@tydiqa\endcsname{93.1}
\expandafter\def\csname res@R100@typed_decisions\endcsname{79.5}
\expandafter\def\csname res@R103@ag_news\endcsname{86.8}
\expandafter\def\csname res@R103@banking77\endcsname{93.0}
\expandafter\def\csname res@R103@belebele_en\endcsname{89.2}
\expandafter\def\csname res@R103@boolq\endcsname{91.6}
\expandafter\def\csname res@R103@clinc150\endcsname{93.7}
\expandafter\def\csname res@R103@cosmosqa\endcsname{89.4}
\expandafter\def\csname res@R103@dbpedia14\endcsname{98.5}
\expandafter\def\csname res@R103@emotion\endcsname{92.5}
\expandafter\def\csname res@R103@helpsteer2\endcsname{60.7}
\expandafter\def\csname res@R103@id\endcsname{85.7}
\expandafter\def\csname res@R103@idece\endcsname{0.025}
\expandafter\def\csname res@R103@kev_transfer\endcsname{74.8}
\expandafter\def\csname res@R103@massive_en\endcsname{91.7}
\expandafter\def\csname res@R103@mnli\endcsname{90.3}
\expandafter\def\csname res@R103@oasst1\endcsname{79.2}
\expandafter\def\csname res@R103@ood\endcsname{89.2}
\expandafter\def\csname res@R103@oodece\endcsname{0.029}
\expandafter\def\csname res@R103@peakgb\endcsname{28.3}
\expandafter\def\csname res@R103@procedural_typed_decisions\endcsname{86.8}
\expandafter\def\csname res@R103@qasc\endcsname{96.9}
\expandafter\def\csname res@R103@snli\endcsname{91.8}
\expandafter\def\csname res@R103@squad2\endcsname{91.2}
\expandafter\def\csname res@R103@sst5\endcsname{63.3}
\expandafter\def\csname res@R103@sstep\endcsname{4.16}
\expandafter\def\csname res@R103@thead\endcsname{1.03}
\expandafter\def\csname res@R103@tydiqa\endcsname{93.3}
\expandafter\def\csname res@R103@typed_decisions\endcsname{76.7}
\expandafter\def\csname res@R106@ag_news\endcsname{86.6}
\expandafter\def\csname res@R106@banking77\endcsname{92.8}
\expandafter\def\csname res@R106@belebele_en\endcsname{87.7}
\expandafter\def\csname res@R106@boolq\endcsname{91.3}
\expandafter\def\csname res@R106@clinc150\endcsname{94.1}
\expandafter\def\csname res@R106@cosmosqa\endcsname{87.6}
\expandafter\def\csname res@R106@dbpedia14\endcsname{98.2}
\expandafter\def\csname res@R106@emotion\endcsname{92.3}
\expandafter\def\csname res@R106@helpsteer2\endcsname{60.9}
\expandafter\def\csname res@R106@id\endcsname{85.7}
\expandafter\def\csname res@R106@idece\endcsname{0.026}
\expandafter\def\csname res@R106@kev_transfer\endcsname{74.7}
\expandafter\def\csname res@R106@massive_en\endcsname{91.8}
\expandafter\def\csname res@R106@mnli\endcsname{90.5}
\expandafter\def\csname res@R106@oasst1\endcsname{79.1}
\expandafter\def\csname res@R106@ood\endcsname{88.7}
\expandafter\def\csname res@R106@oodece\endcsname{0.030}
\expandafter\def\csname res@R106@peakgb\endcsname{28.3}
\expandafter\def\csname res@R106@procedural_typed_decisions\endcsname{87.4}
\expandafter\def\csname res@R106@qasc\endcsname{96.3}
\expandafter\def\csname res@R106@snli\endcsname{92.1}
\expandafter\def\csname res@R106@squad2\endcsname{91.7}
\expandafter\def\csname res@R106@sst5\endcsname{62.1}
\expandafter\def\csname res@R106@sstep\endcsname{4.17}
\expandafter\def\csname res@R106@thead\endcsname{1.01}
\expandafter\def\csname res@R106@tydiqa\endcsname{93.4}
\expandafter\def\csname res@R106@typed_decisions\endcsname{78.6}
\expandafter\def\csname res@R93-R94@ag_news\endcsname{+3.7}
\expandafter\def\csname res@R93-R94@ag_newsp\endcsname{4e-29}
\expandafter\def\csname res@R93-R94@banking77\endcsname{$-$0.3}
\expandafter\def\csname res@R93-R94@banking77p\endcsname{0.4}
\expandafter\def\csname res@R93-R94@belebele_en\endcsname{$-$4.6}
\expandafter\def\csname res@R93-R94@belebele_enp\endcsname{0.003}
\expandafter\def\csname res@R93-R94@boolq\endcsname{+0.6}
\expandafter\def\csname res@R93-R94@boolqp\endcsname{0.4}
\expandafter\def\csname res@R93-R94@clinc150\endcsname{+0.3}
\expandafter\def\csname res@R93-R94@clinc150p\endcsname{0.3}
\expandafter\def\csname res@R93-R94@cosmosqa\endcsname{$-$0.3}
\expandafter\def\csname res@R93-R94@cosmosqap\endcsname{0.8}
\expandafter\def\csname res@R93-R94@dbpedia14\endcsname{+2.3}
\expandafter\def\csname res@R93-R94@dbpedia14p\endcsname{8e-175}
\expandafter\def\csname res@R93-R94@emotion\endcsname{$-$0.4}
\expandafter\def\csname res@R93-R94@emotionp\endcsname{0.5}
\expandafter\def\csname res@R93-R94@helpsteer2\endcsname{+0.3}
\expandafter\def\csname res@R93-R94@helpsteer2p\endcsname{0.4}
\expandafter\def\csname res@R93-R94@id\endcsname{+0.03}
\expandafter\def\csname res@R93-R94@idhi\endcsname{+0.29}
\expandafter\def\csname res@R93-R94@idlo\endcsname{$-$0.24}
\expandafter\def\csname res@R93-R94@kev_transfer\endcsname{$-$1.5}
\expandafter\def\csname res@R93-R94@kev_transferp\endcsname{0.3}
\expandafter\def\csname res@R93-R94@massive_en\endcsname{+0.1}
\expandafter\def\csname res@R93-R94@massive_enp\endcsname{0.9}
\expandafter\def\csname res@R93-R94@mnli\endcsname{+0.3}
\expandafter\def\csname res@R93-R94@mnlip\endcsname{0.3}
\expandafter\def\csname res@R93-R94@oasst1\endcsname{+0.0}
\expandafter\def\csname res@R93-R94@oasst1p\endcsname{0.9}
\expandafter\def\csname res@R93-R94@ood\endcsname{+1.50}
\expandafter\def\csname res@R93-R94@oodhi\endcsname{+2.39}
\expandafter\def\csname res@R93-R94@oodlo\endcsname{+0.63}
\expandafter\def\csname res@R93-R94@procedural_typed_decisions\endcsname{$-$2.7}
\expandafter\def\csname res@R93-R94@procedural_typed_decisionsp\endcsname{7e-73}
\expandafter\def\csname res@R93-R94@qasc\endcsname{+7.7}
\expandafter\def\csname res@R93-R94@qascp\endcsname{5e-12}
\expandafter\def\csname res@R93-R94@snli\endcsname{$-$0.3}
\expandafter\def\csname res@R93-R94@snlip\endcsname{0.3}
\expandafter\def\csname res@R93-R94@squad2\endcsname{$-$0.9}
\expandafter\def\csname res@R93-R94@squad2p\endcsname{0.01}
\expandafter\def\csname res@R93-R94@sst5\endcsname{+2.8}
\expandafter\def\csname res@R93-R94@sst5p\endcsname{0.003}
\expandafter\def\csname res@R93-R94@tydiqa\endcsname{+1.1}
\expandafter\def\csname res@R93-R94@tydiqap\endcsname{0.2}
\expandafter\def\csname res@R93-R94@typed_decisions\endcsname{$-$0.2}
\expandafter\def\csname res@R93-R94@typed_decisionsp\endcsname{0.8}
\expandafter\def\csname res@R97-R100@ag_news\endcsname{$-$2.5}
\expandafter\def\csname res@R97-R100@ag_newsp\endcsname{5e-24}
\expandafter\def\csname res@R97-R100@banking77\endcsname{$-$0.1}
\expandafter\def\csname res@R97-R100@banking77p\endcsname{0.8}
\expandafter\def\csname res@R97-R100@belebele_en\endcsname{$-$0.7}
\expandafter\def\csname res@R97-R100@belebele_enp\endcsname{0.6}
\expandafter\def\csname res@R97-R100@boolq\endcsname{$-$0.3}
\expandafter\def\csname res@R97-R100@boolqp\endcsname{0.5}
\expandafter\def\csname res@R97-R100@clinc150\endcsname{+0.8}
\expandafter\def\csname res@R97-R100@clinc150p\endcsname{0.001}
\expandafter\def\csname res@R97-R100@cosmosqa\endcsname{+2.0}
\expandafter\def\csname res@R97-R100@cosmosqap\endcsname{0.003}
\expandafter\def\csname res@R97-R100@dbpedia14\endcsname{$-$0.1}
\expandafter\def\csname res@R97-R100@dbpedia14p\endcsname{0.001}
\expandafter\def\csname res@R97-R100@emotion\endcsname{+0.6}
\expandafter\def\csname res@R97-R100@emotionp\endcsname{0.2}
\expandafter\def\csname res@R97-R100@helpsteer2\endcsname{$-$0.6}
\expandafter\def\csname res@R97-R100@helpsteer2p\endcsname{0.2}
\expandafter\def\csname res@R97-R100@id\endcsname{+0.20}
\expandafter\def\csname res@R97-R100@idhi\endcsname{+0.42}
\expandafter\def\csname res@R97-R100@idlo\endcsname{$-$0.03}
\expandafter\def\csname res@R97-R100@kev_transfer\endcsname{$-$1.6}
\expandafter\def\csname res@R97-R100@kev_transferp\endcsname{0.2}
\expandafter\def\csname res@R97-R100@massive_en\endcsname{+0.1}
\expandafter\def\csname res@R97-R100@massive_enp\endcsname{0.9}
\expandafter\def\csname res@R97-R100@mnli\endcsname{$-$0.2}
\expandafter\def\csname res@R97-R100@mnlip\endcsname{0.5}
\expandafter\def\csname res@R97-R100@oasst1\endcsname{+0.1}
\expandafter\def\csname res@R97-R100@oasst1p\endcsname{0.7}
\expandafter\def\csname res@R97-R100@ood\endcsname{$-$0.66}
\expandafter\def\csname res@R97-R100@oodhi\endcsname{$-$0.03}
\expandafter\def\csname res@R97-R100@oodlo\endcsname{$-$1.32}
\expandafter\def\csname res@R97-R100@procedural_typed_decisions\endcsname{$-$0.4}
\expandafter\def\csname res@R97-R100@procedural_typed_decisionsp\endcsname{0.0003}
\expandafter\def\csname res@R97-R100@qasc\endcsname{+1.6}
\expandafter\def\csname res@R97-R100@qascp\endcsname{0.02}
\expandafter\def\csname res@R97-R100@snli\endcsname{+0.1}
\expandafter\def\csname res@R97-R100@snlip\endcsname{0.5}
\expandafter\def\csname res@R97-R100@squad2\endcsname{$-$0.1}
\expandafter\def\csname res@R97-R100@squad2p\endcsname{0.8}
\expandafter\def\csname res@R97-R100@sst5\endcsname{+1.1}
\expandafter\def\csname res@R97-R100@sst5p\endcsname{0.2}
\expandafter\def\csname res@R97-R100@tydiqa\endcsname{$-$0.2}
\expandafter\def\csname res@R97-R100@tydiqap\endcsname{0.9}
\expandafter\def\csname res@R97-R100@typed_decisions\endcsname{+0.1}
\expandafter\def\csname res@R97-R100@typed_decisionsp\endcsname{0.9}
\expandafter\def\csname res@R103-R106@ag_news\endcsname{+0.3}
\expandafter\def\csname res@R103-R106@ag_newsp\endcsname{0.3}
\expandafter\def\csname res@R103-R106@banking77\endcsname{+0.2}
\expandafter\def\csname res@R103-R106@banking77p\endcsname{0.6}
\expandafter\def\csname res@R103-R106@belebele_en\endcsname{+1.6}
\expandafter\def\csname res@R103-R106@belebele_enp\endcsname{0.1}
\expandafter\def\csname res@R103-R106@boolq\endcsname{+0.3}
\expandafter\def\csname res@R103-R106@boolqp\endcsname{0.5}
\expandafter\def\csname res@R103-R106@clinc150\endcsname{$-$0.3}
\expandafter\def\csname res@R103-R106@clinc150p\endcsname{0.1}
\expandafter\def\csname res@R103-R106@cosmosqa\endcsname{+1.8}
\expandafter\def\csname res@R103-R106@cosmosqap\endcsname{0.002}
\expandafter\def\csname res@R103-R106@dbpedia14\endcsname{+0.3}
\expandafter\def\csname res@R103-R106@dbpedia14p\endcsname{4e-13}
\expandafter\def\csname res@R103-R106@emotion\endcsname{+0.2}
\expandafter\def\csname res@R103-R106@emotionp\endcsname{0.7}
\expandafter\def\csname res@R103-R106@helpsteer2\endcsname{$-$0.2}
\expandafter\def\csname res@R103-R106@helpsteer2p\endcsname{0.6}
\expandafter\def\csname res@R103-R106@id\endcsname{$-$0.03}
\expandafter\def\csname res@R103-R106@idhi\endcsname{+0.19}
\expandafter\def\csname res@R103-R106@idlo\endcsname{$-$0.22}
\expandafter\def\csname res@R103-R106@kev_transfer\endcsname{+0.1}
\expandafter\def\csname res@R103-R106@kev_transferp\endcsname{1}
\expandafter\def\csname res@R103-R106@massive_en\endcsname{$-$0.1}
\expandafter\def\csname res@R103-R106@massive_enp\endcsname{0.9}
\expandafter\def\csname res@R103-R106@mnli\endcsname{$-$0.1}
\expandafter\def\csname res@R103-R106@mnlip\endcsname{0.6}
\expandafter\def\csname res@R103-R106@oasst1\endcsname{+0.1}
\expandafter\def\csname res@R103-R106@oasst1p\endcsname{0.6}
\expandafter\def\csname res@R103-R106@ood\endcsname{+0.54}
\expandafter\def\csname res@R103-R106@oodhi\endcsname{+1.16}
\expandafter\def\csname res@R103-R106@oodlo\endcsname{$-$0.09}
\expandafter\def\csname res@R103-R106@procedural_typed_decisions\endcsname{$-$0.6}
\expandafter\def\csname res@R103-R106@procedural_typed_decisionsp\endcsname{1e-06}
\expandafter\def\csname res@R103-R106@qasc\endcsname{+0.5}
\expandafter\def\csname res@R103-R106@qascp\endcsname{0.4}
\expandafter\def\csname res@R103-R106@snli\endcsname{$-$0.3}
\expandafter\def\csname res@R103-R106@snlip\endcsname{0.09}
\expandafter\def\csname res@R103-R106@squad2\endcsname{$-$0.5}
\expandafter\def\csname res@R103-R106@squad2p\endcsname{0.03}
\expandafter\def\csname res@R103-R106@sst5\endcsname{+1.2}
\expandafter\def\csname res@R103-R106@sst5p\endcsname{0.08}
\expandafter\def\csname res@R103-R106@tydiqa\endcsname{$-$0.1}
\expandafter\def\csname res@R103-R106@tydiqap\endcsname{1}
\expandafter\def\csname res@R103-R106@typed_decisions\endcsname{$-$1.9}
\expandafter\def\csname res@R103-R106@typed_decisionsp\endcsname{0.005}

\expandafter\def\csname res@R141@ag_news\endcsname{84.1}
\expandafter\def\csname res@R141@banking77\endcsname{94.2}
\expandafter\def\csname res@R141@belebele_en\endcsname{89.8}
\expandafter\def\csname res@R141@boolq\endcsname{91.1}
\expandafter\def\csname res@R141@clinc150\endcsname{94.5}
\expandafter\def\csname res@R141@cosmosqa\endcsname{88.1}
\expandafter\def\csname res@R141@dbpedia14\endcsname{96.7}
\expandafter\def\csname res@R141@emotion\endcsname{93.6}
\expandafter\def\csname res@R141@helpsteer2\endcsname{60.7}
\expandafter\def\csname res@R141@id\endcsname{86.1}
\expandafter\def\csname res@R141@idece\endcsname{0.031}
\expandafter\def\csname res@R141@kev_transfer\endcsname{69.9}
\expandafter\def\csname res@R141@massive_en\endcsname{92.4}
\expandafter\def\csname res@R141@mnli\endcsname{89.4}
\expandafter\def\csname res@R141@oasst1\endcsname{79.2}
\expandafter\def\csname res@R141@ood\endcsname{86.7}
\expandafter\def\csname res@R141@oodece\endcsname{0.040}
\expandafter\def\csname res@R141@peakgb\endcsname{19.5}
\expandafter\def\csname res@R141@procedural_typed_decisions\endcsname{90.8}
\expandafter\def\csname res@R141@qasc\endcsname{92.9}
\expandafter\def\csname res@R141@snli\endcsname{91.7}
\expandafter\def\csname res@R141@squad2\endcsname{91.6}
\expandafter\def\csname res@R141@sst5\endcsname{61.2}
\expandafter\def\csname res@R141@sstep\endcsname{3.91}
\expandafter\def\csname res@R141@thead\endcsname{1.10}
\expandafter\def\csname res@R141@tydiqa\endcsname{94.0}
\expandafter\def\csname res@R141@typed_decisions\endcsname{79.0}
\expandafter\def\csname res@R141@valnll\endcsname{0.304}
\expandafter\def\csname res@R142@ag_news\endcsname{83.1}
\expandafter\def\csname res@R142@banking77\endcsname{93.8}
\expandafter\def\csname res@R142@belebele_en\endcsname{87.8}
\expandafter\def\csname res@R142@boolq\endcsname{90.7}
\expandafter\def\csname res@R142@clinc150\endcsname{94.8}
\expandafter\def\csname res@R142@cosmosqa\endcsname{87.3}
\expandafter\def\csname res@R142@dbpedia14\endcsname{97.1}
\expandafter\def\csname res@R142@emotion\endcsname{93.2}
\expandafter\def\csname res@R142@helpsteer2\endcsname{60.8}
\expandafter\def\csname res@R142@id\endcsname{85.9}
\expandafter\def\csname res@R142@idece\endcsname{0.032}
\expandafter\def\csname res@R142@kev_transfer\endcsname{67.9}
\expandafter\def\csname res@R142@massive_en\endcsname{92.2}
\expandafter\def\csname res@R142@mnli\endcsname{89.6}
\expandafter\def\csname res@R142@oasst1\endcsname{79.4}
\expandafter\def\csname res@R142@ood\endcsname{86.1}
\expandafter\def\csname res@R142@oodece\endcsname{0.043}
\expandafter\def\csname res@R142@peakgb\endcsname{21.2}
\expandafter\def\csname res@R142@procedural_typed_decisions\endcsname{91.0}
\expandafter\def\csname res@R142@qasc\endcsname{94.7}
\expandafter\def\csname res@R142@snli\endcsname{91.6}
\expandafter\def\csname res@R142@squad2\endcsname{91.3}
\expandafter\def\csname res@R142@sst5\endcsname{61.0}
\expandafter\def\csname res@R142@sstep\endcsname{5.08}
\expandafter\def\csname res@R142@thead\endcsname{1.11}
\expandafter\def\csname res@R142@tydiqa\endcsname{93.4}
\expandafter\def\csname res@R142@typed_decisions\endcsname{79.0}
\expandafter\def\csname res@R142@valnll\endcsname{0.301}
\expandafter\def\csname res@R143@ag_news\endcsname{82.1}
\expandafter\def\csname res@R143@banking77\endcsname{93.1}
\expandafter\def\csname res@R143@belebele_en\endcsname{87.0}
\expandafter\def\csname res@R143@boolq\endcsname{92.4}
\expandafter\def\csname res@R143@clinc150\endcsname{94.1}
\expandafter\def\csname res@R143@cosmosqa\endcsname{89.9}
\expandafter\def\csname res@R143@dbpedia14\endcsname{98.1}
\expandafter\def\csname res@R143@emotion\endcsname{93.5}
\expandafter\def\csname res@R143@helpsteer2\endcsname{61.0}
\expandafter\def\csname res@R143@id\endcsname{86.4}
\expandafter\def\csname res@R143@idece\endcsname{0.026}
\expandafter\def\csname res@R143@kev_transfer\endcsname{75.8}
\expandafter\def\csname res@R143@massive_en\endcsname{91.9}
\expandafter\def\csname res@R143@mnli\endcsname{90.0}
\expandafter\def\csname res@R143@oasst1\endcsname{79.1}
\expandafter\def\csname res@R143@ood\endcsname{87.6}
\expandafter\def\csname res@R143@oodece\endcsname{0.037}
\expandafter\def\csname res@R143@peakgb\endcsname{35.3}
\expandafter\def\csname res@R143@procedural_typed_decisions\endcsname{90.1}
\expandafter\def\csname res@R143@qasc\endcsname{94.7}
\expandafter\def\csname res@R143@snli\endcsname{92.2}
\expandafter\def\csname res@R143@squad2\endcsname{92.4}
\expandafter\def\csname res@R143@sst5\endcsname{63.1}
\expandafter\def\csname res@R143@sstep\endcsname{6.34}
\expandafter\def\csname res@R143@thead\endcsname{1.05}
\expandafter\def\csname res@R143@tydiqa\endcsname{93.9}
\expandafter\def\csname res@R143@typed_decisions\endcsname{79.7}
\expandafter\def\csname res@R143@valnll\endcsname{0.307}
\expandafter\def\csname res@R144b@ag_news\endcsname{82.5}
\expandafter\def\csname res@R144b@banking77\endcsname{92.3}
\expandafter\def\csname res@R144b@belebele_en\endcsname{88.0}
\expandafter\def\csname res@R144b@boolq\endcsname{91.8}
\expandafter\def\csname res@R144b@clinc150\endcsname{93.7}
\expandafter\def\csname res@R144b@cosmosqa\endcsname{89.2}
\expandafter\def\csname res@R144b@dbpedia14\endcsname{98.3}
\expandafter\def\csname res@R144b@emotion\endcsname{91.5}
\expandafter\def\csname res@R144b@helpsteer2\endcsname{61.3}
\expandafter\def\csname res@R144b@id\endcsname{85.9}
\expandafter\def\csname res@R144b@idece\endcsname{0.026}
\expandafter\def\csname res@R144b@kev_transfer\endcsname{73.6}
\expandafter\def\csname res@R144b@massive_en\endcsname{91.3}
\expandafter\def\csname res@R144b@mnli\endcsname{90.2}
\expandafter\def\csname res@R144b@oasst1\endcsname{79.2}
\expandafter\def\csname res@R144b@ood\endcsname{87.8}
\expandafter\def\csname res@R144b@oodece\endcsname{0.043}
\expandafter\def\csname res@R144b@peakgb\endcsname{37.0}
\expandafter\def\csname res@R144b@procedural_typed_decisions\endcsname{89.4}
\expandafter\def\csname res@R144b@qasc\endcsname{96.7}
\expandafter\def\csname res@R144b@snli\endcsname{92.4}
\expandafter\def\csname res@R144b@squad2\endcsname{91.7}
\expandafter\def\csname res@R144b@sst5\endcsname{62.8}
\expandafter\def\csname res@R144b@sstep\endcsname{7.69}
\expandafter\def\csname res@R144b@thead\endcsname{1.07}
\expandafter\def\csname res@R144b@tydiqa\endcsname{93.9}
\expandafter\def\csname res@R144b@typed_decisions\endcsname{78.3}
\expandafter\def\csname res@R144b@valnll\endcsname{0.322}
\expandafter\def\csname res@R145@ag_news\endcsname{84.6}
\expandafter\def\csname res@R145@banking77\endcsname{93.8}
\expandafter\def\csname res@R145@belebele_en\endcsname{90.7}
\expandafter\def\csname res@R145@boolq\endcsname{91.9}
\expandafter\def\csname res@R145@clinc150\endcsname{94.3}
\expandafter\def\csname res@R145@cosmosqa\endcsname{91.3}
\expandafter\def\csname res@R145@dbpedia14\endcsname{98.5}
\expandafter\def\csname res@R145@emotion\endcsname{93.2}
\expandafter\def\csname res@R145@helpsteer2\endcsname{61.0}
\expandafter\def\csname res@R145@id\endcsname{86.8}
\expandafter\def\csname res@R145@idece\endcsname{0.026}
\expandafter\def\csname res@R145@kev_transfer\endcsname{75.2}
\expandafter\def\csname res@R145@massive_en\endcsname{92.1}
\expandafter\def\csname res@R145@mnli\endcsname{90.4}
\expandafter\def\csname res@R145@oasst1\endcsname{79.0}
\expandafter\def\csname res@R145@ood\endcsname{89.1}
\expandafter\def\csname res@R145@oodece\endcsname{0.038}
\expandafter\def\csname res@R145@peakgb\endcsname{50.0}
\expandafter\def\csname res@R145@procedural_typed_decisions\endcsname{91.4}
\expandafter\def\csname res@R145@qasc\endcsname{96.4}
\expandafter\def\csname res@R145@snli\endcsname{92.5}
\expandafter\def\csname res@R145@squad2\endcsname{92.8}
\expandafter\def\csname res@R145@sst5\endcsname{63.8}
\expandafter\def\csname res@R145@sstep\endcsname{8.48}
\expandafter\def\csname res@R145@thead\endcsname{1.05}
\expandafter\def\csname res@R145@tydiqa\endcsname{93.8}
\expandafter\def\csname res@R145@typed_decisions\endcsname{80.2}
\expandafter\def\csname res@R145@valnll\endcsname{0.286}
\expandafter\def\csname res@R146@ag_news\endcsname{83.6}
\expandafter\def\csname res@R146@banking77\endcsname{93.7}
\expandafter\def\csname res@R146@belebele_en\endcsname{92.2}
\expandafter\def\csname res@R146@boolq\endcsname{92.4}
\expandafter\def\csname res@R146@clinc150\endcsname{94.2}
\expandafter\def\csname res@R146@cosmosqa\endcsname{91.3}
\expandafter\def\csname res@R146@dbpedia14\endcsname{98.3}
\expandafter\def\csname res@R146@emotion\endcsname{93.0}
\expandafter\def\csname res@R146@helpsteer2\endcsname{60.9}
\expandafter\def\csname res@R146@id\endcsname{86.7}
\expandafter\def\csname res@R146@idece\endcsname{0.026}
\expandafter\def\csname res@R146@kev_transfer\endcsname{74.2}
\expandafter\def\csname res@R146@massive_en\endcsname{92.1}
\expandafter\def\csname res@R146@mnli\endcsname{90.5}
\expandafter\def\csname res@R146@oasst1\endcsname{79.3}
\expandafter\def\csname res@R146@ood\endcsname{89.0}
\expandafter\def\csname res@R146@oodece\endcsname{0.037}
\expandafter\def\csname res@R146@peakgb\endcsname{51.9}
\expandafter\def\csname res@R146@procedural_typed_decisions\endcsname{91.1}
\expandafter\def\csname res@R146@qasc\endcsname{96.8}
\expandafter\def\csname res@R146@snli\endcsname{92.4}
\expandafter\def\csname res@R146@squad2\endcsname{93.1}
\expandafter\def\csname res@R146@sst5\endcsname{63.1}
\expandafter\def\csname res@R146@sstep\endcsname{9.82}
\expandafter\def\csname res@R146@thead\endcsname{1.06}
\expandafter\def\csname res@R146@tydiqa\endcsname{93.7}
\expandafter\def\csname res@R146@typed_decisions\endcsname{80.1}
\expandafter\def\csname res@R146@valnll\endcsname{0.296}
\expandafter\def\csname res@R142-R141@id\endcsname{$-$0.16}
\expandafter\def\csname res@R142-R141@idlo\endcsname{$-$0.37}
\expandafter\def\csname res@R142-R141@idhi\endcsname{+0.03}
\expandafter\def\csname res@R142-R141@ood\endcsname{$-$0.54}
\expandafter\def\csname res@R142-R141@oodlo\endcsname{$-$1.12}
\expandafter\def\csname res@R142-R141@oodhi\endcsname{+0.00}
\expandafter\def\csname res@R142-R141@ag_news\endcsname{$-$1.0}
\expandafter\def\csname res@R142-R141@ag_newsp\endcsname{6e-05}
\expandafter\def\csname res@R142-R141@banking77\endcsname{$-$0.4}
\expandafter\def\csname res@R142-R141@banking77p\endcsname{0.2}
\expandafter\def\csname res@R142-R141@belebele_en\endcsname{$-$2.0}
\expandafter\def\csname res@R142-R141@belebele_enp\endcsname{0.02}
\expandafter\def\csname res@R142-R141@boolq\endcsname{$-$0.5}
\expandafter\def\csname res@R142-R141@boolqp\endcsname{0.3}
\expandafter\def\csname res@R142-R141@clinc150\endcsname{+0.3}
\expandafter\def\csname res@R142-R141@clinc150p\endcsname{0.2}
\expandafter\def\csname res@R142-R141@cosmosqa\endcsname{$-$0.9}
\expandafter\def\csname res@R142-R141@cosmosqap\endcsname{0.2}
\expandafter\def\csname res@R142-R141@dbpedia14\endcsname{+0.4}
\expandafter\def\csname res@R142-R141@dbpedia14p\endcsname{1e-24}
\expandafter\def\csname res@R142-R141@emotion\endcsname{$-$0.4}
\expandafter\def\csname res@R142-R141@emotionp\endcsname{0.4}
\expandafter\def\csname res@R142-R141@helpsteer2\endcsname{+0.1}
\expandafter\def\csname res@R142-R141@helpsteer2p\endcsname{0.9}
\expandafter\def\csname res@R142-R141@kev_transfer\endcsname{$-$2.0}
\expandafter\def\csname res@R142-R141@kev_transferp\endcsname{0.04}
\expandafter\def\csname res@R142-R141@massive_en\endcsname{$-$0.1}
\expandafter\def\csname res@R142-R141@massive_enp\endcsname{0.5}
\expandafter\def\csname res@R142-R141@mnli\endcsname{+0.2}
\expandafter\def\csname res@R142-R141@mnlip\endcsname{0.5}
\expandafter\def\csname res@R142-R141@oasst1\endcsname{+0.2}
\expandafter\def\csname res@R142-R141@oasst1p\endcsname{0.1}
\expandafter\def\csname res@R142-R141@procedural_typed_decisions\endcsname{+0.2}
\expandafter\def\csname res@R142-R141@procedural_typed_decisionsp\endcsname{0.1}
\expandafter\def\csname res@R142-R141@qasc\endcsname{+1.8}
\expandafter\def\csname res@R142-R141@qascp\endcsname{0.003}
\expandafter\def\csname res@R142-R141@snli\endcsname{$-$0.1}
\expandafter\def\csname res@R142-R141@snlip\endcsname{0.6}
\expandafter\def\csname res@R142-R141@squad2\endcsname{$-$0.2}
\expandafter\def\csname res@R142-R141@squad2p\endcsname{0.3}
\expandafter\def\csname res@R142-R141@sst5\endcsname{$-$0.2}
\expandafter\def\csname res@R142-R141@sst5p\endcsname{0.8}
\expandafter\def\csname res@R142-R141@tydiqa\endcsname{$-$0.6}
\expandafter\def\csname res@R142-R141@tydiqap\endcsname{0.3}
\expandafter\def\csname res@R142-R141@typed_decisions\endcsname{+0.0}
\expandafter\def\csname res@R142-R141@typed_decisionsp\endcsname{1}
\expandafter\def\csname res@R144b-R143@id\endcsname{$-$0.47}
\expandafter\def\csname res@R144b-R143@idlo\endcsname{$-$0.67}
\expandafter\def\csname res@R144b-R143@idhi\endcsname{$-$0.27}
\expandafter\def\csname res@R144b-R143@ood\endcsname{+0.26}
\expandafter\def\csname res@R144b-R143@oodlo\endcsname{$-$0.29}
\expandafter\def\csname res@R144b-R143@oodhi\endcsname{+0.82}
\expandafter\def\csname res@R144b-R143@ag_news\endcsname{+0.4}
\expandafter\def\csname res@R144b-R143@ag_newsp\endcsname{0.1}
\expandafter\def\csname res@R144b-R143@banking77\endcsname{$-$0.8}
\expandafter\def\csname res@R144b-R143@banking77p\endcsname{0.02}
\expandafter\def\csname res@R144b-R143@belebele_en\endcsname{+1.0}
\expandafter\def\csname res@R144b-R143@belebele_enp\endcsname{0.2}
\expandafter\def\csname res@R144b-R143@boolq\endcsname{$-$0.5}
\expandafter\def\csname res@R144b-R143@boolqp\endcsname{0.2}
\expandafter\def\csname res@R144b-R143@clinc150\endcsname{$-$0.4}
\expandafter\def\csname res@R144b-R143@clinc150p\endcsname{0.1}
\expandafter\def\csname res@R144b-R143@cosmosqa\endcsname{$-$0.6}
\expandafter\def\csname res@R144b-R143@cosmosqap\endcsname{0.3}
\expandafter\def\csname res@R144b-R143@dbpedia14\endcsname{+0.2}
\expandafter\def\csname res@R144b-R143@dbpedia14p\endcsname{3e-10}
\expandafter\def\csname res@R144b-R143@emotion\endcsname{$-$2.0}
\expandafter\def\csname res@R144b-R143@emotionp\endcsname{0.0003}
\expandafter\def\csname res@R144b-R143@helpsteer2\endcsname{+0.3}
\expandafter\def\csname res@R144b-R143@helpsteer2p\endcsname{0.5}
\expandafter\def\csname res@R144b-R143@kev_transfer\endcsname{$-$2.2}
\expandafter\def\csname res@R144b-R143@kev_transferp\endcsname{0.02}
\expandafter\def\csname res@R144b-R143@massive_en\endcsname{$-$0.6}
\expandafter\def\csname res@R144b-R143@massive_enp\endcsname{0.02}
\expandafter\def\csname res@R144b-R143@mnli\endcsname{+0.2}
\expandafter\def\csname res@R144b-R143@mnlip\endcsname{0.3}
\expandafter\def\csname res@R144b-R143@oasst1\endcsname{+0.1}
\expandafter\def\csname res@R144b-R143@oasst1p\endcsname{0.8}
\expandafter\def\csname res@R144b-R143@procedural_typed_decisions\endcsname{$-$0.7}
\expandafter\def\csname res@R144b-R143@procedural_typed_decisionsp\endcsname{8e-09}
\expandafter\def\csname res@R144b-R143@qasc\endcsname{+1.9}
\expandafter\def\csname res@R144b-R143@qascp\endcsname{0.0009}
\expandafter\def\csname res@R144b-R143@snli\endcsname{+0.2}
\expandafter\def\csname res@R144b-R143@snlip\endcsname{0.4}
\expandafter\def\csname res@R144b-R143@squad2\endcsname{$-$0.7}
\expandafter\def\csname res@R144b-R143@squad2p\endcsname{0.0005}
\expandafter\def\csname res@R144b-R143@sst5\endcsname{$-$0.3}
\expandafter\def\csname res@R144b-R143@sst5p\endcsname{0.7}
\expandafter\def\csname res@R144b-R143@tydiqa\endcsname{+0.0}
\expandafter\def\csname res@R144b-R143@tydiqap\endcsname{1}
\expandafter\def\csname res@R144b-R143@typed_decisions\endcsname{$-$1.3}
\expandafter\def\csname res@R144b-R143@typed_decisionsp\endcsname{0.05}
\expandafter\def\csname res@R146-R145@id\endcsname{$-$0.05}
\expandafter\def\csname res@R146-R145@idlo\endcsname{$-$0.22}
\expandafter\def\csname res@R146-R145@idhi\endcsname{+0.12}
\expandafter\def\csname res@R146-R145@ood\endcsname{$-$0.08}
\expandafter\def\csname res@R146-R145@oodlo\endcsname{$-$0.55}
\expandafter\def\csname res@R146-R145@oodhi\endcsname{+0.38}
\expandafter\def\csname res@R146-R145@ag_news\endcsname{$-$1.0}
\expandafter\def\csname res@R146-R145@ag_newsp\endcsname{5e-06}
\expandafter\def\csname res@R146-R145@banking77\endcsname{$-$0.1}
\expandafter\def\csname res@R146-R145@banking77p\endcsname{0.7}
\expandafter\def\csname res@R146-R145@belebele_en\endcsname{+1.6}
\expandafter\def\csname res@R146-R145@belebele_enp\endcsname{0.03}
\expandafter\def\csname res@R146-R145@boolq\endcsname{+0.5}
\expandafter\def\csname res@R146-R145@boolqp\endcsname{0.1}
\expandafter\def\csname res@R146-R145@clinc150\endcsname{$-$0.1}
\expandafter\def\csname res@R146-R145@clinc150p\endcsname{0.6}
\expandafter\def\csname res@R146-R145@cosmosqa\endcsname{$-$0.0}
\expandafter\def\csname res@R146-R145@cosmosqap\endcsname{1}
\expandafter\def\csname res@R146-R145@dbpedia14\endcsname{$-$0.2}
\expandafter\def\csname res@R146-R145@dbpedia14p\endcsname{5e-17}
\expandafter\def\csname res@R146-R145@emotion\endcsname{$-$0.1}
\expandafter\def\csname res@R146-R145@emotionp\endcsname{0.8}
\expandafter\def\csname res@R146-R145@helpsteer2\endcsname{$-$0.1}
\expandafter\def\csname res@R146-R145@helpsteer2p\endcsname{0.9}
\expandafter\def\csname res@R146-R145@kev_transfer\endcsname{$-$1.1}
\expandafter\def\csname res@R146-R145@kev_transferp\endcsname{0.3}
\expandafter\def\csname res@R146-R145@massive_en\endcsname{$-$0.0}
\expandafter\def\csname res@R146-R145@massive_enp\endcsname{1}
\expandafter\def\csname res@R146-R145@mnli\endcsname{+0.1}
\expandafter\def\csname res@R146-R145@mnlip\endcsname{0.8}
\expandafter\def\csname res@R146-R145@oasst1\endcsname{+0.3}
\expandafter\def\csname res@R146-R145@oasst1p\endcsname{0.02}
\expandafter\def\csname res@R146-R145@procedural_typed_decisions\endcsname{$-$0.3}
\expandafter\def\csname res@R146-R145@procedural_typed_decisionsp\endcsname{0.002}
\expandafter\def\csname res@R146-R145@qasc\endcsname{+0.3}
\expandafter\def\csname res@R146-R145@qascp\endcsname{0.6}
\expandafter\def\csname res@R146-R145@snli\endcsname{$-$0.1}
\expandafter\def\csname res@R146-R145@snlip\endcsname{0.6}
\expandafter\def\csname res@R146-R145@squad2\endcsname{+0.3}
\expandafter\def\csname res@R146-R145@squad2p\endcsname{0.06}
\expandafter\def\csname res@R146-R145@sst5\endcsname{$-$0.7}
\expandafter\def\csname res@R146-R145@sst5p\endcsname{0.2}
\expandafter\def\csname res@R146-R145@tydiqa\endcsname{$-$0.1}
\expandafter\def\csname res@R146-R145@tydiqap\endcsname{1}
\expandafter\def\csname res@R146-R145@typed_decisions\endcsname{$-$0.1}
\expandafter\def\csname res@R146-R145@typed_decisionsp\endcsname{0.9}

\expandafter\def\csname res@R117@ag_news\endcsname{82.4}
\expandafter\def\csname res@R117@banking77\endcsname{92.9}
\expandafter\def\csname res@R117@belebele_en\endcsname{66.2}
\expandafter\def\csname res@R117@boolq\endcsname{84.1}
\expandafter\def\csname res@R117@clinc150\endcsname{92.4}
\expandafter\def\csname res@R117@cosmosqa\endcsname{74.4}
\expandafter\def\csname res@R117@dbpedia14\endcsname{93.6}
\expandafter\def\csname res@R117@emotion\endcsname{92.5}
\expandafter\def\csname res@R117@helpsteer2\endcsname{60.0}
\expandafter\def\csname res@R117@id\endcsname{81.8}
\expandafter\def\csname res@R117@idece\endcsname{0.032}
\expandafter\def\csname res@R117@kev_transfer\endcsname{56.0}
\expandafter\def\csname res@R117@massive_en\endcsname{91.2}
\expandafter\def\csname res@R117@mnli\endcsname{84.2}
\expandafter\def\csname res@R117@oasst1\endcsname{78.8}
\expandafter\def\csname res@R117@ood\endcsname{76.8}
\expandafter\def\csname res@R117@oodece\endcsname{0.066}
\expandafter\def\csname res@R117@peakgb\endcsname{6.3}
\expandafter\def\csname res@R117@procedural_typed_decisions\endcsname{81.2}
\expandafter\def\csname res@R117@qasc\endcsname{86.0}
\expandafter\def\csname res@R117@snli\endcsname{87.8}
\expandafter\def\csname res@R117@squad2\endcsname{84.7}
\expandafter\def\csname res@R117@sst5\endcsname{56.7}
\expandafter\def\csname res@R117@sstep\endcsname{3.07}
\expandafter\def\csname res@R117@thead\endcsname{1.08}
\expandafter\def\csname res@R117@tydiqa\endcsname{91.3}
\expandafter\def\csname res@R117@typed_decisions\endcsname{76.8}
\expandafter\def\csname res@R117@valnll\endcsname{0.455}
\expandafter\def\csname res@R118@ag_news\endcsname{83.6}
\expandafter\def\csname res@R118@banking77\endcsname{93.0}
\expandafter\def\csname res@R118@belebele_en\endcsname{68.2}
\expandafter\def\csname res@R118@boolq\endcsname{83.9}
\expandafter\def\csname res@R118@clinc150\endcsname{92.6}
\expandafter\def\csname res@R118@cosmosqa\endcsname{74.1}
\expandafter\def\csname res@R118@dbpedia14\endcsname{93.5}
\expandafter\def\csname res@R118@emotion\endcsname{93.2}
\expandafter\def\csname res@R118@helpsteer2\endcsname{60.4}
\expandafter\def\csname res@R118@id\endcsname{81.9}
\expandafter\def\csname res@R118@idece\endcsname{0.033}
\expandafter\def\csname res@R118@kev_transfer\endcsname{55.9}
\expandafter\def\csname res@R118@massive_en\endcsname{91.0}
\expandafter\def\csname res@R118@mnli\endcsname{84.0}
\expandafter\def\csname res@R118@oasst1\endcsname{78.9}
\expandafter\def\csname res@R118@ood\endcsname{77.3}
\expandafter\def\csname res@R118@oodece\endcsname{0.060}
\expandafter\def\csname res@R118@peakgb\endcsname{6.3}
\expandafter\def\csname res@R118@procedural_typed_decisions\endcsname{81.3}
\expandafter\def\csname res@R118@qasc\endcsname{85.4}
\expandafter\def\csname res@R118@snli\endcsname{87.6}
\expandafter\def\csname res@R118@squad2\endcsname{84.8}
\expandafter\def\csname res@R118@sst5\endcsname{56.2}
\expandafter\def\csname res@R118@sstep\endcsname{3.11}
\expandafter\def\csname res@R118@thead\endcsname{1.08}
\expandafter\def\csname res@R118@tydiqa\endcsname{91.5}
\expandafter\def\csname res@R118@typed_decisions\endcsname{76.6}
\expandafter\def\csname res@R118@valnll\endcsname{0.448}
\expandafter\def\csname res@R119@ag_news\endcsname{80.4}
\expandafter\def\csname res@R119@banking77\endcsname{93.1}
\expandafter\def\csname res@R119@belebele_en\endcsname{67.2}
\expandafter\def\csname res@R119@boolq\endcsname{83.9}
\expandafter\def\csname res@R119@clinc150\endcsname{92.1}
\expandafter\def\csname res@R119@cosmosqa\endcsname{74.2}
\expandafter\def\csname res@R119@dbpedia14\endcsname{94.0}
\expandafter\def\csname res@R119@emotion\endcsname{93.1}
\expandafter\def\csname res@R119@helpsteer2\endcsname{60.2}
\expandafter\def\csname res@R119@id\endcsname{82.0}
\expandafter\def\csname res@R119@idece\endcsname{0.033}
\expandafter\def\csname res@R119@kev_transfer\endcsname{54.3}
\expandafter\def\csname res@R119@massive_en\endcsname{91.6}
\expandafter\def\csname res@R119@mnli\endcsname{83.3}
\expandafter\def\csname res@R119@oasst1\endcsname{78.9}
\expandafter\def\csname res@R119@ood\endcsname{78.0}
\expandafter\def\csname res@R119@oodece\endcsname{0.060}
\expandafter\def\csname res@R119@peakgb\endcsname{6.3}
\expandafter\def\csname res@R119@procedural_typed_decisions\endcsname{81.5}
\expandafter\def\csname res@R119@qasc\endcsname{81.5}
\expandafter\def\csname res@R119@snli\endcsname{87.5}
\expandafter\def\csname res@R119@squad2\endcsname{84.8}
\expandafter\def\csname res@R119@sst5\endcsname{56.4}
\expandafter\def\csname res@R119@sstep\endcsname{3.10}
\expandafter\def\csname res@R119@thead\endcsname{1.07}
\expandafter\def\csname res@R119@tydiqa\endcsname{91.1}
\expandafter\def\csname res@R119@typed_decisions\endcsname{75.6}
\expandafter\def\csname res@R119@valnll\endcsname{0.447}
\expandafter\def\csname res@R120@ag_news\endcsname{81.9}
\expandafter\def\csname res@R120@banking77\endcsname{93.0}
\expandafter\def\csname res@R120@belebele_en\endcsname{67.4}
\expandafter\def\csname res@R120@boolq\endcsname{84.7}
\expandafter\def\csname res@R120@clinc150\endcsname{92.5}
\expandafter\def\csname res@R120@cosmosqa\endcsname{72.5}
\expandafter\def\csname res@R120@dbpedia14\endcsname{94.7}
\expandafter\def\csname res@R120@emotion\endcsname{92.7}
\expandafter\def\csname res@R120@helpsteer2\endcsname{60.1}
\expandafter\def\csname res@R120@id\endcsname{82.0}
\expandafter\def\csname res@R120@idece\endcsname{0.033}
\expandafter\def\csname res@R120@kev_transfer\endcsname{56.4}
\expandafter\def\csname res@R120@massive_en\endcsname{91.5}
\expandafter\def\csname res@R120@mnli\endcsname{83.7}
\expandafter\def\csname res@R120@oasst1\endcsname{79.0}
\expandafter\def\csname res@R120@ood\endcsname{78.1}
\expandafter\def\csname res@R120@oodece\endcsname{0.059}
\expandafter\def\csname res@R120@peakgb\endcsname{6.3}
\expandafter\def\csname res@R120@procedural_typed_decisions\endcsname{81.5}
\expandafter\def\csname res@R120@qasc\endcsname{85.5}
\expandafter\def\csname res@R120@snli\endcsname{87.7}
\expandafter\def\csname res@R120@squad2\endcsname{84.8}
\expandafter\def\csname res@R120@sst5\endcsname{57.4}
\expandafter\def\csname res@R120@sstep\endcsname{3.15}
\expandafter\def\csname res@R120@thead\endcsname{1.08}
\expandafter\def\csname res@R120@tydiqa\endcsname{91.1}
\expandafter\def\csname res@R120@typed_decisions\endcsname{76.1}
\expandafter\def\csname res@R120@valnll\endcsname{0.448}

\providecommand{\res}[2]{\ifcsname res@#1@#2\endcsname\csname res@#1@#2\endcsname\else\tbd\fi}
\expandafter\def\csname res@DA27@dbacc\endcsname{90.0}
\expandafter\def\csname res@DA27@dbacclo\endcsname{86.9}
\expandafter\def\csname res@DA27@dbacchi\endcsname{92.9}
\expandafter\def\csname res@DA27@dbpair\endcsname{81.8}
\expandafter\def\csname res@DA27@dbhard\endcsname{87.6}
\expandafter\def\csname res@DA27@dbcost\endcsname{328}
\expandafter\def\csname res@DA27@dbpfifty\endcsname{1{,}262}
\expandafter\def\csname res@DA27@dbrank\endcsname{6}
\expandafter\def\csname res@DA27@dbn\endcsname{31}
\expandafter\def\csname res@DA27@dbselfrank\endcsname{2}
\expandafter\def\csname res@DA27@dbselfn\endcsname{22}
\expandafter\def\csname res@pplx@dbacc\endcsname{92.2}
\expandafter\def\csname res@pplx@dbpair\endcsname{86.2}
\expandafter\def\csname res@clef@dbacc\endcsname{85.3}
\expandafter\def\csname res@clef@dbpair\endcsname{72.4}
\expandafter\def\csname res@jev@dbacc\endcsname{88.0}
\expandafter\def\csname res@jev@dbpair\endcsname{76.4}
\expandafter\def\csname res@DA27@bhorig\endcsname{100.0}
\expandafter\def\csname res@DA27@bheasy\endcsname{100.0}
\expandafter\def\csname res@DA27@bhhard\endcsname{75.7}
\expandafter\def\csname res@DA27@bhhardece\endcsname{0.080}
\expandafter\def\csname res@DA27@bhrank\endcsname{5}
\expandafter\def\csname res@DA27@bhn\endcsname{53}
\expandafter\def\csname res@jev@bhhard\endcsname{73.0}
\expandafter\def\csname res@reflex@bhhard\endcsname{75.7}
\expandafter\def\csname res@think@bhhard\endcsname{76.6}
\expandafter\def\csname res@frontier@bhhard\endcsname{96.4}
\expandafter\def\csname res@DA27@diindex\endcsname{65.39}
\expandafter\def\csname res@DA27@diclean\endcsname{59.9}
\expandafter\def\csname res@DA27@ditrained\endcsname{68.1}
\expandafter\def\csname res@DA27@dicleanrank\endcsname{4}
\expandafter\def\csname res@DA27@diwins\endcsname{7}
\expandafter\def\csname res@DA27@diwinstrained\endcsname{6}
\expandafter\def\csname res@torchcast@diindex\endcsname{65.10}
\expandafter\def\csname res@torchcast@diclean\endcsname{62.3}
\expandafter\def\csname res@torchcast@ditrained\endcsname{61.9}
\expandafter\def\csname res@pplx@diindex\endcsname{62.25}
\expandafter\def\csname res@pplx@diclean\endcsname{61.3}
\expandafter\def\csname res@pplx@ditrained\endcsname{56.5}
\expandafter\def\csname res@clef@diindex\endcsname{61.71}
\expandafter\def\csname res@clef@diclean\endcsname{60.2}
\expandafter\def\csname res@clef@ditrained\endcsname{57.3}
\expandafter\def\csname res@jev@diindex\endcsname{57.96}
\expandafter\def\csname res@jev@diclean\endcsname{55.8}
\expandafter\def\csname res@jev@ditrained\endcsname{53.3}
\expandafter\def\csname res@DA27@cohoverall\endcsname{81.5}
\expandafter\def\csname res@DA27@cohlo\endcsname{80.2}
\expandafter\def\csname res@DA27@cohhi\endcsname{82.6}
\expandafter\def\csname res@DA27@cohrep\endcsname{88.0}
\expandafter\def\csname res@DA27@cohlog\endcsname{96.6}
\expandafter\def\csname res@DA27@cohbat\endcsname{70.3}
\expandafter\def\csname res@DA27@cohmarg\endcsname{66.6}
\expandafter\def\csname res@DA27@cohorder\endcsname{25.8}
\expandafter\def\csname res@DA27@cohsolo\endcsname{77.1}
\expandafter\def\csname res@DA27@cohdup\endcsname{97.1}
\expandafter\def\csname res@DA27@cohorderexcess\endcsname{0.06}

\newcommand{\hlS}[1]{{\setlength{\fboxsep}{0pt}\colorbox{blue!12}{#1}}}
\newcommand{\hlQ}[1]{{\setlength{\fboxsep}{0pt}\colorbox{orange!25}{#1}}}
\newcommand{\hlO}[1]{{\setlength{\fboxsep}{0pt}\colorbox{green!20}{#1}}}

\title{Small Heads, Strong Decisions: A Controlled Study of Typed-Decision Models}

\author{Ritvik Rastogi \\ \mbox{} \\ \texttt{rastogiritvik99@gmail.com} \and Elvis Saravia \\ DAIR.AI \\ \texttt{elvissaravia@dair.ai}}
\date{}

\begin{document}
\maketitle

\begin{abstract}
In September 2026, TypeSafe announced Jev as the first model in its \emph{System One} category: unstructured state goes in, and typed, calibrated decisions that software can use directly come out.
Jev's architecture and training data are unpublished, but many open typed-decision models have followed, including Nimble, the Decision 1.0--3.0 family and Clef.
They answer structured questions about a piece of text, such as a yes/no check, a choice among labelled options or a graded score, by reading the text with a language model and scoring the options with a small readout.
Each pairs a different readout with a different backbone, training mix and recipe, so it is unclear which design choices matter.
We re-implement five readouts and train all of them under one protocol: the same backbone, the same 15 public English training sources, the same recipe with a per-head learning-rate sweep, and the same seed.
Our experiments demonstrate three things.
First, the complexity of the head does not affect accuracy: on Qwen3-0.6B every readout reaches 81.3--82.0\% in-domain accuracy, and a 0.5M-parameter \emph{cosine head}, which scores each option by its cosine similarity to the question, matches the 27M-parameter joint schema head adapted from Clef in-domain at half its inference time.
Second, reading all of a text's questions from one prompt matches one prompt per question in accuracy, with 1.6 times faster inference and 2.5 times faster training on one H100, though under causal attention its answers remain sensitive to question order.
Third, the backbone's size matters more than either choice: moving from Qwen3-0.6B to Qwen3-4B adds 4 points in-domain and 11 points on held-out sources, while the base, instruct and thinking checkpoints of the same model tie. Checkpoints trained further for embedding or reranking bring no gain.
A 27B model built on these findings with a larger training mix reaches \res{DA27}{dbacc}\% accuracy on DecideBench, \res{DA27}{bhhard}\% on the JevBench hard tier, \res{DA27}{diindex} on the Decision Index and \res{DA27}{cohoverall} on coherence JevBench.
We recommend spending effort on the backbone and keeping the head minimal.
\end{abstract}

\section{Introduction}
\label{sec:intro}

Many systems built around large language models need fast, structured judgements about text rather than free-form generation: does the answer follow the policy, is the question answerable from the passage, how helpful is the response on a 1--5 scale.
\emph{Typed-decision models} serve this need.
They read a piece of text (the \emph{state}) together with one or more typed questions, each with a closed set of options, and return a probability distribution over each question's options without generating any text.
TypeSafe announced Jev in September 2026 as the first model in its \emph{System One} category~\citep{jev}.
Questions come in three types~\citep{jevdocs}: a \emph{noul} asks a yes/no question, a \emph{choice} picks one option from a list, and a \emph{score} rates the state on a rubric.

Jev's architecture, weights and training data are unpublished; it is available only through an API.
Many open Jev-like models have followed; one public leaderboard lists about 150~\citep{jevindex}.
We focus on those whose architectures are documented~\citep{nimble2026,vllmsr2026decision,clef2026}.
They differ along two axes.
The \emph{prompt regime}: Nimble and Decision 1.0--3.0 build one prompt per question, while Clef encodes the state and all its questions in one prompt.
The \emph{readout}: an \emph{LM letter readout} that uses the LM-head rows of the answer letters (Nimble), a \emph{learned letter readout} that learns a copy of those rows (Decision 3.0), a \emph{bilinear-MLP head} over option endpoints (Decision 1.0/2.0), or a 27M-parameter \emph{joint schema head} (Clef).
Each release also uses its own backbone, training data and recipe, so published numbers do not show which design choice matters.

In a controlled comparison, we re-implement the four published readouts, verify our implementations against the original code, and add a minimal one, the \emph{cosine head}: the cosine similarity between the pooled question and each pooled option, scaled by a learned temperature.
All five train on the same backbone, the same 15 public English sources and the same recipe, with each head's learning rate chosen by the same sweep.
We evaluate on the 15 training sources and on five held-out sources, with paired, row-clustered statistics and a one-point non-inferiority margin.
We then fix the head and vary the backbone: scale (Qwen3 0.6B to 14B, three Gemma~4 and three Ministral~3 models), base vs.\ post-trained checkpoints, instruct vs.\ thinking post-training, and checkpoints trained further for embedding and reranking.
Finally, we apply the recommendations at 27B, with a full fine-tune and a larger training mix, and evaluate the resulting model on four public typed-decision benchmarks. \\
Our contributions are:
\begin{enumerate}
  \item A taxonomy of open typed-decision architectures by prompt regime and readout (\cref{sec:related}, \cref{tab:taxonomy});
  \item A controlled comparison of their readouts and prompt regimes under one backbone, data mix and recipe (\cref{sec:results});
  \item The cosine head, a minimal readout that matches the largest head at 2\% of its parameters (\cref{sec:cosine-head});
  \item A backbone study across scale, base vs.\ post-trained checkpoints, instruct vs.\ thinking variants, and retrieval-tuned checkpoints (\cref{sec:ablations});
  \item decide-anything-27b, a model that applies the recommendations at 27B, evaluated on DecideBench, JevBench, the Decision Index and coherence JevBench (\cref{sec:scale-up}).
\end{enumerate}
Our main findings are:
\begin{itemize}
  \item \textbf{The complexity of the head does not affect accuracy.} On Qwen3-0.6B, every readout reaches 81.3--82.0\% in-domain accuracy, and the 0.5M-parameter cosine head matches the 27M-parameter joint schema head in-domain at half its inference time (\cref{sec:res-heads,sec:res-cost}).
  \item \textbf{One prompt per row is enough.} Reading all of a row's questions from one prompt matches one prompt per question in accuracy, with 1.6 times faster inference and 2.5 times faster training on one H100 (\cref{sec:res-regime,sec:res-cost}).
  \item \textbf{The backbone dominates.} Moving from Qwen3-0.6B to Qwen3-4B adds 4 points in-domain and 11 points on held-out sources, while the base, instruct and thinking checkpoints of the same model tie. Checkpoints trained further for embedding or reranking bring no gain in-domain and trail by up to 2.1 points held-out (\cref{sec:ablations}).
  \item \textbf{The recipe holds up at 27B.} A 27B model with a cosine head reaches \res{DA27}{dbacc}\% accuracy on DecideBench, \res{DA27}{bhhard}\% on the JevBench hard tier, \res{DA27}{diindex} on the Decision Index and \res{DA27}{cohoverall} on coherence JevBench (\cref{sec:scale-up}).
\end{itemize}

The code will be open sourced at \url{https://github.com/Ritvik19/DecideAnything}.

\section{Related Work}
\label{sec:related}

We cover Jev, TypeSafe's closed \emph{System One} model, the open typed-decision models whose architectures are documented, and the work on decoder representations and similarity-based scoring that their readouts build on.
We characterise each model by two design choices: its \emph{prompt regime}, meaning whether the backbone encodes one question per forward pass or all of a state's questions jointly, and its \emph{readout}, the function that maps the backbone's final hidden states to one logit per option.
\Cref{tab:taxonomy} places the models along these two axes.
The figures in this section draw each readout on its original prompt, and each caption states what our implementation changes.
For the controlled head comparison, we implement the three readouts whose released architectures score only one question per prompt (Nimble, Decision 1.0/2.0 and Decision 3.0) to the shared joint-prompt template of \cref{sec:joint-prompt}, and report their costs in their native per-question regime; Clef and our cosine head use one joint prompt per row.

\begin{table}[!htbp]
\centering
\small
\begin{tabular}{@{}lllrl@{}}
\toprule
Model & Prompt regime & Readout & New params & Attention \\
\midrule
Jev~\citep{jev} & \multicolumn{4}{c}{closed; architecture not published} \\
\midrule
Nimble~\citep{nimble2026} & one per question & LM letter readout & 0 & causal \\
Decision 1.0/2.0~\citep{vllmsr2026decision,vllmsr2026decision2} & one per question & bilinear-MLP head & 1.05M & causal \\
Decision 3.0~\citep{vllmsr2026d3} & one per question & learned letter readout & 0.26M & mixed$^\dagger$ \\
Clef~\citep{clef2026} & all questions jointly & joint schema head & 27.3M & causal \\
\midrule
Ours & all questions jointly & cosine head & 0.53M & causal \\
\bottomrule
\end{tabular}

\smallskip
\raggedright\footnotesize $^\dagger$Bidirectional in its full-attention layers; not ported: our learned letter readout keeps causal attention.
\caption{Typed-decision architectures. ``New params'' counts the readout's new parameters in our port at hidden size $H{=}1024$ (Qwen3-0.6B), excluding LoRA adapters.}
\label{tab:taxonomy}
\end{table}

\subsection{Jev}
TypeSafe announced Jev on 15 September 2026 as the first \emph{System One} model.
An input is a pair $(s, \mathcal{Q})$ of a state $s$, given as text or JSON, and a set of typed questions $\mathcal{Q} = \{q_1, \dots, q_Q\}$.
Each question has one of three types~\citep{jevdocs}.
A \emph{noul} is a yes/no question, answered with $p(\text{yes}) \in [0, 1]$.
A \emph{choice} has a closed option set and is answered with a distribution over it.
A \emph{score} has an ordered rubric and is answered with a distribution over its levels.
Both multi-option types also return a scalar confidence summarising that distribution.
The documentation says each question is evaluated ``in parallel and in isolation against the same state''~\citep{jevdocs}, so the output for $q_j$ depends only on $(s, q_j)$; it does not say whether the questions share a forward pass.

\subsection{One question per prompt}
Most open designs build a separate prompt for every question: the state, one question and its options.
The state is therefore re-encoded once per question.
A causal backbone can share the state's key--value cache across a state's questions at inference, but training still pays for every copy.

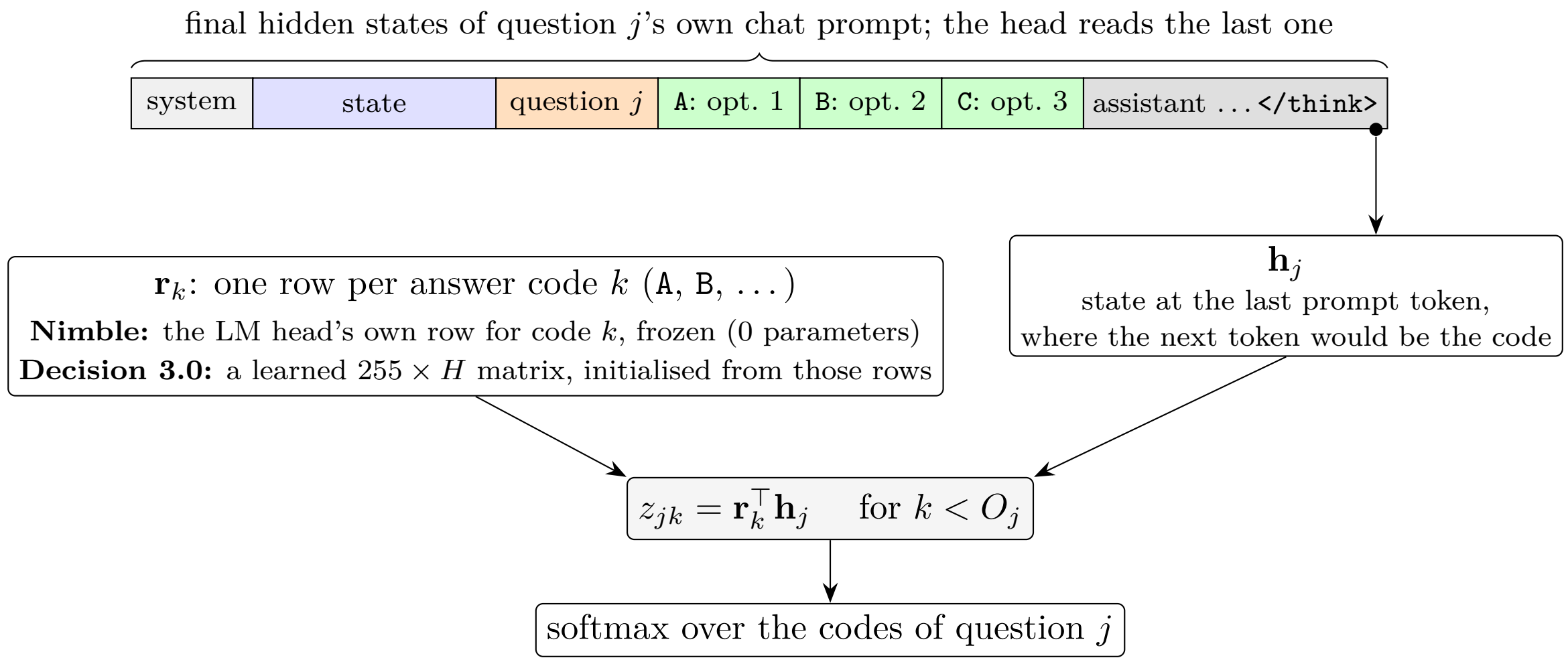
\begin{figure}[!htbp]
\centering
\begin{tikzpicture}[x=1mm, y=1mm, font=\small,
    tok/.style={draw, minimum height=5mm, inner sep=1pt, font=\scriptsize},
    box/.style={draw, rounded corners=2pt, inner sep=3pt, align=center},
    ar/.style={-{Stealth[length=2mm]}},
    dot/.style={circle, fill=black, inner sep=0pt, minimum size=1.3mm}]
  \node[tok, fill=gray!12,   minimum width=12mm] at (6,0)    (pre) {system};
  \node[tok, fill=blue!12,   minimum width=24mm] at (24,0)   (st)  {state};
  \node[tok, fill=orange!25, minimum width=16mm] at (44,0)   (q)   {question $j$};
  \node[tok, fill=green!20,  minimum width=14mm] at (59,0)   (o1)  {\texttt{A}: opt.\ 1};
  \node[tok, fill=green!20,  minimum width=14mm] at (73,0)   (o2)  {\texttt{B}: opt.\ 2};
  \node[tok, fill=green!20,  minimum width=14mm] at (87,0)   (o3)  {\texttt{C}: opt.\ 3};
  \node[tok, fill=gray!25,   minimum width=30mm] at (109,0)  (ans) {assistant \dots\texttt{</think>}};
  \draw[decorate, decoration={brace, amplitude=4pt}] (0,3.5) -- (124,3.5)
    node[midway, above=4pt, font=\footnotesize] {final hidden states of question $j$'s own chat prompt; the head reads the last one};
  \node[dot] at ([xshift=-1.2mm]ans.south east) (ansd) {};

  \node[box] at (114,-19) (h) {$\vect{h}_j$\\[-1pt]\scriptsize state at the last prompt token,\\[-2pt]\scriptsize where the next token would be the code};
  \draw[ar] (ansd) -- (h.north -| ansd);

  \node[box] at (34,-22) (rows) {$\vect{r}_k$: one row per answer code $k$ (\texttt{A}, \texttt{B}, \dots)\\[1pt]
    \scriptsize\textbf{Nimble:} the LM head's own row for code $k$, frozen (0 parameters)\\[-1pt]
    \scriptsize\textbf{Decision 3.0:} a learned $255 \times H$ matrix, initialised from those rows};

  \node[box, fill=gray!8] at (69,-40) (z) {$z_{jk} = \vect{r}_k^{\top} \vect{h}_j$ \quad for $k < O_j$};
  \draw[ar] (rows.south) -- (z.north west);
  \draw[ar] (h.south) -- (z.north east);
  \node[box] at (69,-52) (p) {softmax over the codes of question $j$};
  \draw[ar] (z) -- (p);
\end{tikzpicture}
\caption{The LM letter readout of Nimble and the learned letter readout of Decision 3.0. Each question gets its own chat prompt, with its options labelled by single-token answer codes, ending right after the assistant's empty thinking block. Option $k$ scores the dot product between the hidden state at the last token and code $k$'s row. With Nimble's frozen LM-head rows, this is exactly the language model's next-token logit for the code, restricted to the question's codes; Decision 3.0 learns the rows instead, starting from the same values. Nimble's prompt also lists the other fields in its schema and names the requested one. In our implementation, a row's questions share the joint prompt, and each field ends with \texttt{ANSWER:}, read at its colon.}
\label{fig:code-readout}
\end{figure}

\paragraph{Nimble.}
Nimble~\citep{nimble2026} adds no layers.
Each question becomes one field of a JSON schema whose options are labelled with single-token answer codes (A, B, C, \dots); a chat prompt holds the state and the full schema and ends by naming the requested field.
The readout is an \emph{LM letter readout}: the language model's own next-token logits at the code tokens, so the classifier is the pretrained LM-head rows of those codes, and only the LoRA adapters are trained (\cref{fig:code-readout}).
The loss is cross-entropy over the code logits alone, with hard labels and option order shuffled per row.
The trainer allows up to 26 options per field, and serving allows up to 255.
The released model has 9B parameters and ships with a fitted temperature.

\begin{figure}[!htbp]
\centering
\begin{tikzpicture}[x=1mm, y=1mm, font=\small,
    tok/.style={draw, minimum height=5mm, inner sep=1pt, font=\scriptsize},
    box/.style={draw, rounded corners=2pt, inner sep=3pt, align=center},
    ar/.style={-{Stealth[length=2mm]}},
    dot/.style={circle, fill=black, inner sep=0pt, minimum size=1.3mm}]
  \node[tok, fill=blue!12,   minimum width=28mm] at (14,0)   (st)  {\texttt{Context:} state};
  \node[tok, fill=gray!12,   minimum width=12mm] at (34,0)   (ty)  {type};
  \node[tok, fill=orange!25, minimum width=18mm] at (49,0)   (q)   {question $j$};
  \node[tok, fill=green!20,  minimum width=16mm] at (66,0)   (o1)  {option 1};
  \node[tok, fill=green!20,  minimum width=16mm] at (82,0)   (o2)  {option 2};
  \node[tok, fill=green!20,  minimum width=16mm] at (98,0)   (o3)  {option 3};
  \node[tok, fill=gray!25,   minimum width=20mm] at (116,0)  (end) {\dots\texttt{Decision:}};
  \draw[decorate, decoration={brace, amplitude=4pt}] (0,3.5) -- (126,3.5)
    node[midway, above=4pt, font=\footnotesize] {final hidden states of question $j$'s own prompt; the head reads $K{+}1$ of them};
  \foreach \o in {o1, o2, o3} \node[dot] at ([xshift=-1.2mm]\o.south east) (\o-d) {};
  \node[dot] at ([xshift=-1.2mm]end.south east) (endd) {};

  \node[box] at (52,-16) (c) {$\vect{c}_i = \operatorname{LN}_c(\vect{h}_{e_i})$, one per option\\[-1pt]\scriptsize state at the closing \texttt{>} of the option block};
  \node[box] at (110,-16) (qq) {$\vect{q}_j = \operatorname{LN}_q(\vect{h}_{r_j})$\\[-1pt]\scriptsize state at the final \texttt{:} of \texttt{Decision:}};
  \foreach \o in {o1-d, o2-d, o3-d} \draw[ar] (\o) -- (c.north);
  \draw[ar] (endd) -- (qq.north);

  \node[box] at (38,-33) (bil) {$s_i = (\vect{W}_k \vect{c}_i)^{\top} (\vect{W}_q \vect{q}_j) / \sqrt{d}$\\[-1pt]\scriptsize scaled dot product, as in attention};
  \node[box] at (100,-33) (mlp) {$r_i = \vect{w}^{\top} \operatorname{GELU}(\vect{A}_c \vect{c}_i + \vect{b} + \vect{A}_q \vect{q}_j)$\\[-1pt]\scriptsize learned correction; $\vect{w}$ starts near 0};
  \draw[ar] (c.south) -- (bil.north);
  \draw[ar] (c.south) -- (mlp.north);
  \draw[ar] (qq.south) -- (bil.north);
  \draw[ar] (qq.south) -- (mlp.north);

  \node[box, fill=gray!8] at (69,-48) (z) {$z_i = s_i + r_i$};
  \draw[ar] (bil.south) -- (z.north west);
  \draw[ar] (mlp.south) -- (z.north east);
  \node[box] at (69,-59) (p) {softmax over the options of question $j$};
  \draw[ar] (z) -- (p);
\end{tikzpicture}
\caption{The bilinear-MLP head of Decision 1.0/2.0. Each question gets its own plain-text prompt: the state, the question type, the question, one block per option, and an instruction ending in \texttt{Decision:}. The head reads one hidden state per option (the endpoint $e_i$, the closing \texttt{>} of its block) and one per question (the query $r_j$, the final colon), layer-normalises each, and adds a scaled dot product to a small MLP correction; $d{=}256$, giving 1.05M parameters at $H{=}1024$. No term compares two options, so options interact only through the causal backbone. In our implementation, a row's questions share the joint prompt: the endpoint is the last token of each option's JSON and the query is the field's \texttt{END FIELD} line.}
\label{fig:candidate-head}
\end{figure}
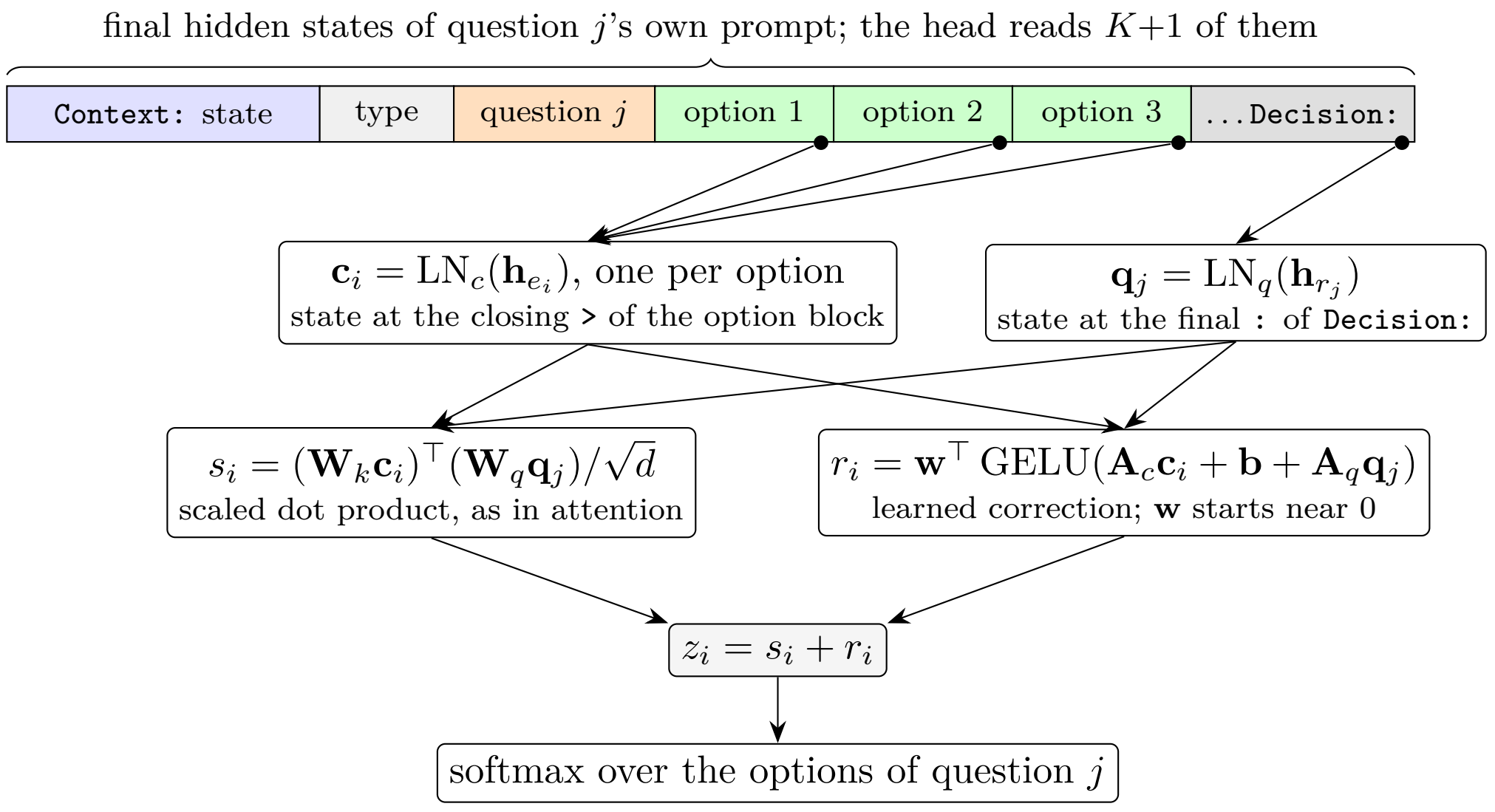

\paragraph{Decision 1.0 and 2.0.}
The Decision models from the vLLM Semantic Router project~\citep{vllmsr2026decision} write each question as a plain-text prompt (context, task type, question, then the options in delimited blocks) ending in \texttt{Decision:}.
The head reads $K{+}1$ hidden states: the last token of each of the $K$ option blocks, and the final \texttt{:} as a query.
A \emph{bilinear-MLP head} scores option $i$ as a scaled dot product between projections of its endpoint and of the query, plus a small MLP correction on their sum (\cref{fig:candidate-head}):
$z_i = (\vect{W}_k \vect{c}_i)^\top (\vect{W}_q \vect{q})/\sqrt{d} + \vect{w}^\top \mathrm{GELU}(\vect{A}_c \vect{c}_i + \vect{A}_q \vect{q} + \vect{b})$.
No term compares two options directly; options interact only through the causal backbone, which also makes the scores depend on option order.
Decision 1.0's causal models and all of Decision 2.0 share this function, on Qwen3.5 backbones from 0.8B to 9B, plus a Qwen3-0.6B-Base model (2.0 Kai) and a 27B model (2.0 Vega).
Decision 1.0 also includes encoder models, which we do not study.
The Decision 1.0 report lists its public training sources~\citep[\S4]{vllmsr2026decision}, which our mix follows (\cref{sec:setup}).

\paragraph{Decision 3.0.}
Decision 3.0~\citep{vllmsr2026d3} keeps the Qwen3.5 backbones but changes everything above them.
The prompt lists the options under answer codes, as in Nimble, and a \emph{learned letter readout} multiplies the last prompt token's hidden state by a learned $[255, H]$ matrix initialised from the code tokens' LM-head rows (\cref{fig:code-readout}).
The full-attention layers are made bidirectional, while the linear-attention (Gated DeltaNet) layers stay causal.
Bidirectional attention lets the state's tokens see the question that follows them, but it rules out sharing a key--value cache across questions.
No training data is published.

\subsection{Multiple questions per prompt}

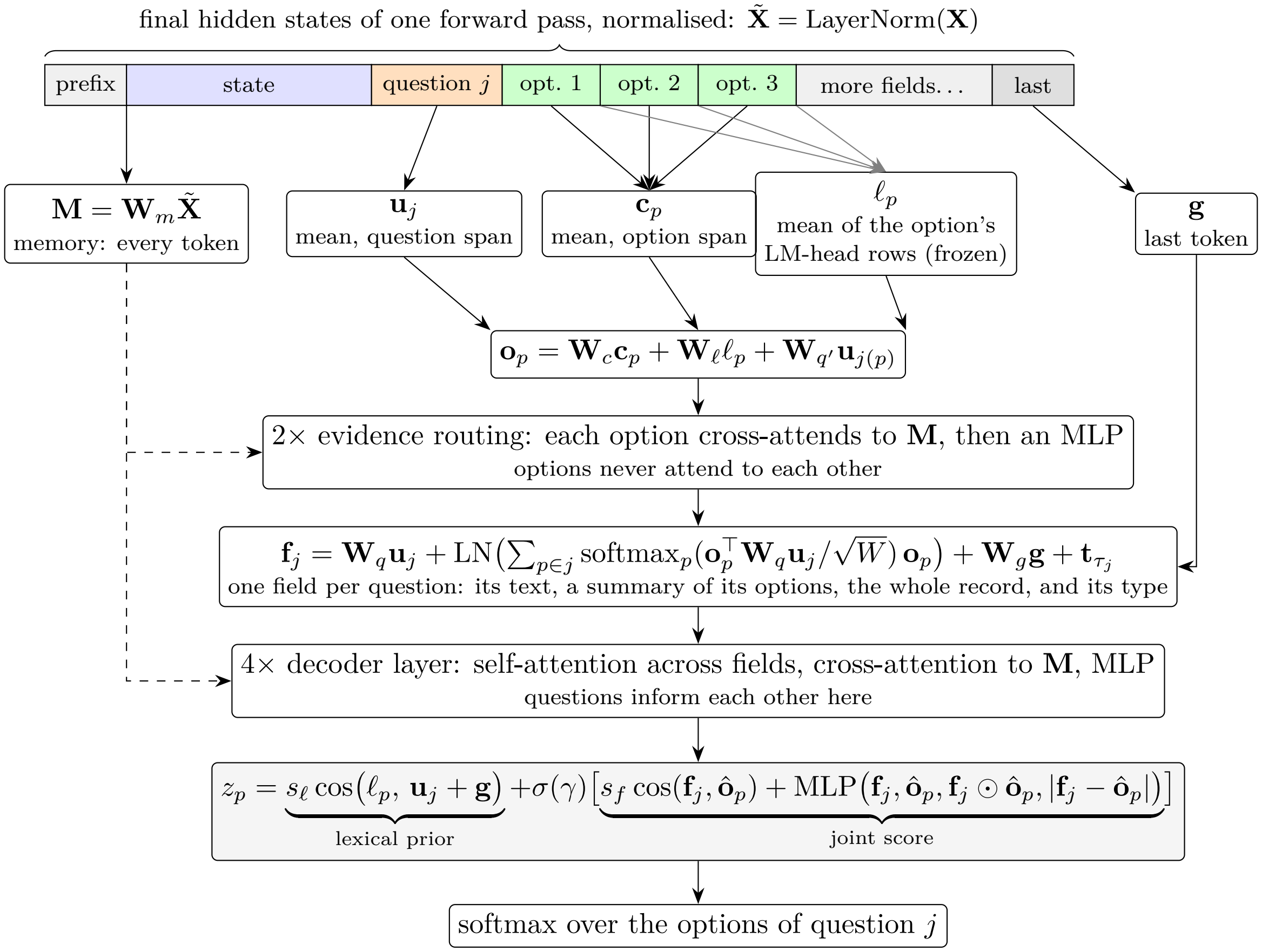
\begin{figure}[!htbp]
\centering
\begin{tikzpicture}[x=1mm, y=1mm, font=\small,
    tok/.style={draw, minimum height=5mm, inner sep=1pt, font=\scriptsize},
    box/.style={draw, rounded corners=2pt, inner sep=3pt, align=center},
    ar/.style={-{Stealth[length=2mm]}},
    mem/.style={-{Stealth[length=2mm]}, dashed}]
  \node[tok, fill=gray!12,   minimum width=10mm] at (5,0)    (pre)  {prefix};
  \node[tok, fill=blue!12,   minimum width=30mm] at (25,0)   (st)   {state};
  \node[tok, fill=orange!25, minimum width=16mm] at (48,0)   (q)    {question $j$};
  \node[tok, fill=green!20,  minimum width=12mm] at (62,0)   (o1)   {opt.\ 1};
  \node[tok, fill=green!20,  minimum width=12mm] at (74,0)   (o2)   {opt.\ 2};
  \node[tok, fill=green!20,  minimum width=12mm] at (86,0)   (o3)   {opt.\ 3};
  \node[tok, fill=gray!12,   minimum width=24mm] at (104,0)  (more) {more fields\dots};
  \node[tok, fill=gray!25,   minimum width=10mm] at (121,0)  (last) {last};
  \draw[decorate, decoration={brace, amplitude=4pt}] (0,3.5) -- (126,3.5)
    node[midway, above=4pt, font=\footnotesize] {final hidden states of one forward pass, normalised: $\tilde{\vect{X}} = \operatorname{LayerNorm}(\vect{X})$};

  \node[box] at (10,-17)  (M) {$\vect{M} = \vect{W}_m \tilde{\vect{X}}$\\[-1pt]\scriptsize memory: every token};
  \node[box] at (44,-17)  (u) {$\vect{u}_j$\\[-1pt]\scriptsize mean, question span};
  \node[box] at (74,-17)  (c) {$\vect{c}_p$\\[-1pt]\scriptsize mean, option span};
  \node[box] at (103,-17) (l) {$\vect{\ell}_p$\\[-1pt]\scriptsize mean of the option's\\[-2pt]\scriptsize LM-head rows (frozen)};
  \node[box] at (141,-17) (g) {$\vect{g}$\\[-1pt]\scriptsize last token};
  \draw[ar] (10,-2.5) -- (M.north);
  \draw[ar] (q.south) -- (u.north);
  \foreach \o in {o1, o2, o3} \draw[ar] (\o.south) -- (c.north);
  \foreach \o in {o1, o2, o3} \draw[ar, gray] (\o.south east) -- (l.north);
  \draw[ar] (last.south) -- (g.north west);

  \node[box] at (80,-33) (o0) {$\vect{o}_p = \vect{W}_c \vect{c}_p + \vect{W}_\ell \vect{\ell}_p + \vect{W}_{q'} \vect{u}_{j(p)}$};
  \draw[ar] (u.south) -- (o0.north west);
  \draw[ar] (c.south) -- (o0.north);
  \draw[ar] (l.south) -- (o0.north east);
  \node[box] at (80,-45) (route) {$2\times$ evidence routing: each option cross-attends to $\vect{M}$, then an MLP\\[-1pt]\scriptsize options never attend to each other};
  \draw[ar] (o0) -- (route);

  \node[box] at (80,-59) (f0) {$\vect{f}_j = \vect{W}_q \vect{u}_j + \operatorname{LN}\bigl(\textstyle\sum_{p \in j} \operatorname{softmax}_p(\vect{o}_p^{\top} \vect{W}_q \vect{u}_j / \sqrt{W})\, \vect{o}_p\bigr) + \vect{W}_g \vect{g} + \vect{t}_{\tau_j}$\\[-1pt]\scriptsize one field per question: its text, a summary of its options, the whole record, and its type};
  \draw[ar] (route) -- (f0);
  \draw[ar] (g.south) |- (f0.east);
  \node[box] at (80,-73) (dec) {$4\times$ decoder layer: self-attention across fields, cross-attention to $\vect{M}$, MLP\\[-1pt]\scriptsize questions inform each other here};
  \draw[ar] (f0) -- (dec);
  \draw[mem] (M.south) |- (route.west);
  \draw[mem] (M.south) |- (dec.west);

  \node[box, fill=gray!8] at (80,-89) (z) {$z_p = \underbrace{s_\ell \cos\bigl(\vect{\ell}_p,\, \vect{u}_j + \vect{g}\bigr)}_{\text{lexical prior}} + \sigma(\gamma)\bigl[\underbrace{s_f \cos(\vect{f}_j, \hat{\vect{o}}_p) + \operatorname{MLP}\bigl(\vect{f}_j, \hat{\vect{o}}_p, \vect{f}_j \odot \hat{\vect{o}}_p, |\vect{f}_j - \hat{\vect{o}}_p|\bigr)}_{\text{joint score}}\bigr]$};
  \draw[ar] (dec) -- (z);
  \node[box] at (80,-103) (p) {softmax over the options of question $j$};
  \draw[ar] (z) -- (p);
\end{tikzpicture}
\caption{Clef's joint schema head. From the normalised hidden states it builds a token memory $\vect{M}$ and pooled question, option, lexical and last-token vectors. Each option gathers evidence from the memory; each question becomes a field vector, and four decoder layers refine the fields jointly. An option's logit adds a lexical prior to a gated joint score that compares its question's field with the option ($\hat{\vect{o}}_p = \operatorname{LN}(\vect{o}_p)$). The width is $W{=}512$, with 8 attention heads and $s_\ell, s_f, \gamma$ learned scalars, giving 27.3M parameters at $H{=}1024$.}
\label{fig:clef-head}
\end{figure}

\paragraph{Clef.}
Cloudflare's Clef~\citep{clef2026} puts the state, every question and every option into one prompt, so a single forward pass answers all of a state's questions.
Its \emph{joint schema head} pools each question span and each option span from the final hidden states.
Each option then cross-attends to a projected copy of the token states to gather evidence, and each question is summarised into a field vector that four Transformer decoder layers refine, with self-attention across the state's questions and cross-attention back to the tokens.
An option's logit combines a lexical prior (the cosine between the option's averaged LM-head rows and the question vector plus the final-token state) with a gated joint score: a cosine plus an MLP over the decoded field and the routed option.
The head has 27.3M parameters at $H{=}1024$ and 32M at $H{=}2560$ (\cref{fig:clef-head}).
The head is open source (Apache-2.0) and implemented, with a training and calibration recipe, in Unsloth, an open-source framework for fine-tuning language models~\citep{unsloth}. Clef's training data is private.
Our joint schema head follows that implementation, with the same prompt format and module names, so its weights load into the reference head.

\subsection{Decoder language models as encoders}
In a causal model only the last token has seen the whole input, which favours last-token pooling, while averaging over a span spreads the read across tokens that each saw only a prefix.
SGPT~\citep{muennighoff2022sgpt} uses position-weighted mean pooling for decoder embeddings; LLM2Vec~\citep{behnamghader2024llm2vec} enables bidirectional attention and adapts the model so that mean pooling works well; NV-Embed~\citep{lee2025nvembed} removes the causal mask during contrastive training and pools with a learned latent-attention layer.
Qwen3 Embedding~\citep{zhang2025qwen3embedding} keeps causal attention, takes the last token's state as the embedding and trains with cosine similarity. Our cosine head makes the same three choices, and \cref{sec:abl-retrieval} tests the Qwen3 Embedding checkpoints as backbones.
The existing decision models already split along this line: Decision 1.0/2.0 read the last token of each option block, Nimble and Decision 3.0 read the last prompt token, and Clef mean-pools spans.
Our pooling ablation (\cref{sec:abl-pooling}) crosses the two choices for question and option spans with causal and bidirectional attention under the same adaptation recipe.

\subsection{Scoring options by similarity}
Scoring a label by its similarity to the input, rather than with a fixed output layer, is well established.
Zero-shot text classification can be posed as entailment between the input and a natural-language statement of each label~\citep{yin2019benchmarking}, and label embeddings can be learned jointly with word embeddings~\citep{wang2018leam}.
Multiple-choice heads on pretrained encoders score each (context, option) pair with a shared linear layer~\citep{devlin2019bert}.
CLIP~\citep{radford2021clip} scores image--text pairs by cosine similarity scaled by a learned temperature capped at 100, the same form as our head's final step.
The cosine head applies this idea to typed decisions inside one shared forward pass: the question and its options are encoded together with the state, then compared by cosine.

\section{Method}
\label{sec:method}

\subsection{Problem formulation}
A training row consists of a state $s$ (the text being judged) and questions $q_1, \dots, q_Q$.
Question $q_j$ carries instructions and a set of $O_j$ options, and the gold label is a distribution $\vect{y}_j \in \Delta^{O_j - 1}$.
A decision model outputs logits $\vect{z}_j \in \mathbb{R}^{O_j}$ for each question, and its prediction is $p(\cdot \mid q_j) = \operatorname{softmax}(\vect{z}_j / T_j)$, where $T_j$ is a temperature fitted after training (\cref{sec:calibration}).
The predicted option is $\arg\max_k z_{jk}$.

\subsection{The joint prompt}
\label{sec:joint-prompt}
\begin{figure}[!htbp]
\centering
\begin{minipage}{\linewidth}
\begin{Verbatim}[commandchars=\\\{\},fontsize=\scriptsize,frame=single,framesep=2mm]
<|im_start|>system
Read the complete state and schema. Decide every field jointly. Each answer must be exactly one of
that field's allowed options.<|im_end|>
<|im_start|>user
STATE:
\hlS{The backpack looks great, but one strap ripped after a week of normal use. For $80 I expected}
\hlS{much better stitching. I'd like a refund.}

SCHEMA FIELDS:

FIELD 1
ID: topic
TYPE: choice
INSTRUCTION: \hlQ{What is the review mainly about?}
ALLOWED OPTIONS:
OPTION 1: \hlO{\{"description":"how support staff treated the customer","option_id":"customer_service"\}}
OPTION 2: \hlO{\{"description":"shipping speed, lost or damaged parcels","option_id":"delivery"\}}
OPTION 3: \hlO{\{"description":"cost and value for money","option_id":"price"\}}
OPTION 4: \hlO{\{"description":"defects, durability, materials, how well it works","option_id":"product_quality"\}}
END FIELD

FIELD 2
ID: sentiment
TYPE: score
INSTRUCTION: \hlQ{How positive is the review overall?}
ALLOWED OPTIONS:
OPTION 1: \hlO{\{"description":"very negative","option_id":"0"\}}
\textit{[options 2--4 elided]}
OPTION 5: \hlO{\{"description":"very positive","option_id":"4"\}}
END FIELD

FIELD 3
ID: wants_refund
TYPE: noul
INSTRUCTION: \hlQ{Does the reviewer ask for a refund?}
ALLOWED OPTIONS:
OPTION 1: \hlO{\{"description":"The proposition is true or the answer is yes.","option_id":"true"\}}
OPTION 2: \hlO{\{"description":"The proposition is false or the answer is no.","option_id":"false"\}}
END FIELD

<|im_end|>
<|im_start|>assistant
<think>

</think>

JOINT SCHEMA DECISIONS:
\end{Verbatim}
\end{minipage}
\caption{The joint prompt for one row with three questions, one of each type. The cosine head reads the \colorbox{blue!12}{state}, each question span (the \colorbox{orange!25}{instruction text}) and each option span (the \colorbox{green!20}{option's JSON}). Line breaks inside the system prompt and the state are added for layout, and the middle options of the score question are elided.}
\label{fig:prompt}
\end{figure}

We use Clef's prompt format~\citep{clef2026}, as implemented in Unsloth~\citep{unsloth}, so the cosine head and the joint schema head see identical tokens and differ only in the head.
\Cref{fig:prompt} shows one row.
The prompt is a chat in the backbone's template.
A fixed system instruction comes first, then a user turn that holds the state followed by a schema with one field per question.
A field gives the question's id, type and instruction text, then lists its options, each rendered as compact JSON with a description and an option id.
The assistant turn holds an empty thinking block and the prefilled words \texttt{JOINT SCHEMA DECISIONS:}.

Each question type maps to a fixed list of options.
A noul has two options, \emph{true} then \emph{false}, with default descriptions unless the source supplies its own.
A choice lists its options sorted by id.
A score lists its rubric levels in order, with ids $0, \dots, O_j{-}1$.
The same order indexes the logits and the targets.

As the state comes first, every question and option token attends to the whole state.
Each option token also sees its own question and every earlier field, so questions can inform each other through the backbone even though the head scores them independently.
When a row exceeds the maximum length, we cut only the end of the state and always keep the schema whole.

One forward pass of the backbone gives final hidden states $\vect{X} \in \mathbb{R}^{L \times H}$ for the whole row, and the head reads every question's answer from them.

The one-prompt-per-question models use the same format with a single field per prompt.

\subsection{The cosine head}
\label{sec:cosine-head}
\begin{figure}[!htbp]
\centering
\begin{tikzpicture}[x=1mm, y=1mm, font=\small,
    tok/.style={draw, minimum height=5mm, inner sep=1pt, font=\scriptsize},
    box/.style={draw, rounded corners=2pt, inner sep=3pt, align=center},
    ar/.style={-{Stealth[length=2mm]}}]
  \node[tok, fill=gray!12,   minimum width=10mm] at (5,0)   (pre)  {prefix};
  \node[tok, fill=blue!12,   minimum width=30mm] at (25,0)  (st)   {state};
  \node[tok, fill=orange!25, minimum width=16mm] at (48,0)  (q)    {question $j$};
  \node[tok, fill=green!20,  minimum width=12mm] at (62,0)  (o1)   {opt.\ 1};
  \node[tok, fill=green!20,  minimum width=12mm] at (74,0)  (o2)   {opt.\ 2};
  \node[tok, fill=green!20,  minimum width=12mm] at (86,0)  (o3)   {opt.\ 3};
  \node[tok, fill=gray!12,   minimum width=20mm] at (102,0) (more) {more fields\dots};
  \draw[decorate, decoration={brace, amplitude=4pt}] (0,3.5) -- (112,3.5)
    node[midway, above=4pt, font=\footnotesize]
    {final hidden states of one forward pass, normalised: $\tilde{\vect{X}} = \operatorname{RMSNorm}_h(\vect{X})$};

  \node[box] at (28,-16) (u) {$\vect{u}_j$\\[-1pt]\scriptsize the question span's last token (or its mean)};
  \node[box] at (90,-16) (v) {$\vect{v}_p$, one per option\\[-1pt]\scriptsize the option span's last token (or its mean)};
  \draw[ar] (q.south) -- (u.north);
  \foreach \o in {o1, o2, o3} \draw[ar] (\o.south) -- (v.north);

  \node[box] at (28,-31) (qh) {$\hat{\vect{q}}_j = \operatorname{RMSNorm}_q(\vect{W}_q \vect{u}_j)$};
  \node[box] at (90,-31) (oh) {$\hat{\vect{o}}_p = \operatorname{RMSNorm}_o(\vect{W}_o \vect{v}_p)$};
  \draw[ar] (u) -- (qh);
  \draw[ar] (v) -- (oh);

  \node[box, fill=gray!8] at (59,-45) (z) {$z_p = s \cdot \tfrac{1}{W}\,\hat{\vect{q}}_{j(p)}^{\top}\hat{\vect{o}}_p$, \quad $s = \min(e^{\theta}, 100)$\\[-1pt]\scriptsize a scaled cosine while the RMS gains are 1};
  \draw[ar] (qh.south) -- (z.north -| qh.south);
  \draw[ar] (oh.south) -- (z.north -| oh.south);
  \node[box] at (59,-58) (p) {softmax over the options of question $j$};
  \draw[ar] (z) -- (p);
\end{tikzpicture}
\caption{The cosine head. One forward pass encodes the state and every question. For each question, the head pools the question span and each of its option spans, projects both into a shared $W$-dimensional space, and scores each option by its scaled cosine with its own question. Options are never compared with another question, and the head has no per-type parameters.}
\label{fig:cosine-head}
\end{figure}
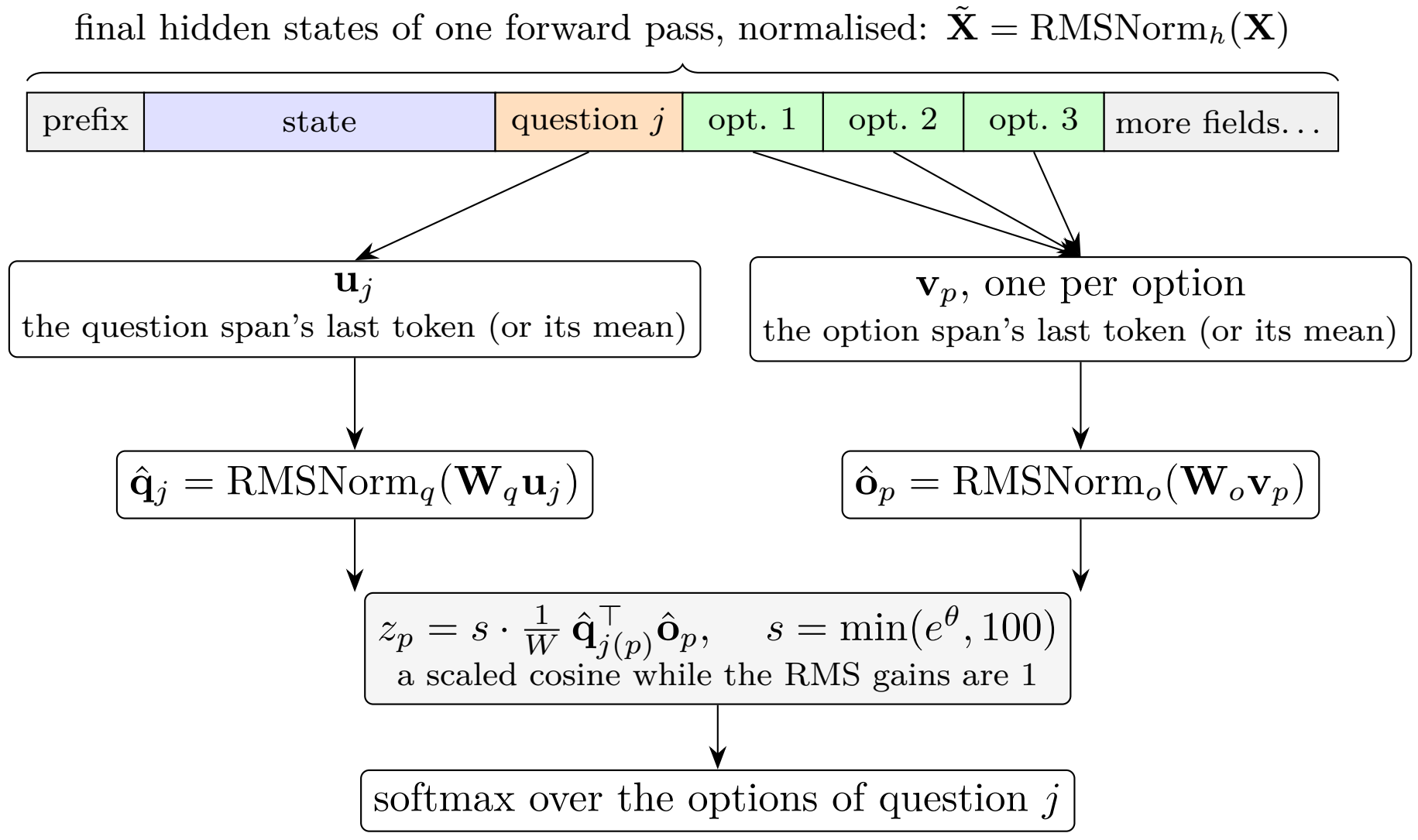

The head turns the hidden states into one logit per option with two projections, three RMS norms~\citep{zhang2019rmsnorm} and one scalar (\cref{fig:cosine-head}):
\begin{align}
  \tilde{\vect{X}} &= \operatorname{RMSNorm}_h(\vect{X}), \\
  \vect{u}_j &= \tilde{\vect{x}}_{b_j}, \qquad
  \vect{v}_p = \tilde{\vect{x}}_{e_p}, \\
  \hat{\vect{q}}_j &= \operatorname{RMSNorm}_q(\vect{W}_q \vect{u}_j), \qquad
  \hat{\vect{o}}_p = \operatorname{RMSNorm}_o(\vect{W}_o \vect{v}_p), \\
  z_p &= \min(e^{\theta}, 100) \cdot \tfrac{1}{W}\, \hat{\vect{q}}_{j(p)}^{\top} \hat{\vect{o}}_p .
\end{align}

Here $j(p)$ is the question that owns option $p$, $\vect{W}_q, \vect{W}_o \in \mathbb{R}^{W \times H}$ have no bias, $W{=}256$ (varied in \cref{sec:abl-width}), $b_j$ is the last token of question $j$'s span, and $e_p$ is the last token of option $p$'s JSON. In a causal backbone $b_j$ has read the question and the state, and $e_p$ has read the whole option, its question and the state. Causal attention and last-token pooling are the default. The ablation in \cref{sec:abl-pooling} crosses attention direction (causal or bidirectional) with all four combinations of last-token and span-mean pooling for the question and options. Span-mean pooling reads $\operatorname*{mean}_{t \in \mathcal{S}(\cdot)} \tilde{\vect{x}}_t$ over a span's token positions $\mathcal{S}(\cdot)$.

An RMS norm rescales a $W$-dimensional vector to norm $\sqrt{W}$ before applying its learned per-dimension gain.
While the gains are 1, $\frac{1}{W}\hat{\vect{q}}^{\top}\hat{\vect{o}}$ is therefore exactly the cosine of the two projected vectors; once the gains move, it becomes a cosine with learned per-dimension weights.
Normalising after the projections means that neither vector's length, which would otherwise grow with the backbone's activations or with span length, can act as a hidden temperature.
Confidence is set by one learned scale $s = \min(e^{\theta}, 100)$, initialised to 10 and capped at 100, as in CLIP~\citep{radford2021clip}.

Each option is compared only with its own question, so the head never mixes information across questions; any interaction between them happens inside the backbone.
Unused option slots are filled with $-10^4$ when a batch's questions are stacked into one matrix, so they receive no probability.

\paragraph{Size.}
The head has $2HW + H + 2W + 1$ parameters: 525,825 at $H{=}1024$ (Qwen3-0.6B), almost all of them in $\vect{W}_q$ and $\vect{W}_o$, against 27.3M for the joint schema head on the same backbone.
The projections use PyTorch's default initialisation and the RMS gains start at 1, so the untrained head already outputs a scaled cosine of the pooled backbone states.

\subsection{Training objective and optimisation}
\label{sec:training}
The loss and the split into LoRA and head learning rates follow Unsloth's implementation for Clef~\citep{unsloth}.

\paragraph{Loss.}
Every model is trained with the same objective: a soft cross-entropy between each question's gold distribution and its predicted distribution, averaged over the questions in a batch,
\begin{equation}
  \mathcal{L} = -\frac{1}{N} \sum_{j=1}^{N} \sum_{k=1}^{O_j} y_{jk} \log \operatorname{softmax}(\vect{z}_j)_k ,
\end{equation}
where $N$ counts questions, not rows.
Averaging over questions gives every question the same weight, so a row with five questions counts five times as much as a row with one.
Option slots beyond a question's $O_j$ are set to $-10^4$ before the softmax, so they get no probability and no gradient.
We add no label smoothing, Brier term or ordinal term for score questions.

\paragraph{Trained parameters.}
The backbone's weights stay frozen.
We train LoRA adapters~\citep{hu2022lora} (rank 32, $\alpha{=}64$, no dropout) on every attention and MLP projection of every layer, together with all of the head's parameters.
Gradient checkpointing keeps activation memory low, the backbone runs under bf16 autocast, and the head computes in fp32.

\paragraph{Optimiser.}
We use AdamW ($\beta = (0.9, 0.999)$, $\epsilon = 10^{-8}$) with two learning rates: $2 \times 10^{-4}$ for the LoRA adapters, and a separate rate for each head, chosen by a small sweep on validation rows (\cref{sec:lr-sweep}).
The learning rate warms up linearly over the first 5\% of steps and then follows a cosine decay to zero, and gradients are clipped to a global norm of 1.

\paragraph{Sampling.}
Each epoch draws up to 10,000 rows per source. Where a source has more, the second epoch draws rows the first did not use, so the large pools are not memorised. The epoch's rows from all sources are shuffled together.
Rows vary widely in length (a CLINC150 row, with 151 options, runs to about 2,300 tokens), so padding a random batch to its longest row wastes most of the compute.
We therefore sort each step's rows by length, split them into micro-batches of four, and backpropagate each micro-batch's loss weighted by its share of the step's questions.
Since $\mathcal{L}$ is a mean over questions, a step whose $N$ questions fall into micro-batches $c$ with $N_c$ questions and mean losses $\mathcal{L}_c$ has
\begin{equation}
  \mathcal{L} = \sum_c \frac{N_c}{N}\, \mathcal{L}_c ,
\end{equation}
so the accumulated gradient equals the full batch's gradient exactly (up to floating-point summation order), at a fraction of the padding.

\paragraph{Divergence check.}
Every 50 steps from step 600, training stops if the loss is not finite or if the mean loss over the last 200 steps exceeds 1.5 times the best 200-step mean so far. A stopped run is repeated with a lower LoRA learning rate (\cref{sec:stability}).

\subsection{Calibration}
\label{sec:calibration}
A model trained with cross-entropy can rank the options well and still state the wrong confidence; decision models in particular are used by thresholding probabilities, so we calibrate every model after training with temperature scaling~\citep{guo2017calibration}.

Before training, we set aside every fifth row of each source's validation split as calibration data.
For sources without a validation split, we first hold out whole training rows, 10\% of the source's questions and at most 400, and take every fifth of those.
Calibration rows are never trained on and never used in the reported results.

After training, we run the model once over the pooled calibration rows of all sources and fit two levels of temperature on the frozen logits.
A head temperature $T_{\text{head}}$ minimises the cross-entropy against the gold distributions:
\begin{equation}
  T_{\text{head}} = \arg\min_{T} \; -\frac{1}{M} \sum_{j=1}^{M} \vect{y}_j^{\top} \log \operatorname{softmax}(\vect{z}_j / T),
\end{equation}
where $M$ counts calibration questions. We fit against the soft gold distributions as they are, so a source whose annotators disagree does not push the temperature toward over-confidence.
We optimise $\log T$ with L-BFGS and a strong-Wolfe line search, and clamp $T_{\text{head}}$ to $[0.05, 20]$.
Then, for each question type with at least ten calibration questions, a relative temperature $T_{\tau}$ is fitted the same way on the logits already divided by $T_{\text{head}}$, and clamped to $[0.5, 5]$.
A question of type $\tau$ is predicted as $\operatorname{softmax}(\vect{z}_j / (T_{\text{head}} T_{\tau}))$.
This adds four scalars per model and leaves every predicted option unchanged.
The two-level scheme and its clamps follow Unsloth's calibration recipe for Clef~\citep{unsloth}, which fits against one-hot labels instead.

The type-level temperatures correct confidence that is off in the same direction for every source of a type.
They cannot correct a source that is over-confident while another of the same type is under-confident, because both share one temperature.
We report ECE after calibration alongside accuracy, which calibration leaves unchanged, both averaged over sources (\cref{sec:setup}).

\section{Experimental Setup}
\label{sec:setup}

\subsection{Data}
\label{sec:data}
\begin{table}[!htbp]
\centering
\small
\begin{tabular}{@{}lrrrlr@{}}
\toprule
& \multicolumn{3}{c}{Training rows} & \multicolumn{2}{c}{Evaluation} \\
\cmidrule(lr){2-4}\cmidrule(l){5-6}
Source & Pool & Per epoch & Distinct & Split & Questions \\
\midrule
BoolQ & 9,427 & 9,427 & 9,427 & val. & 2,616 \\
SST-5 & 8,544 & 8,544 & 8,544 & test & 2,210 \\
Emotion & 16,000 & 10,000 & 16,000 & test & 2,000 \\
SNLI & 549,367 & 10,000 & 20,000 & test & 9,824 \\
MNLI & 392,702 & 10,000 & 20,000 & test & 9,832 \\
MASSIVE & 11,514 & 10,000 & 11,514 & test & 5,948 \\
Banking77 & 10,003 & 9,600 & 9,603 & test & 3,080 \\
CLINC150 & 15,250 & 10,000 & 15,250 & test & 5,500 \\
TyDi QA & 8,806 & 8,806 & 8,806 & val. & 952 \\
Cosmos QA & 25,262 & 10,000 & 20,000 & val. & 2,388 \\
SQuAD 2.0 & 130,319 & 10,000 & 20,000 & val. & 9,498 \\
HelpSteer2 & 20,324 & 10,000 & 20,000 & val. & 4,150 \\
OASST1 & 23,499 & 10,000 & 20,000 & val. & 12,525 \\
typed-decisions & 1,200 & 1,120 & 1,120 & test & 2,000 \\
procedural & 360,000 & 10,000 & 20,000 & test & 57,016 \\
\midrule
Total &  & 137,497 & 220,264 &  & 129,539 \\
\bottomrule
\end{tabular}
\caption{The 15 training sources. ``Pool'' is the number of training rows available, ``per epoch'' how many each epoch draws (at most 10,000), and ``distinct'' how many different rows the two epochs use. Evaluation is on the test split where one exists, otherwise on the validation rows not used for calibration.}
\label{tab:data}
\end{table}

We train every model on the same mixture of 15 public English sources (\cref{tab:data}), converted into the row format of \cref{sec:method}.
These are BoolQ~\citep{clark2019boolq}, SST-5~\citep{socher2013sst}, Emotion~\citep{saravia2018carer}, SNLI~\citep{bowman2015snli}, MNLI~\citep{williams2018mnli}, MASSIVE (English; scenario and intent)~\citep{fitzgerald2023massive}, Banking77~\citep{casanueva2020banking77}, CLINC150~\citep{larson2019clinc}, TyDi QA (English, as passage-level answerability)~\citep{clark2020tydiqa}, Cosmos QA~\citep{huang2019cosmosqa}, SQuAD 2.0 (answerability)~\citep{rajpurkar2018squad2}, HelpSteer2 (five rubric scores)~\citep{wang2024helpsteer2}, OASST1 (English, 13 questions per message)~\citep{kopf2023oasst}, typed-decisions~\citep{typeddecisions} and procedural-typed-decisions~\citep{proceduraltd}.
Together they cover all three question types, between 2 and 151 options per question, and between one and 13 questions per row.
They follow the public training sources named by the Decision 1.0 report~\citep{vllmsr2026decision}, and add the two typed-decision sets, which have several questions per row.
procedural-typed-decisions is an independent rule-generated set, not the Decision 1.0 procedural curriculum, which is unpublished.

\paragraph{Held-out sources.}
Five further sources are never trained on: AG News and DBpedia-14~\citep{zhang2015charcnn} (topic classification with 4 and 14 options), QASC~\citep{khot2020qasc} (eight-option science questions, with both supporting facts given), Belebele (English)~\citep{bandarkar2024belebele} (four-option reading comprehension) and Kev transfer~\citep{kevtransfer}, the 1,046 transfer questions used by the Decision 1.0 report.
Kev transfer is that report's exact question set. The others are the full public evaluation splits, not the report's samples, so our numbers are not comparable with the report's.
We call accuracy on these five sources \emph{held-out} or out-of-domain (OOD), and accuracy on the evaluation splits of the 15 training sources \emph{in-domain}.

\subsection{Models}
\label{sec:arms}
The main comparison (\cref{tab:main}) trains five readouts on Qwen3-0.6B~\citep{yang2025qwen3}: the cosine head, which pools the question and each option at their last token, the joint schema head (from Clef), the bilinear-MLP head (from Decision 1.0/2.0), the learned letter readout (from Decision 3.0) and the LM letter readout (from Nimble).

For the controlled head comparison, all five readouts use one shared joint-prompt template (\cref{sec:joint-prompt}): one forward pass per row, with each head reading the positions it needs (\cref{fig:clef-head,fig:candidate-head,fig:code-readout}).
The bilinear-MLP and letter readout architectures score only one question per prompt.
The letter readouts also use their required answer-code labels and read positions (Appendix~\ref{app:prompts}).
The implementations let us compare the readouts on a shared row-level prompt, but they do not make the released architectures support joint prompts.
In the cost comparison we charge those three designs in their native one-question-per-prompt regime (\cref{sec:res-cost}).
A sixth model, the cosine head with one prompt per question, compares the two prompt regimes with the readout fixed (\cref{sec:res-regime}).
We check our implementations against the original code (\cref{app:ports}).
The controlled head comparison keeps causal attention for every readout, including the learned letter readout. Separately, we ablate causal versus bidirectional attention in the cosine head (\cref{sec:abl-pooling}); this does not reproduce Decision 3.0's complete training procedure.

\paragraph{Head learning rates.}
\label{sec:lr-sweep}
Each head with parameters gets the same small learning-rate sweep on a subset (2,000 rows per source, one epoch, Qwen3-0.6B), and we keep the rate with the lowest calibrated NLL on the pooled validation (calibration) rows (\cref{tab:lr-sweep}).

The chosen rates are $10^{-3}$ for the cosine head, $3 \times 10^{-4}$ for the joint schema head, $3 \times 10^{-3}$ for the bilinear-MLP head and $10^{-4}$ for the learned letter readout; the LM letter readout has no head parameters.

Among the four causal pooling configurations (mean/mean, mean/last, last/mean and last/last for question/options), the same validation NLL chose the last token for both spans, which is the cosine head's default. \Cref{sec:abl-pooling} tests all four pooling combinations under both attention directions.

\paragraph{Backbones.}
\label{sec:backbones}
We use Qwen3 as the main backbone family because it lets us vary one property of the backbone at a time.
Its dense models from 0.6B to 14B share a tokenizer, chat template and pretraining recipe, and each size is released both as a base checkpoint and post-trained.
At 4B, separate Instruct and Thinking checkpoints (2507) sit beside the hybrid-thinking one, and Qwen3-Embedding and Qwen3-Reranker are trained further from the 0.6B, 4B and 8B checkpoints~\citep{zhang2025qwen3embedding}.
The three post-training comparisons (base vs.\ post-trained, instruct vs.\ thinking, general vs.\ retrieval-tuned) therefore each change only how far, and toward what, the same pretrained model was trained further.
In the controlled study we do not use the Qwen3.5 backbones of the Decision models, which interleave full-attention layers with linear-attention (Gated DeltaNet) layers. Qwen3 uses full attention in every layer, so the attention stack is standard and every layer can be switched to bidirectional attention (\cref{sec:abl-pooling}). The 27B model of \cref{sec:scale-up} is built on such a hybrid backbone.
Three Gemma~4~\citep{gemma4} and three Ministral~3~\citep{ministral3} models test whether the scaling results hold in other families, each with its own pretraining data, tokenizer and architecture.

\subsection{Training}
Every run uses the recipe of \cref{sec:training,sec:calibration}: bf16 without quantisation, LoRA rank 32 with $\alpha{=}64$, 32 rows per step in micro-batches of four, a maximum length of 8,192 tokens, and up to 10,000 rows per source per epoch.
Backbones up to 4B train for two epochs (8,594 steps). Backbones of 8B and above (Qwen3-8B, Qwen3-14B, Gemma~4 12B, Ministral~3 8B and 14B, and the two 8B retrieval-tuned checkpoints) train for one (4,297 steps) to fit the compute budget, and a one-epoch Qwen3-4B run bridges the two protocols (\cref{tab:scale}).

The pilot experiments found that a third epoch raised in-domain accuracy by about half a point but cost 0.7 to 2.0 points on the held-out sources for every head, so we stop at two.
Each run uses one NVIDIA H100 80GB GPU.

\subsection{Evaluation}
We report accuracy (the argmax against the gold argmax), expected calibration error (ECE; 15 equal-width bins over the top-1 confidence) and NLL against the gold distribution, after calibration unless stated.
Each source is weighted equally: in-domain and held-out numbers are unweighted means over sources.

\paragraph{Statistics.}
Two runs are compared on the same questions.
For the difference of the dataset-averaged accuracy, a 95\% interval comes from 2,000 bootstrap resamples that draw whole rows within each source, so that a row's questions stay together.
Per source, we give an exact McNemar test on the questions that one run gets right and the other wrong, uncorrected for multiplicity.
We call two models tied when the interval of their difference lies within $\pm 1$ point, the non-inferiority margin we fixed before the main runs, and we read smaller gaps as ties even when they are statistically detectable.

\section{Results}
\label{sec:results}

\begin{table}[!htbp]
\centering
\small
\setlength{\tabcolsep}{4pt}
\begin{tabular}{@{}lrrlrlrr@{}}
\toprule
& & \multicolumn{2}{c}{In-domain} & \multicolumn{2}{c}{Held-out} & \multicolumn{2}{c}{ECE} \\
\cmidrule(lr){3-4}\cmidrule(lr){5-6}\cmidrule(l){7-8}
Readout & Params & Acc. & $\Delta$ & Acc. & $\Delta$ & ID & OOD \\
\midrule
Cosine & 0.53M & 82.0 & \textit{ref.} & 78.0 & \textit{ref.} & 0.033 & 0.060 \\
Joint schema & 27.3M & 81.9 & $-$0.10 {\scriptsize[$-$0.33, +0.13]} & 76.1 & $-$1.91 {\scriptsize[$-$2.65, $-$1.13]} & 0.034 & 0.077 \\
Bilinear-MLP & 1.05M & 82.0 & +0.00 {\scriptsize[$-$0.23, +0.24]} & 76.5 & $-$1.46 {\scriptsize[$-$2.27, $-$0.66]} & 0.034 & 0.070 \\
Learned letter & 0.26M & 81.3 & $-$0.61 {\scriptsize[$-$0.85, $-$0.35]} & 76.4 & $-$1.60 {\scriptsize[$-$2.49, $-$0.71]} & 0.033 & 0.057 \\
LM letter & 0 & 81.6 & $-$0.31 {\scriptsize[$-$0.55, $-$0.07]} & 79.4 & +1.41 {\scriptsize[+0.54, +2.25]} & 0.033 & 0.063 \\
\midrule
Cosine, per question & 0.53M & 81.9 & $-$0.08 {\scriptsize[$-$0.33, +0.18]} & 77.4 & $-$0.55 {\scriptsize[$-$1.35, +0.32]} & 0.034 & 0.058 \\
\bottomrule
\end{tabular}
\caption{Every readout on Qwen3-0.6B. The cosine head pools both the question and each option by their last token (other poolings: \cref{tab:pooling}). Accuracy (\%) is the unweighted mean over the 15 in-domain sources and over the 5 held-out sources. $\Delta$ is the paired difference from the cosine head on the same questions, with a 95\% bootstrap interval that resamples whole rows (test-set noise only, not seed noise). ECE is after calibration, averaged over sources. All rows use the joint prompt except the last, which gives every question its own prompt with the same cosine head and pooling.}
\label{tab:main}
\end{table}

\subsection{A small head on one prompt is enough}
\label{sec:res-heads}
\label{sec:res-regime}
\Cref{tab:main} compares the five readouts on the joint prompt, plus the cosine head with one prompt per question.

\paragraph{Head size.}
The cosine head has 52 times fewer parameters than the joint schema head and matches it in-domain (\res{R01-R68}{id} points). Held-out, the joint schema head differs by \res{R01-R68}{ood} points, mostly on AG News and DBpedia.

\paragraph{The other readouts.}
All five readouts lie between 81.3\% and 82.0\% in-domain.
Held-out, the LM letter readout leads the cosine head by \res{R10-R68}{ood} points, all from Belebele (\res{R10-R68}{belebele_en}), which \cref{sec:res-differ} traces to the language model's pre-training prior over answer codes.

\subsection{Cost}
\label{sec:res-cost}
\begin{table}[!htbp]
\centering
\small
\begin{tabular}{@{}lllrrrr@{}}
\toprule
& & & & In-domain & Training & Inference \\
Readout & Design & Prompt & Params & acc.\ (\%) & (GPU-h) & (ms/row) \\
\midrule
Cosine & ours & joint & 0.53M & 82.0 & 6.9 & 6.4 \\
Joint schema & Clef & joint & 27.3M & 81.9 & 8.7 & 13.6 \\
Bilinear-MLP & Decision 1.0/2.0 & per question & 1.05M & 82.0 & 17.4 & 9.9 \\
Learned letter & Decision 3.0 & per question & 0.26M & 81.3 & 17.6 & 9.4 \\
LM letter & Nimble & per question & 0 & 81.6 & 17.8 & 9.4 \\
\bottomrule
\end{tabular}
\caption{Accuracy and cost of the five readouts on Qwen3-0.6B. Each readout's cost is charged in its design's prompt regime: one joint prompt per row, or one sequence per question. Training is the GPU-hours of one run (8,594 steps of 32 rows) on one idle H100, from the median step time. Inference is the average time per row (one text with all its questions, 2.5 on average) with the GPU kept full by length-sorted batches of up to 32k tokens.}
\label{tab:cost}
\end{table}

\begin{figure}[!htbp]
\centering
\begin{tikzpicture}
\begin{axis}[width=0.7\linewidth, height=5.5cm, xlabel={training time (GPU-hours per run)}, ylabel={in-domain accuracy (\%)},
  ymin=80.8, ymax=82.3, enlarge x limits=0.12, grid=major, tick label style={font=\scriptsize}, label style={font=\small}]
\addplot[only marks, mark=*, black, error bars/.cd, y dir=both, y explicit] coordinates {(6.9083,81.955) += (0,0.000) -= (0,0.000) (8.7134,81.857) += (0,0.229) -= (0,0.230)};
\addplot[only marks, mark=*, mark options={fill=white}, black, error bars/.cd, y dir=both, y explicit] coordinates {(17.4261,81.960) += (0,0.232) -= (0,0.239) (17.5997,81.349) += (0,0.257) -= (0,0.241) (17.7805,81.642) += (0,0.243) -= (0,0.242)};
\node[font=\scriptsize, anchor=west, inner sep=4pt] at (axis cs:6.9083,81.955) {Cosine};
\node[font=\scriptsize, anchor=west, inner sep=4pt] at (axis cs:8.7134,81.857) {Joint schema};
\node[font=\scriptsize, anchor=east, inner sep=4pt] at (axis cs:17.4261,81.960) {Bilinear-MLP};
\node[font=\scriptsize, anchor=east, inner sep=4pt] at (axis cs:17.5997,81.349) {Learned letter};
\node[font=\scriptsize, anchor=east, inner sep=4pt] at (axis cs:17.7805,81.642) {LM letter};
\end{axis}
\end{tikzpicture}
\caption{In-domain accuracy against training time on Qwen3-0.6B (\cref{tab:cost}). Filled: one joint prompt per row; hollow: one sequence per question, as in the readout's original design. Bars are the 95\% paired bootstrap interval of \cref{tab:main} against the cosine head, drawn around each readout's own accuracy.}
\label{fig:cost-train}
\end{figure}
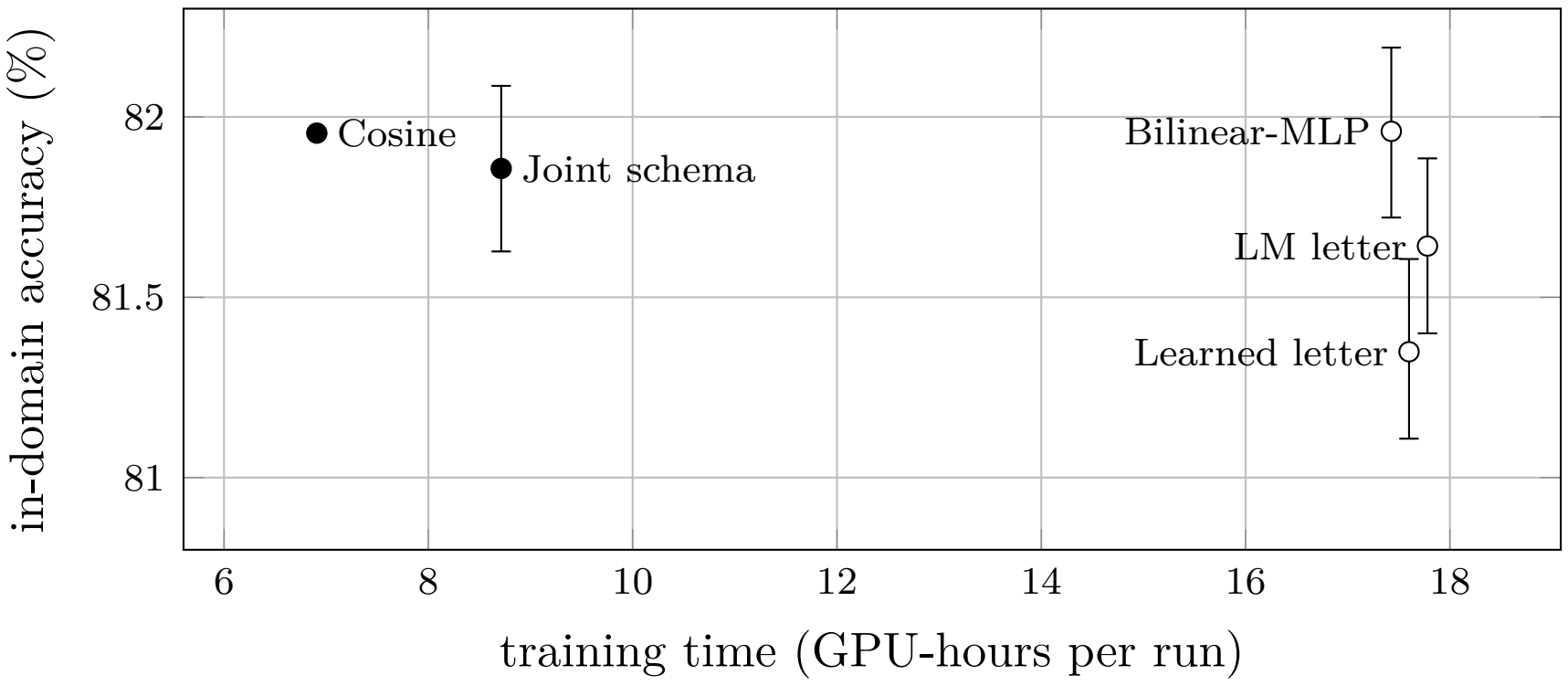
\begin{figure}[!htbp]
\centering
\begin{tikzpicture}
\begin{axis}[width=0.7\linewidth, height=5.5cm, xlabel={inference time (ms per row)}, ylabel={in-domain accuracy (\%)},
  ymin=80.8, ymax=82.3, enlarge x limits=0.12, grid=major, tick label style={font=\scriptsize}, label style={font=\small}]
\addplot[only marks, mark=*, black, error bars/.cd, y dir=both, y explicit] coordinates {(6.3523,81.955) += (0,0.000) -= (0,0.000) (13.5821,81.857) += (0,0.229) -= (0,0.230)};
\addplot[only marks, mark=*, mark options={fill=white}, black, error bars/.cd, y dir=both, y explicit] coordinates {(9.9453,81.960) += (0,0.232) -= (0,0.239) (9.4075,81.349) += (0,0.257) -= (0,0.241) (9.4088,81.642) += (0,0.243) -= (0,0.242)};
\node[font=\scriptsize, anchor=west, inner sep=4pt] at (axis cs:6.3523,81.955) {Cosine};
\node[font=\scriptsize, anchor=east, inner sep=4pt] at (axis cs:13.5821,81.857) {Joint schema};
\node[font=\scriptsize, anchor=south east, inner sep=4pt] at (axis cs:9.9453,81.960) {Bilinear-MLP};
\node[font=\scriptsize, anchor=east, inner sep=4pt] at (axis cs:9.4075,81.349) {Learned letter};
\node[font=\scriptsize, anchor=east, inner sep=4pt] at (axis cs:9.4088,81.642) {LM letter};
\end{axis}
\end{tikzpicture}
\caption{In-domain accuracy against inference time per row on Qwen3-0.6B (\cref{tab:cost}). Filled: one joint prompt per row; hollow: one sequence per question, as in the readout's original design. Bars are the 95\% paired bootstrap interval of \cref{tab:main} against the cosine head, drawn around each readout's own accuracy.}
\label{fig:cost-infer}
\end{figure}
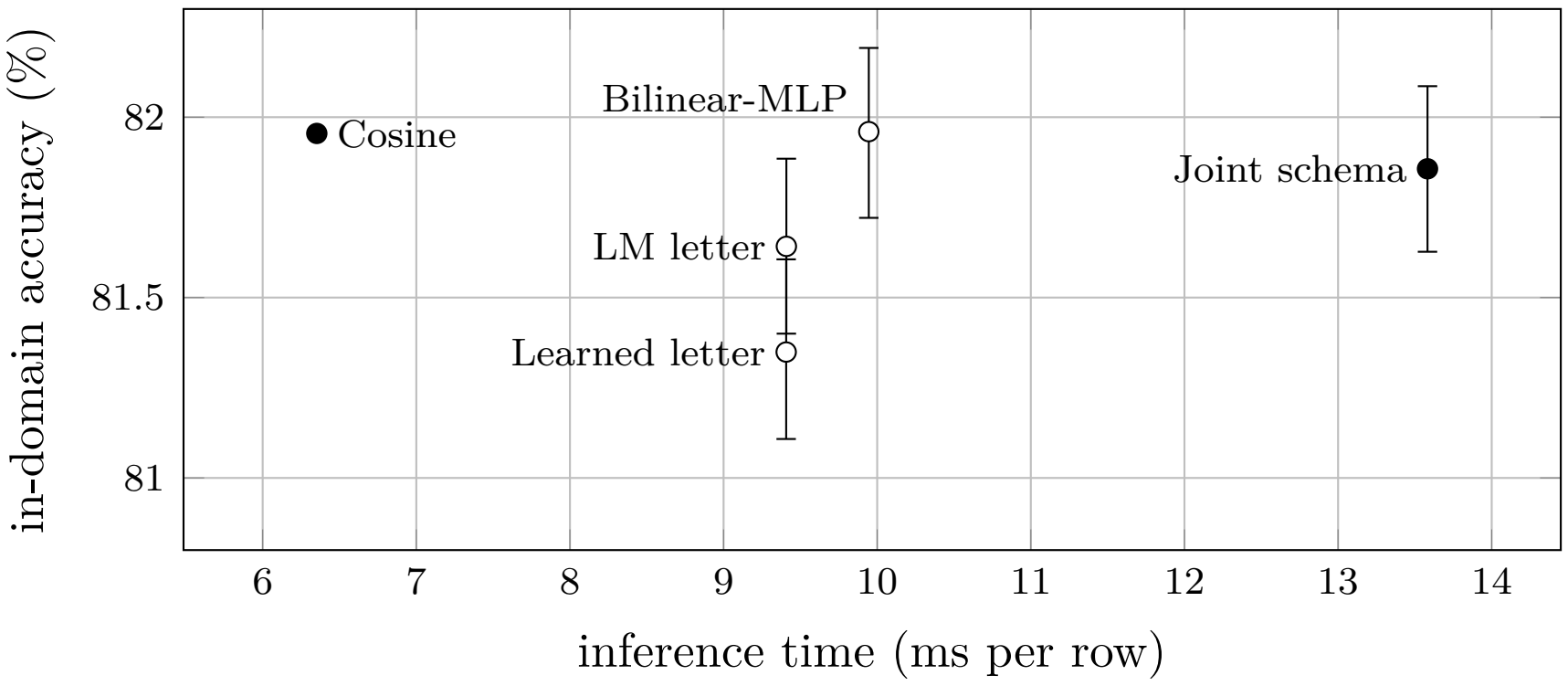
\Cref{tab:cost} charges each readout in its own design's prompt regime: one joint prompt per row for the cosine head and the joint schema head, one sequence per question for the bilinear-MLP head and the two letter readouts.
At the same accuracy, the cosine head is the cheapest to train and to run: \res{cost}{cosine@trainh} GPU-hours per run and \res{cost}{cosine@inferms}\,ms per row, against \res{cost}{candidate@trainh}--\res{cost}{nimble@trainh} GPU-hours and \res{cost}{d3@inferms}--\res{cost}{candidate@inferms}\,ms for the three designs that give every question its own sequence, and \res{cost}{clef@trainh} GPU-hours and \res{cost}{clef@inferms}\,ms for the joint schema head (\cref{fig:cost-train,fig:cost-infer}).

\subsection{Where the readouts do differ: the language-model prior}
\label{sec:res-differ}
\begin{table}[!htbp]
\centering
\small
\begin{tabular}{@{}llrr@{}}
\toprule
Readout & Codes at evaluation & Belebele & Kev MMLU \\
\midrule
Cosine & -- & 66.3 & 38.8 \\
\midrule
LM letter & uppercase letters (\texttt{A}, \texttt{B}, \dots) & 76.1 & 48.8 \\
LM letter & number words & 72.3 & 47.5 \\
LM letter & lowercase letters (\texttt{a}, \texttt{b}, \dots) & 67.4 & 40.0 \\
LM letter & symbols & 59.3 & 32.5 \\
\midrule
LM letter, trained lowercase & lowercase letters & 69.2 & 41.2 \\
\bottomrule
\end{tabular}
\caption{The LM letter readout's held-out advantage depends on its answer codes. Accuracy (\%) of the Qwen3-0.6B checkpoints, all trained with uppercase codes, when the codes are swapped at evaluation only. Belebele has 900 questions; Kev's MMLU subset has 80. The last row is a separate model trained with lowercase codes, scored by the main evaluation (where the uppercase-trained model scores \res{R10}{belebele_en}). Paired against the cosine head on Belebele: uppercase $p{=}2{\times}10^{-10}$, number words $p{=}1.3{\times}10^{-5}$, lowercase $p{=}0.28$, symbols $p{=}3{\times}10^{-4}$ (worse).}
\label{tab:codes}
\end{table}

The LM letter readout's held-out lead comes entirely from Belebele, the one held-out source with options labelled A to D: \res{R10-R68}{belebele_en} points over the cosine head (McNemar $p{=}\res{R10-R68}{belebele_enp}$). On no other held-out source is it significantly ahead.

Swapping its codes at evaluation changes its Belebele accuracy from 76.1\% (uppercase letters) to 72.3\% (number words), 67.4\% (lowercase letters) and 59.3\% (symbols), against 66.3\% for the cosine head (\cref{tab:codes}).
The LM letter readout trained with lowercase codes reaches \res{R67}{belebele_en}\% on Belebele, against \res{R10}{belebele_en}\% with uppercase ones, at the same in-domain accuracy (\res{R67-R10}{id} points).
Relabelling Belebele's options from letters to digits changes no readout by more than 0.7 points.
\section{Ablations}
\label{sec:ablations}

\subsection{Pooling and attention direction}
\label{sec:abl-pooling}
\begin{table}[!htbp]
\centering
\small
\setlength{\tabcolsep}{5pt}
\begin{tabular}{@{}lrrrrrr@{}}
\toprule
& \multicolumn{3}{c}{In-domain} & \multicolumn{3}{c}{Held-out} \\
\cmidrule(lr){2-4}\cmidrule(l){5-7}
Question / option & Causal & Bidir. & $\Delta$ & Causal & Bidir. & $\Delta$ \\
\midrule
last / last & \textbf{\res{R68}{id}} & 77.1 & $-4.9$ & \textbf{\res{R68}{ood}} & 66.2 & $-11.8$ \\
mean / last & \res{R07}{id} & 77.2 & $-4.5$ & \res{R07}{ood} & \textbf{71.3} & $-5.4$ \\
mean / mean & \res{R04}{id} & \textbf{78.3} & $-3.5$ & \res{R04}{ood} & 70.7 & $-6.1$ \\
last / mean & 81.8 & 75.8 & $-6.0$ & 76.2 & 64.4 & $-11.8$ \\
\bottomrule
\end{tabular}
\caption{Pooling and attention direction in the Qwen3-0.6B cosine head. Accuracy (\%), averaged equally over the 15 in-domain or five held-out sources; $\Delta$ is bidirectional minus causal, in percentage points. Bold marks the highest value in each accuracy column. All models use rank-32 LoRA for two epochs. Differences are computed from the displayed means.}
\label{tab:pooling}
\end{table}

We compare question and option pooling (last token or span mean) under causal and bidirectional attention in the Qwen3-0.6B cosine head, using the same rank-32 LoRA and two-epoch training recipe (\cref{tab:pooling}).

Causal accuracy spans 0.3 percentage points in-domain and 1.8 points held-out across the four pooling combinations.
Last/last pooling has the highest accuracy on both averages, and we keep it as the default.
The small in-domain differences do not establish statistical equivalence between poolings.

Bidirectional attention reduces accuracy for every pooling combination, by 3.5--6.0 points in-domain and 5.4--11.8 points held-out.
Mean question pooling raises bidirectional held-out accuracy over last-token question pooling by 5.1 points with last-token option pooling and 6.2 points with mean option pooling, while changing option pooling moves it by only 0.6--1.8 points.
No pooling closes the gap to causal attention under this adaptation recipe, and the best pooling depends on attention direction: last/last under causal attention, mean question pooling under bidirectional attention.

The losses are large on passage comprehension and rule-based tasks (\cref{app:pooling-attention}); on Belebele, bidirectional accuracy falls by 13.9--25.1 points across matched poolings.
This is consistent with the backbone struggling to adapt its contextual representations once the attention direction changes, but the experiments do not isolate the cause or test whether more adaptation capacity or longer training would recover the deficit.

Bidirectional models also have higher held-out ECE (0.074--0.110 versus 0.060--0.071 for causal attention).
Per-source results and further calibration details are in \cref{app:pooling-attention}.

\subsection{Projection width}
\label{sec:abl-width}
\begin{table}[!htbp]
\centering
\small
\begin{tabular}{@{}lrrrrr@{}}
\toprule
& & \multicolumn{2}{c}{In-domain} & \multicolumn{2}{c}{Held-out} \\
\cmidrule(lr){3-4}\cmidrule(l){5-6}
$W$ & Params & Acc. & ECE & Acc. & ECE \\
\midrule
64 & 0.13M & \res{R117}{id} & \res{R117}{idece} & \res{R117}{ood} & \res{R117}{oodece} \\
128 & 0.26M & \res{R118}{id} & \res{R118}{idece} & \res{R118}{ood} & \res{R118}{oodece} \\
256 & 0.53M & \res{R119}{id} & \res{R119}{idece} & \res{R119}{ood} & \res{R119}{oodece} \\
512 & 1.05M & \res{R120}{id} & \res{R120}{idece} & \res{R120}{ood} & \res{R120}{oodece} \\
\bottomrule
\end{tabular}
\caption{Projection width $W$ of the Qwen3-0.6B cosine head (last/last pooling, two epochs). Accuracy (\%) and ECE are averaged over the 15 in-domain or five held-out sources.}
\label{tab:width}
\end{table}

We vary the cosine head's projection width $W$ (\cref{sec:cosine-head}) over 64, 128, 256 and 512 on Qwen3-0.6B, with last/last pooling and the two-epoch recipe (\cref{tab:width}).
This takes the head from 0.13M to 1.05M parameters.

In-domain accuracy spans 0.2 points across the four widths, while held-out accuracy spans 1.3 points, rising from 76.8\% at $W{=}64$ to 78.1\% at $W{=}512$.
We keep $W{=}256$ as the default: it is within 0.1 points of $W{=}512$ held-out at half the parameters.
Broken down by questions per row or by number of options, in-domain accuracy varies by at most 1.3 points across widths in any bucket, with no width best throughout (\cref{app:width}).

\subsection{Scaling the backbone}
\label{sec:abl-scale}
\begin{table}[!htbp]
\centering
\small
\begin{tabular}{@{}llrrrrrr@{}}
\toprule
& & \multicolumn{3}{c}{In-domain} & \multicolumn{3}{c}{Held-out} \\
\cmidrule(lr){3-5}\cmidrule(l){6-8}
Backbone & Ep. & Cosine & Joint schema & $\Delta$ & Cosine & Joint schema & $\Delta$ \\
\midrule
Qwen3-0.6B & 2 & 82.0 & 81.9 & $-$0.10 & 78.0 & 76.1 & $-$1.91 \\
Qwen3-1.7B & 2 & 84.3 & 84.5 & +0.14 & 83.1 & 82.3 & $-$0.80 \\
Qwen3-4B & 2 & 85.9 & 85.9$^{\ddagger}$ & +0.05 & 88.9 & 89.3$^{\ddagger}$ & +0.43 \\
Qwen3-4B & 1 & 85.2 & -- & -- & 87.2 & -- & -- \\
Qwen3-8B & 1 & 85.8$^{\ddagger}$ & 85.9 & +0.07 & 89.8$^{\ddagger}$ & 89.0 & $-$0.80 \\
Qwen3-14B & 1 & 86.6 & 86.4$^{\ddagger}$ & $-$0.17 & 90.8 & 91.4$^{\ddagger}$ & +0.59 \\
\midrule
Gemma 4 E2B & 2 & 85.0 & -- & -- & 85.6 & -- & -- \\
Gemma 4 E4B & 2 & 86.4 & 86.3 & $-$0.06 & 88.1 & 88.0 & $-$0.08 \\
Gemma 4 12B & 1 & 85.0$^{\S}$ & -- & -- & 90.5$^{\S}$ & -- & -- \\
\midrule
Ministral 3 3B & 2 & \res{R141}{id} & \res{R142}{id} & \res{R142-R141}{id} & \res{R141}{ood} & \res{R142}{ood} & \res{R142-R141}{ood} \\
Ministral 3 8B & 1 & \res{R143}{id} & \res{R144b}{id}$^{\S}$ & \res{R144b-R143}{id} & \res{R143}{ood} & \res{R144b}{ood}$^{\S}$ & \res{R144b-R143}{ood} \\
Ministral 3 14B & 1 & \res{R145}{id} & \res{R146}{id} & \res{R146-R145}{id} & \res{R145}{ood} & \res{R146}{ood} & \res{R146-R145}{ood} \\
\bottomrule
\end{tabular}
\caption{Scaling the backbone. Accuracy (\%) of the cosine head and the joint schema head; $\Delta$ is joint schema minus cosine, paired on the same questions. The cosine head pools each option and question at its last token. Backbones of 8B and above, and the bridge row for Qwen3-4B, train for one epoch instead of two. The LoRA learning rate is $2 \times 10^{-4}$ for Qwen3 and Gemma~4, and $10^{-4}$ for Ministral~3, chosen by its own sweep (\cref{tab:lora-sweep}). $^{\ddagger}$LoRA learning rate $10^{-4}$ and $^{\S}$$3 \times 10^{-5}$ instead, after that rate diverged (\cref{sec:stability}).}
\label{tab:scale}
\end{table}

From Qwen3-0.6B to Qwen3-4B, the cosine head rises from \res{R68}{id}\% to \res{R26}{id}\% in-domain and from \res{R68}{ood}\% to \res{R26}{ood}\% held-out (\cref{tab:scale}).
With one epoch, Qwen3-8B and Qwen3-14B reach \res{R30b}{id}\% and \res{R32}{id}\% in-domain, and \res{R30b}{ood}\% and \res{R32}{ood}\% held-out.
On Qwen3-4B, one epoch instead of two costs \res{R66-R26}{id} points in-domain and \res{R66-R26}{ood} held-out.
On Qwen3 and Gemma~4, the joint schema head and the cosine head stay within 0.2 points in-domain at every size. Held-out, they differ by under two points, with no consistent sign.
Gemma~4 \texttt{google/gemma-4-E2B-it} reaches \res{R56}{id}\% in-domain and \res{R56}{ood}\% held-out, \texttt{google/gemma-4-E4B-it} reaches \res{R57}{id}\% and \res{R57}{ood}\%, and \texttt{google/gemma-4-12B-it} (one epoch, LoRA learning rate $3 \times 10^{-5}$) reaches \res{R63c}{id}\% and \res{R63c}{ood}\%~\citep{gemma4,gemma4report}.

Ministral~3 shows the same pattern~\citep{ministral3}.
From 3B to 14B, the cosine head goes from \res{R141}{id}\% to \res{R145}{id}\% in-domain but from \res{R141}{ood}\% to \res{R145}{ood}\% held-out.
The joint schema head stays within 0.6 points of the cosine head at every size, and the two tie everywhere except held-out at 3B, where the interval reaches \res{R142-R141}{oodlo} points and the comparison is inconclusive.
At 8B, the in-domain gap of \res{R144b-R143}{id} points is the only one whose interval excludes zero, but that run used a lower learning rate after diverging at $10^{-4}$ (\cref{sec:stability}), so the gap cannot be attributed to the head alone.

\subsection{Post-training and retrieval tuning}
\label{sec:abl-base}
\label{sec:abl-thinking}
\label{sec:abl-retrieval}
\begin{table}[!htbp]
\centering
\small
\begin{tabular}{@{}lrrrrrr@{}}
\toprule
& \multicolumn{3}{c}{In-domain} & \multicolumn{3}{c}{Held-out} \\
\cmidrule(lr){2-4}\cmidrule(l){5-7}
Backbone & Post-trained & Base & $\Delta$ & Post-trained & Base & $\Delta$ \\
\midrule
Qwen3-0.6B & 82.0 & 81.7 & $-$0.23 & 78.0 & 77.7 & $-$0.29 \\
Qwen3-1.7B & 84.3 & 84.3 & $-$0.03 & 83.1 & 82.7 & $-$0.36 \\
Qwen3-4B & 85.9 & 85.9$^{\ddagger}$ & +0.05 & 88.9 & 89.2$^{\ddagger}$ & +0.35 \\
Qwen3-8B & 85.8$^{\ddagger}$ & 85.9 & +0.09 & 89.8$^{\ddagger}$ & 88.5 & $-$1.33 \\
Qwen3-14B & 86.6 & 86.4 & $-$0.18 & 90.8 & 90.9 & +0.10 \\
\bottomrule
\end{tabular}
\caption{Base versus post-trained Qwen3 backbones, cosine head. ``Post-trained'' is the released hybrid-thinking checkpoint; $\Delta$ is base minus post-trained, paired. 8B and 14B train for one epoch. $^{\ddagger}$LoRA learning rate $10^{-4}$ (\cref{sec:stability}).}
\label{tab:base}
\end{table}

Each comparison in this subsection asks whether training the same pretrained model further helps the decision head, which reads hidden states and never generates text.
The three comparisons go from general post-training, to the style of that post-training, to further training for retrieval, a task close to ours (\cref{sec:backbones}).

\paragraph{Base vs.\ post-trained.}
The base checkpoint has not seen the chat template of our prompt (\cref{sec:joint-prompt}), so post-training could help through familiarity with the prompt format alone.
At every size from 0.6B to 14B, the base and post-trained checkpoints are within 0.25 points in-domain (\cref{tab:base}).
Held-out, they are within one point except at 8B, where the base model is \res{R53-R30b}{ood} points behind, mostly on Belebele (\res{R53-R30b}{belebele_en}) and Kev (\res{R53-R30b}{kev_transfer}). The two 8B runs also use different LoRA learning rates.

\begin{table}[!htbp]
\centering
\small
\begin{tabular}{@{}lrlrl@{}}
\toprule
& \multicolumn{2}{c}{In-domain} & \multicolumn{2}{c}{Held-out} \\
\cmidrule(lr){2-3}\cmidrule(l){4-5}
Backbone & Acc. & $\Delta$ & Acc. & $\Delta$ \\
\midrule
Qwen3-4B (hybrid thinking) & 85.9 & \textit{ref.} & 88.9 & \textit{ref.} \\
Qwen3-4B-Instruct-2507 & 85.9 & +0.06 {\scriptsize[$-$0.16, +0.27]} & 89.4 & +0.53 {\scriptsize[$-$0.03, +1.06]} \\
Qwen3-4B-Thinking-2507 & 85.9 & +0.05 {\scriptsize[$-$0.17, +0.27]} & 89.0 & +0.09 {\scriptsize[$-$0.61, +0.69]} \\
Qwen3-4B-Base$^{\ddagger}$ & 85.9 & +0.05 {\scriptsize[$-$0.17, +0.27]} & 89.2 & +0.35 {\scriptsize[$-$0.23, +0.94]} \\
\bottomrule
\end{tabular}
\caption{Four post-training variants of the same 4B architecture, cosine head, two epochs. $\Delta$ is the paired difference from the hybrid-thinking checkpoint, with a 95\% row-bootstrap interval. $^{\ddagger}$LoRA learning rate $10^{-4}$ (\cref{sec:stability}).}
\label{tab:post-training}
\end{table}

\paragraph{Instruct vs.\ thinking.}
This comparison asks whether post-training for reasoning leaves hidden states that serve decisions better, even though the head generates no reasoning.
4B is the only dense Qwen3 size with separate 2507 Instruct and Thinking checkpoints.
The 2507 checkpoints were also trained further than the hybrid one, so they differ from it in more than the style of post-training.
Qwen3-4B-Instruct-2507, Qwen3-4B-Thinking-2507 and the hybrid Qwen3-4B all reach 85.9\% in-domain (\cref{tab:post-training}). Held-out, Instruct and Thinking differ from the hybrid by \res{R38-R26}{ood} and \res{R41-R26}{ood} points.

\begin{table}[!htbp]
\centering
\small
\begin{tabular}{@{}lrrrrrr@{}}
\toprule
& \multicolumn{3}{c}{In-domain} & \multicolumn{3}{c}{Held-out} \\
\cmidrule(lr){2-4}\cmidrule(l){5-7}
Size & Qwen3 & Embedding & Reranker & Qwen3 & Embedding & Reranker \\
\midrule
0.6B & \res{R68}{id} & \res{R93}{id} & \res{R94}{id} & \res{R68}{ood} & \res{R93}{ood} & \res{R94}{ood} \\
4B & \res{R26}{id} & \res{R97}{id} & \res{R100}{id} & \res{R26}{ood} & \res{R97}{ood} & \res{R100}{ood} \\
8B$^{\ddagger}$ & \res{R30b}{id} & \res{R103}{id} & \res{R106}{id} & \res{R30b}{ood} & \res{R103}{ood} & \res{R106}{ood} \\
\bottomrule
\end{tabular}
\caption{Retrieval-tuned Qwen3 checkpoints as backbones. Accuracy (\%) of the cosine head (last-token question and option pooling), averaged equally over the 15 in-domain or five held-out sources. Qwen3 is the general hybrid-thinking checkpoint of \cref{tab:scale}; Embedding and Reranker are Qwen3-Embedding and Qwen3-Reranker~\citep{zhang2025qwen3embedding} of the same size. 0.6B and 4B train for two epochs, 8B for one. $^{\ddagger}$LoRA learning rate $10^{-4}$, as for Qwen3-8B (\cref{sec:stability}).}
\label{tab:retrieval}
\end{table}

\paragraph{Retrieval-tuned checkpoints.}
Qwen3-Embedding and Qwen3-Reranker~\citep{zhang2025qwen3embedding} are Qwen3 checkpoints trained further for retrieval.
The embedding model compares last-token states by cosine similarity, as the cosine head does (\cref{sec:cosine-head}), and the reranker answers a yes/no relevance question, as a noul does.
Neither improves on the general checkpoint (\cref{tab:retrieval}).
In-domain, both stay within 0.4 points of it at every size. Held-out, neither is ahead at any size.
The largest loss is at 4B, where Embedding and Reranker reach \res{R97}{ood}\% and \res{R100}{ood}\% held-out, below all four general 4B checkpoints (\cref{tab:post-training}), mostly on Belebele and Kev (\cref{app:retrieval}).
At 8B, most of the gap is on Kev, where the general checkpoint may simply score high: the base checkpoint and the joint schema head on the same backbone score lower there.
Neither retrieval checkpoint is consistently ahead of the other.
We use last-token pooling for the question and options, the most accurate pooling we tried on these checkpoints.
\citet{dinzinger2026systemone} find the mirror image: fine-tuning Qwen backbones for typed decisions does not make them better rerankers than reranker fine-tuning does, which wins by 5--29 nDCG@10 points up to 9B and ties at 27B.

\subsection{Training stability}
\label{sec:stability}
The default LoRA learning rate of $2 \times 10^{-4}$ diverged on four larger runs: Qwen3-4B and Qwen3-14B with the joint schema head, and Qwen3-8B and Qwen3-4B-Base with the cosine head. Qwen3-4B-Base finished at \res{R50}{id}\% in-domain and \res{R50}{ood}\% held-out.
Each spike came at or soon after the end of warm-up.
At $10^{-4}$, all four trained normally.
Gemma~4 12B collapsed after warm-up at $10^{-4}$ and trained normally at $3 \times 10^{-5}$.
At the rate chosen by the Ministral~3 sweep, $10^{-4}$, five of the six Ministral~3 runs trained normally. The 8B joint schema run hit a NaN loss at step 1,643 of 4,297, long after warm-up and with no rise in loss before it, and trained normally at $3 \times 10^{-5}$.

The tables mark every run with a lowered rate.

\section{Scaling Up}
\label{sec:scale-up}
The controlled study holds the data and recipe fixed and stops at 14B.
To see how far its recommendations carry outside that setting, we built one model that follows them at 27B with a much larger training mix, and evaluated it on four public typed-decision benchmarks.

\paragraph{Model.}
decide-anything-27b keeps what the study recommends: a cosine head with $W{=}256$ on a post-trained backbone, all of a text's questions read from one prompt, and one temperature per question type fitted on data it never trained on (\cref{sec:calibration}).
The backbone is Qwen3.8-27B~\citep{qwen38}, which interleaves 48 Gated DeltaNet layers with 16 full-attention layers.
Training runs in two stages over 1.54M questions.
The first stage covers 1.18M questions from public datasets and procedural and rule-based generators; the second, 0.36M questions.
The second stage runs with three seeds, and the released model is the uniform average of their weights.

\paragraph{Benchmarks.}
DecideBench v2~\citep{decidebench} has 450 items in 225 contrast pairs from nine decision families, each shown with one example per option; pair accuracy counts a pair only when both of its items are right.
JevBench by Benchmark Heaven~\citep{bhjevbench} scores accuracy, calibration, latency and cost over four tiers, partly on held-out items. We ran the 231 public items of v1.2 and compare on the one tier that every system on the board answered in full, the 111 public hard items.
The Decision Index~\citep{jevindex,jevindexkit} combines 37 benchmarks in five areas into one chance-corrected index; its public 0.3 suite has 140,620 requests. The live board now ranks by a Full score that adds private tests, so we report only this public index.
Coherence JevBench~\citep{jevbenchcoh} does not score accuracy: it edits each request in ways that should leave the answer unchanged, or change it in a known way, and checks that it does, over 240 cases.

\begin{table}[!htbp]
\centering
\small
\begin{tabular}{@{}llrrrr@{}}
\toprule
Benchmark & Metric & Ours & Jev & Clef & pplx-decider \\
\midrule
\multirow{2}{*}{DecideBench v2} & Accuracy & \res{DA27}{dbacc} & \res{jev}{dbacc} & \res{clef}{dbacc} & \res{pplx}{dbacc} \\
& Pair accuracy & \res{DA27}{dbpair} & \res{jev}{dbpair} & \res{clef}{dbpair} & \res{pplx}{dbpair} \\
\midrule
JevBench (BH) & Hard-tier accuracy & \res{DA27}{bhhard} & \res{jev}{bhhard} & -- & -- \\
\midrule
\multirow{3}{*}{Decision Index 0.3} & Index (all 37) & \res{DA27}{diindex} & \res{jev}{diindex} & \res{clef}{diindex} & \res{pplx}{diindex} \\
& 13 with train split$^{\dagger}$ & \res{DA27}{ditrained} & \res{jev}{ditrained} & \res{clef}{ditrained} & \res{pplx}{ditrained} \\
& 24 without & \res{DA27}{diclean} & \res{jev}{diclean} & \res{clef}{diclean} & \res{pplx}{diclean} \\
\bottomrule
\end{tabular}
\caption{decide-anything-27b on three public typed-decision benchmarks. BH: Benchmark Heaven. Baselines are the versions each board lists: Jev 1.14 (DecideBench) and 1.13 (JevBench), Clef 27B, and pplx-decider 27B v1.2 (DecideBench) and v1.1 (Decision Index). $^{\dagger}$Mean over the 13 Decision Index benchmarks whose training splits are in our stage-2 data (\cref{app:scale-up}); the other 24 are scored separately below. Whether the baselines trained on any of these benchmarks is not known.}
\label{tab:scale-up}
\end{table}

\paragraph{Results.}
On DecideBench, the model reaches \res{DA27}{dbacc}\% accuracy (95\% interval \res{DA27}{dbacclo}--\res{DA27}{dbacchi}), sixth of \res{DA27}{dbn} entries and second of \res{DA27}{dbselfn} self-hosted ones (\cref{tab:scale-up}).
The one self-hosted model ahead, pplx-decider 27B at \res{pplx}{dbacc}\%, lies inside that interval; Clef 27B reaches \res{clef}{dbacc}\% and Jev~1.14 \res{jev}{dbacc}\%.
On the JevBench hard tier it scores \res{DA27}{bhhard}\% with an ECE of \res{DA27}{bhhardece}, tied fifth of \res{DA27}{bhn} with reflex-27b. The four systems ahead all generate text before answering: two frontier language models at \res{frontier}{bhhard}\% and two decision models in thinking mode at \res{think}{bhhard}\%. It answers all 72 standard and 48 easy public items correctly. On the Decision Index it scores \res{DA27}{diindex}, first of the 11 entrants on the kit's 0.3 board. Stage 2 includes the training splits of 13 of its 37 benchmarks (\cref{app:scale-up}): on those it scores \res{DA27}{ditrained}, and on the other 24 \res{DA27}{diclean}, \res{DA27}{dicleanrank}th of the 11.
The model scores \res{DA27}{cohoverall} (95\% interval \res{DA27}{cohlo}--\res{DA27}{cohhi}) on coherence JevBench, but its sub-scores are uneven.
It is near ceiling on logical (\res{DA27}{cohlog}) and duplicate (\res{DA27}{cohdup}) consistency, scores \res{DA27}{cohrep} on repeat, \res{DA27}{cohsolo} on solo, \res{DA27}{cohbat} on batch and \res{DA27}{cohmarg} on marginal consistency, and only \res{DA27}{cohorder} on question order.
The order test checks that reordering a request's questions leaves each answer unchanged.
One prompt per question passes it by construction, since each question sees only the state. In the joint prompt under causal attention, each question also sees every question listed before it, and training with shuffled question order did not remove this dependence.

\section{Conclusion}
\label{sec:conclusion}
We compared the readouts of the open typed-decision models under one backbone, one data mix and one recipe, and then varied the backbone with the head held fixed.
A more complex head does not buy accuracy: all five readouts lie between 81.3\% and 82.0\% in-domain, and the 0.5M-parameter cosine head matches the 27M-parameter joint schema head in-domain at half its inference time.
One prompt per row matches one prompt per question in accuracy at a fraction of the cost: inference is 1.6 times faster and training 2.5 times faster. Under causal attention, however, its answers remain sensitive to question order, scoring only \res{DA27}{cohorder} on the order test of coherence JevBench (\cref{sec:scale-up}).
The backbone matters most: from Qwen3-0.6B to Qwen3-4B, accuracy rises by four points in-domain and eleven held-out, while base, instruct and thinking checkpoints of the same model tie. Checkpoints trained further for retrieval add nothing in-domain and trail by up to 2.1 points held-out. This holds even for Qwen3-Embedding, which is trained with the cosine similarity our head computes.
A 27B model built this way, with a larger training mix, ranks among the strongest self-hosted typed-decision models on DecideBench, JevBench and the Decision Index.
For practitioners, this means choosing the backbone carefully, encoding all of a text's questions in one prompt, and keeping the head minimal.

\section{Future Work}
\label{sec:future}
\Cref{sec:scale-up} scales the backbone and the data together, to 27B and 1.54M questions.
Other open questions are how the findings carry over to multimodal inputs, and whether more adaptation capacity or a different training schedule can recover the accuracy lost under bidirectional attention with our current recipe (\cref{sec:abl-pooling}).
The joint prompt matches one prompt per question in accuracy, but its answers depend on question order (\cref{sec:scale-up}), and shuffling the order in training does not remove this.
A block-isolated attention mask, in which each question attends to the state but not to its sibling questions, would keep the single forward pass over the state while removing the interaction between questions, and so separate the two effects of the joint prompt regime (\cref{sec:res-regime}).

\bibliographystyle{plainnat}
\bibliography{refs}

\appendix
\section{Prompt variants}
\label{app:prompts}
All models use the joint prompt of \cref{fig:prompt}, with two variants.
For the two letter readouts, each option line reads \texttt{OPTION A:}, \texttt{OPTION B:}, \dots{} instead of \texttt{OPTION 1:}, using the first 255 answer codes that are single tokens in the backbone's vocabulary, and each field ends with an \texttt{ANSWER:} line whose colon is the read position.
For the one-prompt-per-question models, each prompt holds the state and a single field; a row's prompts stay together for calibration.

\FloatBarrier
\section{Hyperparameters}
\label{app:hparams}
\Cref{tab:hparams} lists the training settings shared by every run.
\Cref{tab:lr-sweep} gives the head learning-rate sweep of \cref{sec:lr-sweep}, and \cref{tab:lora-sweep} the LoRA learning-rate sweep for the Ministral~3 backbones.
\begin{table}[!htbp]
\centering
\small
\begin{tabular}{@{}ll@{}}
\toprule
Setting & Value \\
\midrule
Precision & bf16, no quantisation \\
LoRA & rank 32, $\alpha{=}64$, dropout 0, all attention and MLP projections \\
LoRA learning rate & $2 \times 10^{-4}$; $10^{-4}$ for Ministral~3 (\cref{tab:lora-sweep}); $10^{-4}$ or $3 \times 10^{-5}$ where marked (\cref{sec:stability}) \\
Head learning rate & cosine $10^{-3}$, joint schema $3 \times 10^{-4}$, bilinear-MLP $3 \times 10^{-3}$, learned letter $10^{-4}$ \\
Optimiser & AdamW, $\beta = (0.9, 0.999)$, $\epsilon = 10^{-8}$, weight decay 0.01 (0 on norms and biases) \\
Schedule & linear warm-up over 5\% of steps, then cosine decay to 0 \\
Gradient clipping & global norm 1.0 \\
Batch & 32 rows per step (77 questions on average), in length-sorted micro-batches of 4 \\
Rows & up to 10,000 per source per epoch, fresh rows each epoch where available \\
Epochs & 2 (8,594 steps); 1 (4,297 steps) for backbones of 8B and above \\
Maximum length & 8,192 tokens \\
Calibration & every 5th validation row; head and per-type temperatures, soft targets \\
Seed & 1 \\
Hardware & one NVIDIA H100 80GB per run \\
\bottomrule
\end{tabular}
\caption{Training settings shared by every run, and the per-head learning rates.}
\label{tab:hparams}
\end{table}
\begin{table}[!htbp]
\centering
\small
\begin{tabular}{@{}ll@{}}
\toprule
Head & Learning rate (validation NLL) \\
\midrule
Cosine & 3e-4 (0.669), \textbf{1e-3} (0.660), 3e-3 (0.662) \\
Joint schema & 3e-5 (0.668), 1e-4 (0.671), \textbf{3e-4} (0.659), 1e-3 (0.664) \\
Bilinear-MLP & 3e-4 (0.668), 1e-3 (0.668), \textbf{3e-3} (0.667) \\
Learned letter & 3e-6 (0.719), 1e-5 (0.721), 3e-5 (0.710), \textbf{1e-4} (0.709) \\
\bottomrule
\end{tabular}
\caption{Head learning-rate sweep: calibrated NLL on the pooled validation (calibration) rows, on a cheap protocol (2,000 rows per source, one epoch, Qwen3-0.6B). Bold marks the chosen rate. The LM letter readout has no head parameters and is not swept.}
\label{tab:lr-sweep}
\end{table}

\begin{table}[!htbp]
\centering
\small
\begin{tabular}{@{}lll@{}}
\toprule
Backbone & Head & LoRA learning rate (validation NLL) \\
\midrule
Ministral 3 3B & Cosine & 3e-5 (0.510), \textbf{1e-4} (0.477), 2e-4 (0.499) \\
 & Joint schema & 3e-5 (0.516), \textbf{1e-4} (0.481), 2e-4 (0.502) \\
Ministral 3 8B & Cosine & 3e-5 (0.461), \textbf{1e-4} (0.436), 2e-4 (0.454) \\
 & Joint schema & 3e-5 (0.465), \textbf{1e-4} (0.441), 2e-4 (0.462) \\
Ministral 3 14B & Cosine & 3e-5 (0.445), \textbf{1e-4} (0.418), 2e-4 (0.426) \\
 & Joint schema & 3e-5 (0.455), \textbf{1e-4} (0.424), 2e-4 (0.430) \\
\bottomrule
\end{tabular}
\caption{LoRA learning-rate sweep for Ministral~3: calibrated NLL on the pooled validation (calibration) rows, on the protocol of \cref{tab:lr-sweep} (2,000 rows per source, one epoch), with each head at the learning rate chosen there. Bold marks the chosen rate. The 8B joint schema run diverged at the chosen rate and was retrained at $3 \times 10^{-5}$ (\cref{sec:stability}).}
\label{tab:lora-sweep}
\end{table}

\FloatBarrier
\section{Per-source results}
\label{app:per-dataset}
\Cref{tab:in-domain,tab:ood} give in-domain and held-out accuracy per source for every readout on Qwen3-0.6B.
\begin{table}[!htbp]
\centering
\small
\setlength{\tabcolsep}{4pt}
\begin{tabular}{@{}lrrrrrrr@{}}
\toprule
Source & Questions & Cosine & Joint schema & Bilin.-MLP & Learned let. & LM let. & Cos.\ per-q. \\
\midrule
BoolQ & 2,616 & \textbf{84.8} & 84.5 & 84.1 & 84.4 & 84.1 & 84.7 \\
SST-5 & 2,210 & 56.4 & 55.9 & 56.2 & 55.3 & \textbf{56.6} & 56.1 \\
Emotion & 2,000 & 92.5 & 93.0 & 92.8 & 92.8 & \textbf{93.5} & 92.9 \\
SNLI & 9,824 & 87.4 & \textbf{87.7} & 87.7 & 87.7 & 87.2 & 87.8 \\
MNLI & 9,832 & 84.0 & 84.1 & \textbf{84.2} & 83.7 & 83.4 & 83.9 \\
MASSIVE & 5,948 & 91.1 & \textbf{91.3} & 91.3 & \textbf{91.3} & 91.0 & 90.7 \\
Banking77 & 3,080 & 92.8 & \textbf{93.2} & 93.0 & 93.1 & 93.2 & 92.7 \\
CLINC150 & 5,500 & \textbf{92.7} & 92.3 & \textbf{92.7} & 91.9 & 91.3 & 92.4 \\
TyDi QA & 952 & 91.4 & 91.0 & 91.1 & 91.0 & \textbf{91.7} & 91.2 \\
Cosmos QA & 2,388 & 74.2 & \textbf{74.9} & 74.0 & 72.9 & 74.0 & 73.4 \\
SQuAD 2.0 & 9,498 & \textbf{84.4} & 83.7 & 84.4 & 83.6 & 83.9 & 84.9 \\
HelpSteer2 & 4,150 & \textbf{60.4} & 59.7 & 60.1 & 59.9 & 60.1 & 59.3 \\
OASST1 & 12,525 & 78.8 & 78.8 & \textbf{79.0} & 78.9 & 78.8 & 79.0 \\
typed-decisions & 2,000 & 76.3 & 76.1 & \textbf{76.6} & 76.0 & 76.6 & 76.2 \\
procedural & 57,016 & 81.9 & 81.7 & \textbf{82.3} & 77.9 & 79.3 & 82.9 \\
\midrule
Mean &  & 82.0 & 81.9 & 82.0 & 81.3 & 81.6 & 81.9 \\
\bottomrule
\end{tabular}
\caption{In-domain accuracy (\%) by source on Qwen3-0.6B. ``Questions'' counts each source's final evaluation questions (its test split, or the validation rows not used for calibration). Bold marks the best joint-prompt readout.}
\label{tab:in-domain}
\end{table}

\begin{table}[!htbp]
\centering
\small
\begin{tabular}{@{}lrrrrrr@{}}
\toprule
Readout & AG News & DBpedia & QASC & Belebele & Kev & Mean \\
\textit{questions} & 7,600 & 70,000 & 926 & 900 & 1,046 & \\
\midrule
Cosine & \textbf{82.1} & \textbf{96.9} & \textbf{88.4} & 66.3 & 56.1 & 78.0 \\
Joint schema & 77.9 & 95.1 & 86.0 & 66.0 & 55.4 & 76.1 \\
Bilinear-MLP & 80.6 & 95.4 & 86.6 & 65.3 & 54.6 & 76.5 \\
Learned letter & 77.4 & 94.6 & 85.6 & 68.1 & 56.2 & 76.4 \\
LM letter & 79.8 & 95.8 & 87.5 & \textbf{76.1} & \textbf{57.7} & 79.4 \\
\midrule
Cosine, per question & 78.3 & 93.8 & 86.1 & 68.2 & 60.7 & 77.4 \\
\bottomrule
\end{tabular}
\caption{Held-out accuracy (\%) by source on Qwen3-0.6B. Bold marks the best joint-prompt readout. The second row gives each source's number of questions.}
\label{tab:ood}
\end{table}

\FloatBarrier
\section{Verification of our implementations}
\label{app:ports}
Our bilinear-MLP head keeps the original module names and equation, and loads the state dict of the original Decision 2.0 head strictly, with zero difference in its outputs.
Separately, our from-scratch reproductions use the bilinear-MLP head for Decision 1.0/2.0 and the learned-letter readout for Decision 3.0. They reproduce the outputs of three released checkpoints (Decision-2.0-Kai-0.6B, Decision-1.0-Eos-0.8B and d3-edge) on about 920 questions each: identical token ids, argmax answers and answer fields, with logits equal to within $2.3 \times 10^{-5}$.
The LM letter readout's logits equal the language model's own next-token logits at the code tokens exactly, as in Nimble, and the learned letter readout starts from those same rows.
Our joint schema head keeps the module names of Clef's reference head, so the reference head's weights load into it.

\FloatBarrier
\section{Pooling and attention: detailed results}
\label{app:pooling-attention}
\Cref{tab:pooling-heldout} reports each held-out source for the pooling--attention comparison in \cref{sec:abl-pooling}.
\Cref{tab:pooling-indomain} gives accuracy ranges on selected in-domain sources.
On Belebele, bidirectional attention loses 13.9--25.1 points at matched pooling settings.
This task-level pattern does not separate passage length from task type or reasoning demands.

After calibration, in-domain ECE ranges from 0.026 to 0.029 for the four bidirectional configurations, compared with 0.033--0.034 for the four causal configurations.
Held-out ECE instead worsens under bidirectional attention (\cref{tab:pooling-heldout}), so the lower in-domain ECE does not transfer to the held-out sources.
\begin{table}[!htbp]
\centering
\small
\setlength{\tabcolsep}{3pt}
\begin{tabular}{@{}llrrrrrrr@{}}
\toprule
Pooling & Attention & AG News & DBpedia & QASC & Belebele & Kev & Mean & ECE \\
\midrule
last / last & causal & \res{R68}{ag_news} & \res{R68}{dbpedia14} & \res{R68}{qasc} & \res{R68}{belebele_en} & \res{R68}{kev_transfer} & \res{R68}{ood} & \res{R119}{oodece} \\ 
mean / last & causal & \res{R07}{ag_news} & \res{R07}{dbpedia14} & \res{R07}{qasc} & \res{R07}{belebele_en} & \res{R07}{kev_transfer} & \res{R07}{ood} & \res{R07}{oodece} \\
mean / mean & causal & \res{R04}{ag_news} & \res{R04}{dbpedia14} & \res{R04}{qasc} & \res{R04}{belebele_en} & \res{R04}{kev_transfer} & \res{R04}{ood} & \res{R04}{oodece} \\
last / mean & causal & 80.8 & 94.6 & 84.7 & 66.2 & 54.8 & 76.2 & 0.070 \\
\midrule
last / last & bidir. & 75.1 & 84.6 & 81.7 & 41.2 & 48.4 & 66.2 & 0.107 \\
mean / last & bidir. & 80.2 & 90.9 & 84.8 & 51.8 & 48.9 & 71.3 & 0.074 \\
mean / mean & bidir. & 79.0 & 88.0 & 83.2 & 51.2 & 52.0 & 70.7 & 0.084 \\
last / mean & bidir. & 71.1 & 86.9 & 76.1 & 42.6 & 45.5 & 64.4 & 0.110 \\
\bottomrule
\end{tabular}
\caption{Held-out accuracy (\%) and mean ECE after calibration for the pooling--attention ablation. Pooling is question / option; means weight sources equally. Per-source accuracies are rounded to one decimal. Calibration leaves the predicted option and hence accuracy unchanged.}
\label{tab:pooling-heldout}
\end{table}

\begin{table}[!htbp]
\centering
\small
\begin{tabular}{@{}lrr@{}}
\toprule
Source & Causal & Bidirectional \\
\midrule
procedural-typed-decisions & 81.5--81.9 & 49.7--60.2 \\
typed-decisions & 74.9--77.2 & 61.5--68.7 \\
Cosmos QA & 73.5--74.2 & 61.6--66.9 \\
SQuAD 2.0 & 84.4--84.9 & 75.5--79.4 \\
BoolQ & 83.5--84.8 & 80.1--82.5 \\
MNLI & 84.0--84.3 & 80.3--82.9 \\
\bottomrule
\end{tabular}
\caption{Selected in-domain sources with substantial bidirectional accuracy reductions. Ranges are observed accuracy (\%), not confidence intervals.}
\label{tab:pooling-indomain}
\end{table}

\FloatBarrier
\section{Projection width by row and question shape}
\label{app:width}
\Cref{tab:width-breakdown} breaks down the projection-width sweep of \cref{sec:abl-width} by the number of questions in a row and the number of answer options.
In-domain, the spread across the four widths is at most 1.3 points in every bucket, and the best width changes from bucket to bucket.
Held-out, the largest spread is on questions with 9--20 options (93.3--96.9\%), which are almost all DBpedia-14, so it reflects one source rather than the number of options.
The 2- and 3-option held-out buckets hold 466 and 120 Kev transfer questions, too few to separate the widths.
\begin{table}[!htbp]
\centering
\small
\begin{tabular}{@{}lrrrrr@{}}
\toprule
& & \multicolumn{4}{c}{$W$} \\
\cmidrule(l){3-6}
& Questions & 64 & 128 & 256 & 512 \\
\midrule
\multicolumn{6}{@{}l}{\textit{In-domain, by questions per row}} \\
1 & 50,006 & 85.4 & 85.4 & 85.2 & 85.3 \\
2 & 8,166 & 89.9 & 90.1 & 90.4 & 90.4 \\
3 & 30,336 & 84.2 & 83.8 & 84.9 & 84.1 \\
4--5 & 21,038 & 74.0 & 74.7 & 73.5 & 74.5 \\
6+ & 19,993 & 76.4 & 76.8 & 76.6 & 76.8 \\
\midrule
\multicolumn{6}{@{}l}{\textit{In-domain, by answer options}} \\
2 & 39,694 & 88.2 & 87.9 & 88.3 & 88.2 \\
3 & 27,753 & 85.0 & 84.7 & 85.2 & 84.6 \\
4 & 7,128 & 81.0 & 81.3 & 81.9 & 80.8 \\
5--8 & 25,780 & 71.3 & 71.6 & 71.3 & 71.9 \\
9--20 & 16,764 & 74.2 & 75.0 & 73.7 & 74.7 \\
21+ & 12,420 & 90.7 & 90.8 & 90.7 & 90.9 \\
\midrule
\multicolumn{6}{@{}l}{\textit{Held-out, by answer options}} \\
2 & 466 & 62.9 & 66.7 & 62.7 & 64.4 \\
3 & 120 & 45.0 & 35.8 & 40.5 & 46.3 \\
4 & 8,670 & 80.4 & 81.7 & 81.4 & 81.0 \\
5--8 & 1,034 & 85.2 & 84.5 & 84.1 & 86.0 \\
9--20 & 70,182 & 93.4 & 93.3 & 96.9 & 95.6 \\
\bottomrule
\end{tabular}
\caption{Accuracy (\%) of the projection-width sweep (\cref{tab:width}) by number of questions in the row and by number of answer options. Unlike the other tables, accuracy is pooled over questions rather than averaged over sources, so large sources dominate each bucket. The held-out buckets are close to single sources: the 2- and 3-option buckets are all Kev transfer, and 70,000 of the 70,182 questions with 9--20 options are DBpedia-14.}
\label{tab:width-breakdown}
\end{table}

\FloatBarrier
\section{Retrieval-tuned checkpoints: per-source results}
\label{app:retrieval}
\Cref{tab:retrieval-ood} gives held-out accuracy per source for the retrieval-tuned checkpoints of \cref{sec:abl-retrieval} and the general Qwen3 checkpoints of the same size.
\begin{table}[!htbp]
\centering
\small
\begin{tabular}{@{}llrrrrrr@{}}
\toprule
Size & Backbone & AG News & DBpedia & QASC & Belebele & Kev & Mean \\
\midrule
0.6B & Qwen3 & \res{R68}{ag_news} & \res{R68}{dbpedia14} & \res{R68}{qasc} & \res{R68}{belebele_en} & \res{R68}{kev_transfer} & \res{R68}{ood} \\
 & Embedding & \res{R93}{ag_news} & \res{R93}{dbpedia14} & \res{R93}{qasc} & \res{R93}{belebele_en} & \res{R93}{kev_transfer} & \res{R93}{ood} \\
 & Reranker & \res{R94}{ag_news} & \res{R94}{dbpedia14} & \res{R94}{qasc} & \res{R94}{belebele_en} & \res{R94}{kev_transfer} & \res{R94}{ood} \\
\midrule
4B & Qwen3 & \res{R26}{ag_news} & \res{R26}{dbpedia14} & \res{R26}{qasc} & \res{R26}{belebele_en} & \res{R26}{kev_transfer} & \res{R26}{ood} \\
 & Embedding & \res{R97}{ag_news} & \res{R97}{dbpedia14} & \res{R97}{qasc} & \res{R97}{belebele_en} & \res{R97}{kev_transfer} & \res{R97}{ood} \\
 & Reranker & \res{R100}{ag_news} & \res{R100}{dbpedia14} & \res{R100}{qasc} & \res{R100}{belebele_en} & \res{R100}{kev_transfer} & \res{R100}{ood} \\
\midrule
8B$^{\ddagger}$ & Qwen3 & \res{R30b}{ag_news} & \res{R30b}{dbpedia14} & \res{R30b}{qasc} & \res{R30b}{belebele_en} & \res{R30b}{kev_transfer} & \res{R30b}{ood} \\
 & Embedding & \res{R103}{ag_news} & \res{R103}{dbpedia14} & \res{R103}{qasc} & \res{R103}{belebele_en} & \res{R103}{kev_transfer} & \res{R103}{ood} \\
 & Reranker & \res{R106}{ag_news} & \res{R106}{dbpedia14} & \res{R106}{qasc} & \res{R106}{belebele_en} & \res{R106}{kev_transfer} & \res{R106}{ood} \\
\bottomrule
\end{tabular}
\caption{Held-out accuracy (\%) by source for the retrieval-tuned checkpoints of \cref{tab:retrieval}, cosine head with last-token question and option pooling. Qwen3 is the general hybrid-thinking checkpoint. $^{\ddagger}$One epoch, LoRA learning rate $10^{-4}$.}
\label{tab:retrieval-ood}
\end{table}

\FloatBarrier
\section{decide-anything-27b: training settings}
\label{app:scale-up}
\Cref{tab:scale-up-recipe} lists how the model of \cref{sec:scale-up} was trained.
The learning rate was chosen by a sweep over $\{1, 2, 4, 5, 8\} \times 10^{-6}$ on a 78k-question subset of stage 1: rates from $2$ to $5 \times 10^{-6}$ tied on calibrated loss, and $4 \times 10^{-6}$ was best on held-out sets and calibration error.
\begin{table}[!htbp]
\centering
\small
\begin{tabular}{@{}lp{0.68\linewidth}@{}}
\toprule
Setting & Value \\
\midrule
Backbone & Qwen3.8-27B~\citep{qwen38}, 64 layers (48 Gated DeltaNet, 16 full attention), hidden size 5,120 \\
Trained parameters & every decoder layer and the final norm (24.36B); embeddings frozen \\
Head & cosine, $W{=}256$, last token pooling of question and option \\
Stage 1 & 1,179,792 questions in 810,811 rows (431.6M tokens); 1 epoch, 4,606 steps of 176 rows; learning rate $4 \times 10^{-6}$, constant after 400 warm-up steps \\
Stage 2 & 358,755 questions in 218,332 rows (130M tokens); continues from stage 1 with its optimiser state; 1,364 steps of 160 rows; learning rate $4 \times 10^{-6}$ decaying to 0 on a 1$-$sqrt curve; seeds 1--3 \\
Benchmark train splits & stage 2 includes the training splits of 13 of the 37 Decision Index 0.3 benchmarks \\
Averaging & uniform average of the three stage-2 models, head included \\
Head learning rate & $10^{-4}$ \\
Optimiser & AdamW, $\beta = (0.9, 0.999)$, $\epsilon = 10^{-8}$, weight decay 0.01 (0 on norms, biases and the scale), fp32 master weights, bf16 compute \\
Gradient clipping & global norm 1.0 \\
Augmentation & JSON states re-rendered with probability 0.3; question order shuffled \\
Maximum length & questions over 8,192 tokens dropped \\
Calibration & one temperature per question type, fitted on 10,501 held-out questions: 1.068 (yes/no), 1.041 (choice), 1.015 (score) \\
Hardware & 8 NVIDIA H100 80GB, FSDP2; 22 hours end to end \\
\bottomrule
\end{tabular}
\caption{Training settings for decide-anything-27b.}
\label{tab:scale-up-recipe}
\end{table}
\FloatBarrier

\end{document}